\pdfoutput=1

\documentclass[11pt]{article}

\usepackage[final]{acl}

\usepackage{times}
\usepackage{latexsym}

\usepackage[T1]{fontenc}

\usepackage[utf8]{inputenc}

\usepackage{microtype}

\usepackage{inconsolata}

\usepackage{graphicx}

\newcommand{\smallertexttt}[1]{{\fontsize{10}{12}\selectfont\texttt{#1}}}

\usepackage{subcaption}
\usepackage{mwe}

\usepackage{multirow}
\usepackage{microtype}

\usepackage{amsfonts}
\usepackage{amsmath}
\usepackage{amssymb}

\usepackage{multirow}

\usepackage{float}

\usepackage{svg}

\usepackage{booktabs}
\usepackage{enumitem}
\usepackage{subcaption}
\usepackage[colorinlistoftodos]{todonotes}

\title{Modular Arithmetic: Language Models Solve Math Digit by Digit}

\newcommand{\affilsup}[1]{\rlap{\textsuperscript{\normalfont#1}}}

\author{
    Tanja Baeumel\affilsup{1,2,3}
    \qquad
    Daniil Gurgurov\affilsup{1,2}
    \qquad
    Yusser Al Ghussin\affilsup{1,2} \\
    \textbf{Josef van Genabith}\affilsup{1,2}
    \qquad
    \textbf{Simon Ostermann}\affilsup{1,2,3}
    \\ \\
    $^1$German Research Center for AI (DFKI) \\
    $^2$ Saarland University \\
    $^3$Center for European Research in Trusted AI (CERTAIN) \\
    \smallertexttt{tanja.baeumel@dfki.de}
}

\begin{document}
\maketitle
\begin{abstract}
While recent work has begun to uncover the internal strategies that Large Language Models (LLMs) employ for simple arithmetic tasks, a unified understanding of their underlying mechanisms is still lacking. We extend recent findings showing that LLMs represent numbers in a digit-wise manner and present evidence for the existence of digit-position-specific circuits that LLMs use to perform simple arithmetic tasks, i.e. modular subgroups of MLP neurons that operate independently on different digit positions (units, tens, hundreds).  Notably, such circuits exist independently of model size and of tokenization strategy, i.e. both for models that encode longer numbers digit-by-digit and as one token.
\noindent Using Feature Importance and Causal Interventions, we identify and validate the digit-position-specific circuits, revealing a compositional and interpretable structure underlying the solving of arithmetic problems in LLMs. Our interventions selectively alter the model’s prediction at targeted digit positions, demonstrating the causal role of digit-position circuits in solving arithmetic tasks.
\end{abstract}

\section{Introduction}

The emergence of mathematical abilities in large language models (LLMs) has sparked growing interest in uncovering the internal mechanisms that underlie their arithmetic reasoning capabilities \cite{stolfo_mechanistic_2023, nikankin2024bagofheuristics, zhang2024interpretingimprovinglargelanguage, lindsey2025biology}.  
Understanding whether LLMs solve arithmetic by applying generalizable strategies, relying on superficial heuristics, or merely memorizing training examples can provide crucial insights into the general nature of reasoning abilities in LLMs.

\begin{figure}[t]
    \centering
    \includegraphics[width=0.41\textwidth]{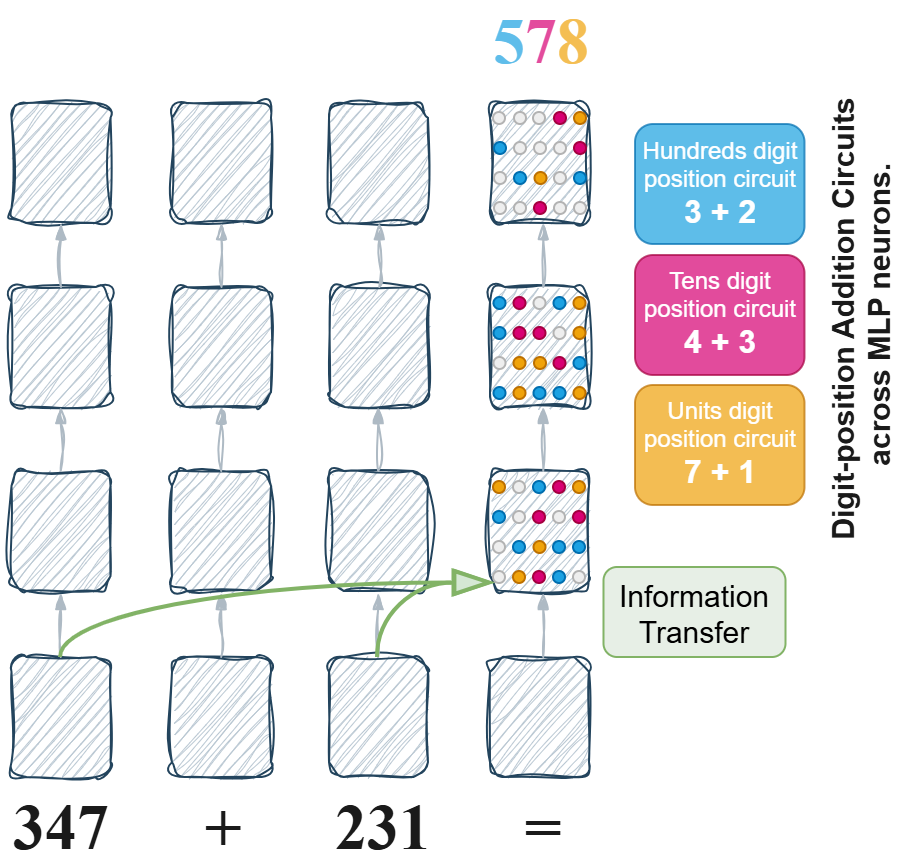} 
    \small
    \caption{Main finding: Simple arithmetic tasks are solved modularly by digit-position-specific circuits distributed across multiple MLP layers. Distinct sets of MLP neurons
    are responsible for generating results digits in parallel and independently for different positions. The hundreds digit position circuit is illustrated in blue, the tens in red, and units in yellow.}
    \label{fig:finding}
\end{figure}

Despite recent progress, a unified account of how LLMs perform basic arithmetic has yet to be established. 
\citet{nikankin2024bagofheuristics} argue that LLMs rely on a \textit{bag of heuristics}---a sparse set of neurons that are sensitive to simple operand or result patterns---rather than implementing coherent algorithms for arithmetic.  
\citet{lindsey2025biology} describe a dual-pathway mechanism for Claude-3.5 Haiku, with separate pathways for estimating rough result magnitude and the result unit digit.
At the same time, \citet{levy2024language} and \citet{gould2023successor} find that digit values -- but not full numeric values -- can be extracted from the residual stream via probing, suggesting that LLMs internally \textit{represent} numbers in a digit-wise manner. This finding is suprising, as many LLMs represent numbers up to a certain value (e.g. 3 digits) as single tokens (e.g., ``347'' is one token in Llama 3 8B). Their work however does not investigate whether these numeric representations are also exploited for solving arithmetic tasks.

In this work, we take the next step in understanding how LLMs solve arithmetic tasks by presenting causal evidence for digit-position-specific arithmetic circuits, i.e., modular subgroups of MLP neurons that independently generate results for different digit positions (Figure~\ref{fig:finding}).  
For instance, to solve \(347 + 231 =\), the model generates results for \(7 + 1\) (units), \(4 + 3\) (tens), and \(3 + 2\) (hundreds) using three distinct circuits, formed by groups of MLP neurons that are selectively sensitive to individual digit positions and distributed across mid-to-late layers.

\begin{figure*}[ht]
    \centering
    \begin{subfigure}[b]{0.94\textwidth}
        \includegraphics[width=\textwidth]{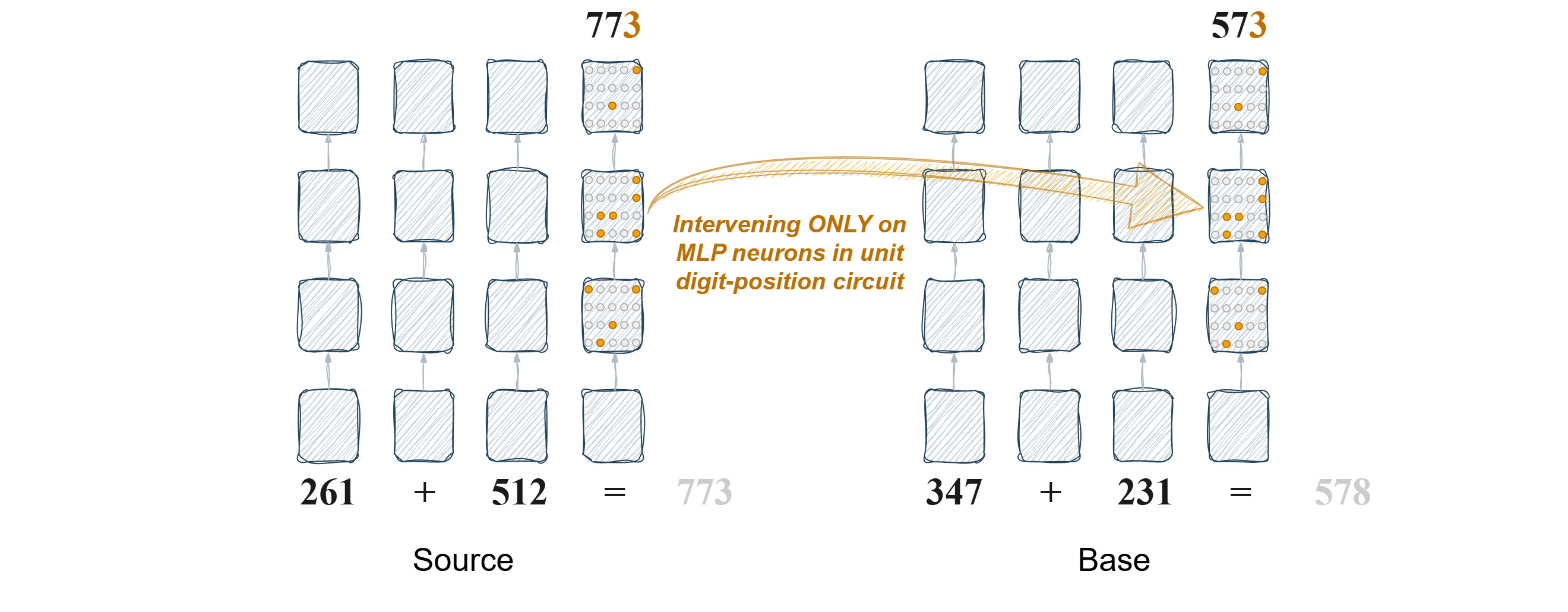}
    \end{subfigure}
    \caption{Intervening on only the MLP neurons that are members of one of the digit-position specific circuits, results in targeted changes only on the corresponding digit-position of the generated result. Here, an intervention on the unit position circuit only affects the unit position of the generated result ( \(3\) instead of \(8\)). This is evidence that the identified digit-position circuits are specific and causally involved in arithmetic result generation.}
    \label{fig:Pipeline}
\end{figure*}

We identify the digit-position-specific circuits using Fisher Score-based feature selection \cite{gu2012generalizedfisherscorefeature, SUN2021887} and validate their causal role via targeted interventions \cite{vig_investigating_2020,meng_locating_2023}: Altering the activation of the units-digit circuit for instance selectively changes only the predicted units digit while leaving tens and hundreds unchanged (Figure~\ref{fig:Pipeline}). 
Our findings offer important novel evidence that LLMs solve arithmetic tasks not merely through heuristics, but through structured, compositional arithmetic processes.

\noindent Our main contributions are: 

(1) We identify digit-position-specific arithmetic circuits by using supervised Fisher Score-based feature selection to detect MLP neuron groups responsible for solving arithmetic digit-level-subtasks.

(2) We validate these circuits via causal interventions, demonstrating that they selectively control individual output digits.

(3) We confirm the existence of digit-position-specific arithmetic circuits across a wide range of models, tasks, and tokenization schemes.

We release all code and data to support reproducibility\footnote{\url{https://github.com/tbaeumel/transformer-digit-arithmetic}}.
\section{Method}
\label{sec:methods}

\subsection{Overview of our Approach}
\label{sec:fisher}

We investigate whether LLMs perform arithmetic in a digit-wise, position-specific manner by a two-step approach, following recent work in mechanistic interpretability:

\begin{enumerate}
    \item \textbf{Identify Digit-Specific Neurons through Feature Selection:} We identify neurons that are involved in generating digit position specific subresults.  
    We use a Fisher Score-based feature selection method to measure neuron sensitivity to sub-tasks at different digit positions. 
    \item \textbf{Causal Verification via Interventions}: To verify the causal role of the identified neuron groups, we perform targeted interventionsby replacing the activation of a digit-specific neuron group in a base prompt with that of a source prompt. This allows us to test whether the altered activations lead to predictable changes in the model's output. Our results demonstrate that these neuron groups indeed implement digit-position-specific circuits: Only the digit corresponding to the intervened circuit changes, while the rest of the output remains unaffected. 
    As illustrated in Figure~\ref{fig:Pipeline}, intervening on the unit digit circuit of the base prompt \texttt{347 + 231} (expected output \texttt{578}) with activations from the source prompt \texttt{261 + 512} (expected output \texttt{773}) yields the output \texttt{573}. This output combines the hundreds and tens digits from the base result (\texttt{57}) with the unit digit from the source result (\texttt{3}), confirming the selective influence of the intervened circuit.
\end{enumerate}

\subsection{Data for Neuron Identification Experiments}
\label{sec:identificationdata}
We generate a simple addition dataset \(D_{add}\) and subtraction dataset \(D_{sub}\) by generating 1000 addition and 1000 subtraction prompts respectively. This data is used to identify digit-specific neurons (Step 1). The prompts are formulated in a one-shot setting of the form \texttt{``157 $\mathrm{o}$ 431 = 588; $A$ $\mathrm{o}$ $B$ = ''}, where $\mathrm{o} \in \{+, -\}$ and $A, B \in \{100, \dots, 999\}$. All addition and subtraction tasks are sampled such that the result is also a 3-digit integer, i.e., $\in \{100, \dots, 999\}$.
To isolate digit-wise computations and prevent interactions between digit positions, we construct all data such that no carry occurs in any digit position. We present an investigation of carry bits in Section \ref{sec:carry}.

\subsection{Data for Intervention Experiments}
\label{sec:interventiondata}

For the causal intervention experiments 
, we construct paired prompts in a one-shot setting consisting of a \textit{base} (e.g., \texttt{``157 $\mathrm{o}$ 431 = 588; $A$ $\mathrm{o}$ $B$ = ''}) and a \textit{source} (e.g., \texttt{``157 $\mathrm{o}$ 431 = 588; $C$ $\mathrm{o}$ $D$ = ''}) where operator $\mathrm{o} \in \{+, -\}$ and operands $A, B, C, D \in \{100, \dots, 999\}$),.

We construct two addition datasets \(D_{add,op1}\) and \(D_{add,op2}\) where operator $\mathrm{o} = +$, such that one operand is shared between base and source while the other varies across all digits (\(D_{add,op1}\): \(B=D\), \( A_{d_i} \neq C_{d_i} \forall \text{ digit positions } d_i \); \(D_{add,op2}\): \(A=C\), \( B_{d_i} \neq D_{d_i} \forall d_i \)). In the same way we construct \(D_{sub,op1}\) and \(D_{sub,op1}\)  where $\mathrm{o} = -$.

All datasets contain 200 pairs of unique arithmetic problems. To isolate digit-wise computations and prevent interactions between digit positions, we construct all data such that no carry occurs in any digit position.

\subsection{Models}
\label{sec:models}We evaluate four decoder-only transformer models: LLaMA3-8B, LLaMA3-70B \cite{grattafiori2024llama3herdmodels}, OLMo 2 7B \cite{olmo20242olmo2furious}, and Gemma 2 9B \cite{gemma_2024} all of which achieve high performance on simple addition and subtraction tasks ($>$95\%,  Table \ref{tab:model_accuracy}).

\begin{table}[h!]
\centering
\small
\resizebox{\columnwidth}{!}{%
\begin{tabular}{|l|l|c|c|}
\hline
\textbf{Numeric } & \textbf{Model} & \textbf{Accuracy } & \textbf{Accuracy } \\
\textbf{Tokenization} & & \textbf{Addition} & \textbf{Subtraction} \\
\hline
\multirow{3}{*}{Multi-digit} 
& LLaMA 3 8B & 100.00\% & 100.00\% \\
& LLaMA 3 70B & 100.00\% & 100.00\% \\
& Olmo 2 7B & 99.00\% & 99.50\% \\
\hline
Single-digit & Gemma 2 9B & 98.50\% & 99.50\% \\
\hline
\end{tabular}
}
\caption{Accuracy of models on \(D_{add}\) and \(D_{sub}\).
}
\label{tab:model_accuracy}
\end{table}

We focus our main analyses on models that employ multi-digit numeric tokenization strategies, i.e., their vocabularies contain multi-digit tokens like ``147'', i.e., LLaMA3-8B, Olmo 2 7B and LLaMA3-70B\footnote{All operands and correct results from all datasets appear as unique vocabulary tokens in all multi-digit tokenization models.}. Any observed digit-wise processing must be a property of the model's internal structure rather than a simple consequence of tokenization for such models, making it a more surprising and informative phenomenon. 
To demonstrate the generalizability of our findings, we also include results from a single-digit tokenization model (Gemma 2 9B) in Appendix \ref{app:single-digit}, which similarly exhibits digit-position-specific processing pathways.

To study the mechanisms that emerge from standard language model pretraining, we evaluate base models without any instruction tuning or arithmetic-specific fine-tuning.
\section{Circuit Localization}

We now describe the circuits localization steps in detail along with the results.
We focus our analyses on LLaMA3-8B in the main text and report detailed analyses showing similar results for Olmo 2 7B and LLaMA3-70B in Appendices \ref{app:circuit_statistics} and \ref{app:more_models}. 

\subsection{Step 1: Identifying Digit-Specific Neurons}
\label{sec:fisherselection}

\paragraph{Intuition.} To identify MLP neurons that are sensitive to digit-level arithmetic subtasks, we perform supervised feature selection. Intuitively, we identify hich neurons are most discriminative for different arithmetic subtasks in individual digit positions.

We employ Fisher Score \cite{dudapattern2001, venkatesh2019review} as our feature selection method, which is an established information-theoretic method for assessing individual feature importance in high-dimensional spaces.
Fisher Score measures how well a feature (in our case, a neuron’s activation) separates data points belonging to different classes (in our case, digit-level arithmetic tasks). 
A high Fisher score \( F_{i,d} \) of neuron \(i\) at digit position \(d\)  indicates that the activation of neuron \( i \) reliably discriminates between subtasks targeting that digit position. We interpret such neurons as likely members of digit-position-specific arithmetic subcircuits. 
For instance,  if a neuron is similarly activated for the addition tasks \texttt{``15\textbf{7} + 43\textbf{1} = ''}, \texttt{``23\textbf{7} + 36\textbf{1} = ''}, and \texttt{``76\textbf{7} + 21\textbf{1} = ''}, chances are that it is involved in the generation of the unit result digit, as all examples have the same subtask \(7+1\) in the unit position.

We collect MLP activations across layers, for all prompts in \(D_{add}\) and \(D_{sub}\) and compute Fisher Scores for each neuron and digit position. We then rank neurons according to their Fisher Score, reflecting how strongly their activations vary with changes to specific digit positions in the input operands. By thresholding the scores (as a hyperparameter), we obtain candidate circuits composed of neurons sensitive to each digit position.

Following \citet{stolfo_mechanistic_2023}, we focus on MLP layers that could plausibly contribute to producing arithmetic output. We identify the earliest layer at which operand information has been propagated to the residual stream at the final token, as earlier layers lack relevant input context and are unlikely to contribute to the result generation.
We therefore restrict our search for arithmetic circuits to layers at which operand information has been propagated. Appendix~\ref{app:injectionLayers} provides the identified ``operand injection'' layers for all models and tasks, alongside a more detailed rationale.

\paragraph{Neuron Selection for Digit-Position Circuits.}
Appendix \ref{app:FisherFormal} presents a detailed formalization of how Fisher Score  \( F_{i,d} \) is calculated for each neuron. 

For each MLP layer \(l\) in model \(m\) and \(\mathrm{o} \in {add, sub}\) in each digit position \( d \):
\begin{enumerate} [itemsep=2pt, topsep=2pt]
    \item Compute average \( F_{i,d} \) for all neurons \( i \) in \(l\) over the dataset \(D_{\mathrm{o}}\).
    \item Select the neurons with \( F_{i,d} \) above a threshold $t$ as candidates for the digit-specific arithmetic circuit \( C_{m, \mathrm{o}, d, t} \).
\end{enumerate}

For all models, we explore circuits \(C_{m, \mathrm{o}, d, t}\)  based on different thresholds \(t \in \{ 0.4, 0.5, 0.6, 0.7, 0.8, 0.9, 1.0\}\). Fisher Scores are a relative metric, therefore choosing a reasonable threshold for circuit membership is a hyperparameter tuning problem. Generally speaking, Fisher scores close to 0 indicate a lack of discriminative power, scores between 0.1 and 1.0 weak to moderate discriminability, and values above 1.0 strong discriminability.

\paragraph{Statistics on Selected Circuits.} 
We provide detailed statistics for all identified circuits \(C_{m, \mathrm{o}, d, t}\) across all models, operators, and digit positions in Appendix \ref{app:circuit_statistics} and provide a brief summary of the findings here.

We find that digit-position-specific circuits consist of a significant portion available MLP neurons, for instance for \(C_{\text{Llama3 8B}, \mathrm{+}, d, t}\) the average number of MLP neurons per layer responsible for one of the digit-position specific circuits is 60.3\% of all MLP neurons. Further, the sets of digit-position-specific MLP neurons are highly sufficient for representing digit-specific arithmetic subtasks, indicated by strong classificiation performance using only the digit circuit neurons. Importantly, we also find that neuron sets in different digit-position circuits are largely distinct (Figure \ref{fig:llama8_circuit_stats}), providing a first indication of digit-positional modularity in LLM arithmetic.

\subsection{Step 2: Causal Verification via Interventions}
\label{sec:intervention}

\begin{table*}[ht!]
\small
\centering
\resizebox{\textwidth}{!}{
\begin{tabular}{|c|c|c|c|c|c|c|c|c|c|c|c|}
\hline
\(m\) &\(\mathrm{o}\) & \(d\) &\(t*\) & bbb & \textbf{bbs} & \textbf{bsb} & \textbf{sbb} & bss & sbs & ssb & sss\\
\hline
\multicolumn{1}{|c|}{} & 
\multicolumn{1}{c|}{} & 
\multicolumn{1}{c|}{} & 
\multicolumn{1}{c|}{} & 
\multicolumn{8}{c|}{\textbf{Absolute change in prediction probability \textbf{\(\Delta p\)} in percentage points (after - before).}} \\
\hline
\multirow{6}{*}{Llama 3 8B} & \multirow{3}{*}{\(+\)} 
                            & unit & \(0.6\) & -78.89\% & \textbf{+30.93\%} & +0.87\% & +0.87\% & +10.59\% & +2.57\% & +0.49\% & +2.76\% \\
                           &  & tens & \(0.5\) & -57.29\% & +0.47\% & \textbf{+22.76\%}& +1.51\% & +1.34\% & +0.17\% & +4.57\% & +0.37\% \\
                           & & hun.s & \(0.9\) & -77.01\% & +0.06\% & +0.50 & \textbf{+45.56\%} & +0.10\% & +0.50\% & +5.04\% & +0.47\% \\
    \cline{2-12} 
    & \multirow{3}{*}{\(-\)} 
                        & unit & \(0.6\) & -75.62\% & \textbf{+28.50\%} & +0.59\% & +1.16\% & +12.40\% & +2.90\% & +0.25\% & +3.41\% \\
                         &       & tens & \(0.5\) & -55.97\% & +0.90\% & \textbf{+14.40\%} & +2.07\% & +2.04\% & +0.22\% & +6.06\% & +1.27\% \\
                         &       & hun.s & \(0.9\) & -62.20\% & +0.14\% & +0.58\% & \textbf{+36.27\%} & +0.12\% & +0.28\% & +1.88\% & +0.16\% \\
                           \hline
\multirow{6}{*}{Llama 3 70B} 
& \multirow{3}{*}{\(+\)} 
                        & unit & 0.5 & -36.93\% & \textbf{+23.16\%} & +0.72\% & +0.82\% & +3.67\% & +0.54\% & +0.15\% & +0.53\% \\
                        &  & tens & 0.5 & -28.67\% & +0.37\% & \textbf{+17.71\%} & +0.57\% & +0.46\% & +0.08\% & +1.73\% & +0.15\% \\
                        &  & hun.s & 0.6 & -41.40\% & +0.17\% & +0.51\% & \textbf{+29.68\%} & +0.12\% & +0.23\% & +1.14\% & +0.18\% \\
\cline{2-12}
& \multirow{3}{*}{\(-\)} 
                            & unit & 0.5 & -34.17\% & \textbf{+20.76\%} & +0.44\% & +0.32\% & +7.43\% & +0.23\% & +0.13\% & +1.08\% \\
                            &  & tens & 0.5 & -14.84\% & +0.07\% & \textbf{+11.32\%} & +0.19\% & +0.08\% & +0.03\% & +1.02\% & +0.06\% \\
                            &  & hun.s & 0.6 & -18.62\% & +0.05\% & +0.19\% & \textbf{+15.91\%} & +0.03\% & +0.06\% & +0.31\% & +0.04\% \\
\hline
\multirow{6}{*}{Olmo 2 7B} 
& \multirow{3}{*}{\(+\)} & unit 
                         & \(0.4\) & -86.16\% & \textbf{+27.81\%} & +0.33\% & +11.27\% & +1.51\% & +5.89\% & +0.95\% & +3.75\% \\
                           &  & tens & \(0.8\) & -83.21\% & +0.11\% & \textbf{+42.57\%} & +3.50\% & -0.02\% & +0.02\% & +7.20\% & +0.34\% \\
                           &  & hun.s & \(0.8\) & -77.76\% & +0.47\% & +0.16\% & \textbf{+37.37\%} & +0.35\% & +2.20\% & +2.77\% & +0.33\% \\
    \cline{2-12} 
    & \multirow{3}{*}{\(-\)} 
                        & unit & \(0.5\) & -89.44\% & \textbf{+50.26\%} & +0.10\% & +5.93\% & +0.83\% & +5.37\% & +0.20\% & +1.77\% \\
                           &  & tens & \(0.9\) & -84.64\% & -0.00\% & \textbf{+44.70\%} & +1.23\% & +0.28\% & +0.07\% & +5.29\% & +0.13\% \\
                           &  & hun.s & \(0.9\) & -86.63\% & +1.07\% & -0.40\% & \textbf{+45.28\%} & +0.19\% & +3.29\% & +3.49\% & +0.37\% \\

\hline
\end{tabular}
}
\caption{Main Results: for all detected circuits (across models, operators, digit positions) we report the change in prediction probabilities in percentage points (effect size) for result variants after interventions on digit-position-specific arithmetic circuits (with optimal \(t*\) as \(t\) in each circuit), on datasets \(D_{add, op2}\) and \(D_{sub, op2}\), with the increase in prediction probability for the targeted result variant shown in bold. Highly similar results for datasets \(D_{add, op1}\) and \(D_{sub, op1}\) are provided in Table \ref{tab:mainresultsop1} below.}
\label{tab:mainresultsop2}
\end{table*}

We perform targeted \textit{interchange interventions} \cite{vig_investigating_2020, meng_locating_2023} on candidate digit-position circuits \(C_{m, \mathrm{o}, d, t}\) replacing MLP activations of digit-position-specific neurons in a \textit{base} prompt with those from a \textit{source} prompt at selected layers.

\paragraph{Circuit Depth.}
Based on the observations in Section \ref{sec:fisherselection} and Appendix \ref{app:injectionLayers}, we choose the MLP layer set considered for digit-position specific arithmetic circuits to be \(L_{Llama 3 8B, add, op2} = \{15, ..., 24\}\) for \(D_{add,op2}\) and \(L_{Llama 3 8B, add, op1} = \{16, ..., 24\}\) for \(D_{add,op1}\). Ablation experiments (Appendix \ref{app:threshold_layer28}) show that choosing deeper circuits \(L_{Llama 3 8B, add, op2} = \{15, ..., 28\}\) and \(L_{Llama 3 8B, add, op2} = \{16, ..., 28\}\) does not make a significant difference. This supports our conclusion that late MLP layers are less responsible for generating digit-position specific arithmetic results.

\paragraph{Experiment.} Given a source (\texttt{123 + 562 = 685}) and base prompt (\texttt{123 + 456 = 579}), we intervene on the neurons in  \(C_{m, \mathrm{o}, d, t}\). We then measure the model's output distribution over all 8 possible digit-wise combinations of the base and source results (e.g., \texttt{bbb}, \texttt{sbb}, \texttt{bsb}, etc., where e.g. \texttt{sbb} indicates the the hundreths digits correspond to the source result and the other two digits to the base result). If our independent digit-wise circuit hypothesis is correct, a position-specific circuit should shift probability mass only toward the expected variant (e.g., \texttt{sbb} for a hundreds-digit intervention), with minimal effects on other positions.
 
We assess position-specificity of the intervention by checking whether the probability of the expected result variant increases selectively (e.g. for an intervention on hundreds digits \texttt{sbb}). 
If other variants like \texttt{ssb} or \texttt{sbs} were to also increase, this would indicate that the intervention affects multiple digit positions. The specificity of the probability shift provides key evidence for digit-position-specific circuits in the MLP layers.

\begin{figure*}[ht!]
    \centering
    \hfill
    \begin{subfigure}[b]{0.32\textwidth}
        \centering
        \includegraphics[width=\textwidth]{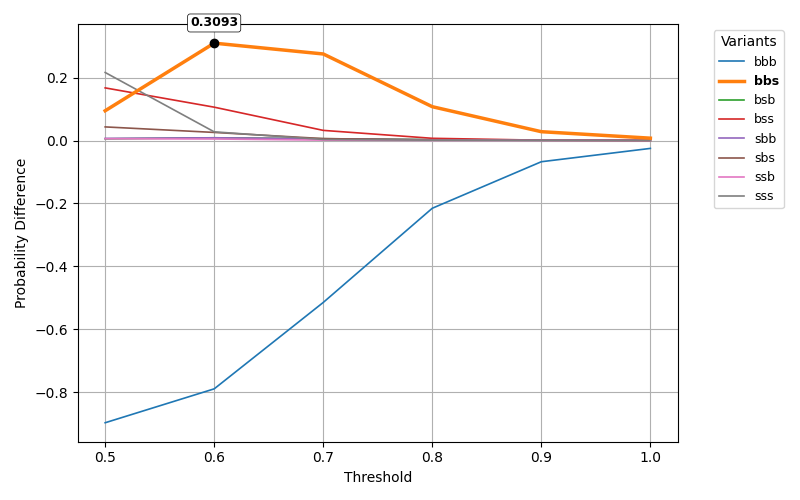}
        \caption{Unit Circuit intervention}
    \end{subfigure}
    \hfill
    \begin{subfigure}[b]{0.32\textwidth}
        \centering
        \includegraphics[width=\textwidth]{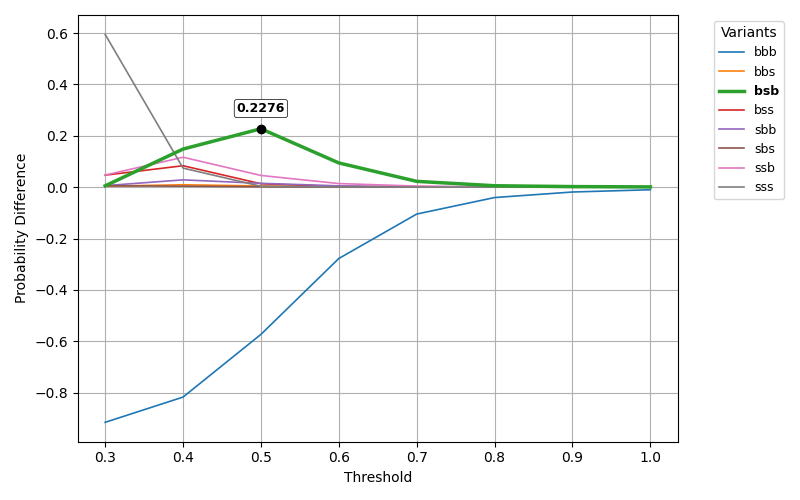} 
        \caption{Tens Circuit intervention}
    \end{subfigure}
    \begin{subfigure}[b]{0.32\textwidth}
        \centering
        \includegraphics[width=\textwidth]{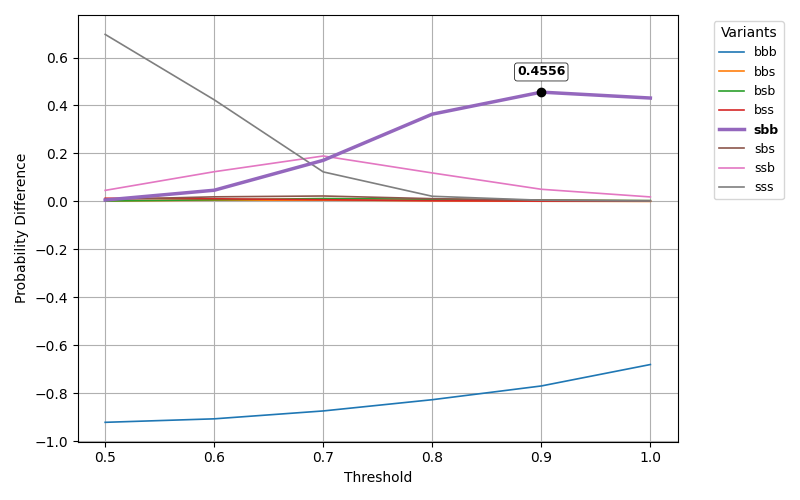}
        \caption{Hundreds Circuit Intervention}
    \end{subfigure}
    \caption{Effect size of digit-circuit interventions on \(D_{add, op2}\) with different thresholds for neuron circuit membership.}
    \label{fig:Threshold_Results}
\end{figure*}

\begin{table}[h]
\centering
\scriptsize
\begin{tabular}{|c|c|c|c|c|}
\hline
\(m\) &\(\mathrm{o}\) & \(d\) &\(t*\) & Flip Rate\\
\hline
\multirow{6}{*}{Llama 3 8B} & \multirow{3}{*}{\(+\)} & unit & \(0.6\) & 51\% \\
                           &  & tens & \(0.5\) & 33\% \\
                           & & hun.s & \(0.9\) & 68.5\% \\
                           \cline{2-5} 
                           & \multirow{3}{*}{\(-\)} & unit & \(0.6\) &  49\% \\
                           &  & tens & \(0.5\) & 19.5\% \\
                           & & hun.s & \(0.9\) & 51\% \\
                           \hline
\multirow{6}{*}{Llama 3 70B} & \multirow{3}{*}{\(+\)} & unit & \(0.5\) & 24\% \\
                           &  & tens & \(0.5\) & 15\% \\
                           & & hun.s & \(0.6\) & 33.5\% \\
                           \cline{2-5} 
                           & \multirow{3}{*}{\(-\)} & unit & \(0.5\) & 20.5\% \\
                           &  & tens & \(0.5\) & 7.5\% \\
                           & & hun.s & \(0.6\) & 10.5\% \\
                           \hline
\multirow{6}{*}{Olmo 2 7B} & \multirow{3}{*}{\(+\)} & unit & \(0.4\) & 44\% \\
                           &  & tens & \(0.8\) & 54\% \\
                           & & hun.s & \(0.8\) & 51.5\% \\
                           \cline{2-5} 
                           & \multirow{3}{*}{\(-\)} & unit & \(0.5\) & 64\% \\
                           &  & tens & \(0.9\) & 53\% \\
                           & & hun.s & \(0.9\) & 54.5\% \\
                           \hline
\end{tabular}
\caption{Flip rate from \texttt{bbb} result to the intended digit-specific result variant (Unit: \texttt{bbs}, Tens: \texttt{bsb}, Hundreds: \texttt{sbb}) at the best threshold \(t*\) for each circuit, on datasets \(D_{add, op2}\) and \(D_{sub, op2}\). Highly similar results for datasets \(D_{add, op1}\) and \(D_{sub, op1}\) in Table \ref{tab:fliprateop1}.}
\label{tab:fliprateop2}
\end{table}

\paragraph{Results.}
Table \ref{tab:mainresultsop2} contains results for the intervention on the best candidate digit-position circuits \(C_{m, \mathrm{o}, d, t*}\), where \(t*\) is the best value for threshold \(t\). 
For all 4 datasets \(D_{\mathrm{o}, op}\) (\(\mathrm{o} \in \{add, sub\}; \text{op} \in \{op1, op2\}\)) and all models, we observe a consistent and substantial increase of the probability of the expected result variant -- \texttt{bbs} for unit, \texttt{bsb} for tens, and \texttt{sbb} for hundreds -- demonstrating that the interventions on the hypothesized digit-position circuits indeed only affect the desired digit position. This is evidence for the digit-position-specific nature of the identified circuits, and thus the \textbf{causal role of digit-position specific processing in solving arithmetic tasks}.
For instance, an intervention on \(C_{\text{Llama 3 8B}, \text{add}, \text{hundreds}, 0.9}\) on \(D_{add, op2}\) yields a 45.56\% probability point increase on the \texttt{sbb} result variant (correct hundreds digit from the source, with tens and unit digits from the base), from 0.23\% before intervention to 45.79\% after intervention.
Specificity of the digit-position circuits is further supported by the low probability changes in result variants that contain multiple source digits (\texttt{bss}, \texttt{sbs}, \texttt{ssb}, and \texttt{sss}).

Additionally, Table \ref{tab:mainresultsop2} shows an expected reduction in the probability of the base result (\texttt{bbb}), indicating effective suppression of the default behavior. Importantly, non-target digits remain largely unaffected, as indicated by minimal changes observed in other result variants, confirming the \textbf{specificity} of the identified digit-position circuits.

We show detailed results on \(C_{\text{Llama 3 8B}, \text{add}, d, t}\) in Figure \ref{fig:Threshold_Results} to show the effect of the chosen threshold \(t\).
Additionally, Table \ref{tab:fliprateop2} shows that the intervention is often successful in flipping the model’s prediction from the \texttt{bbb} result to the intended digit-specific result variant. 

The results support our hypothesis: Intervening on the hypothesized circuits for the unit, tens, and hundreds digits leads to targeted and isolated changes in the corresponding digit of the model’s output. This is strong evidence that distinct circuits underlie the generation of arithmetic results in individual digits, and that the identified digit-position addition circuits are causally involved in generating digit-position specific sub-results.
\section{Supplementary Experiments}
\label{sec:supp}
To further strengthen our findings we present results from a series of supplementary experiments conducted on Llama 3 8B.

\subsection{Addition and Subtraction Circuits are Largely Distinct}
We are interested in whether subtraction and addition circuits are similar in their functionality or rather very distinct. 
To assess the similarity between addition and subtraction circuits, we investigate the overlap of selected MLP neurons for each digit position in each layer. 
Table~\ref{tab:add_sub} shows a low average overlap (Overlap in the top 100 MLP neurons: unit 19\%, tens 9.2\%, hundreds 19.8\%) in selected neurons, indicating that addition and subtraction circuits rely on mostly distinct subsets of neurons.

\subsection{Additions with Carry Bits Employ the Same Mechanisms}
\label{sec:carry}
All previous experiments explicitly avoided carry operations. However, carries are a key reason why fully modular digit-wise processing does not always suffice in arithmetic: the result at one digit position can influence another. For example, \(9 + 5\) in the units position yields a carry of \(1\) to the tens position.

To examine how our digit-wise arithmetic circuits interact with carry propagation, we test whether targeted interventions remain effective when they introduce a carry. Specifically, we ask: \textit{Does a digit-specific circuit encode carry information, or is it handled elsewhere in the network?}

We consider two scenarios involving carry-induced interventions (200 samples each):

\paragraph{(1) Carry from Units to Tens.}  
We intervene on the units-digit circuit of a base prompt with no carry, such as \(347 + 231 = 578\), using activations from a source prompt that induces a carry from the unit to the tens position, such as \(347 + 415 = 762\). After intervention on the units-digit circuit, we evaluate whether the output reflects (a) the base tens and hundreds digits with the source units digit (i.e., \(572 = bbs\)), or (b) if in addition to this a carry is propagated into the tens digit (i.e., \(582 = bb_{+1}s\)). 

\paragraph{(2) Carry from Tens to Hundreds.}  
We intervene on the tens-digit circuit of a no-carry base prompt (\(347 + 231 = 578\)) with a source prompt that causes a carry from tens to hundreds (\(347 + 482 = 829\)). We compare whether the model outputs (a) the base hundreds and units digits with the source tens digit (i.e., \(528 = bsb\)), or (b) if the hundreds digit is carry-adjusted (i.e., \(628 = b_{+1}sb\)).

These interventions help determine whether carry information is localized within digit-specific circuits or handled separately by the model.

\begin{figure}[t]
    \centering
    \includegraphics[width=0.35\textwidth]{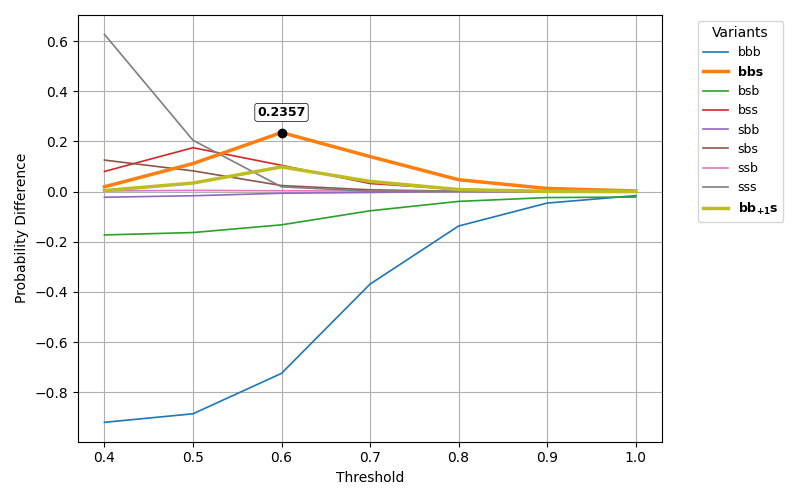} %
    \small
    \caption{Effect of intervening on the unit circuit with a carry originating from unit digit position.}
    \label{fig:carry_unit_pos}
\end{figure}

\begin{figure}[ht]
    \centering
    \includegraphics[width=0.35\textwidth]{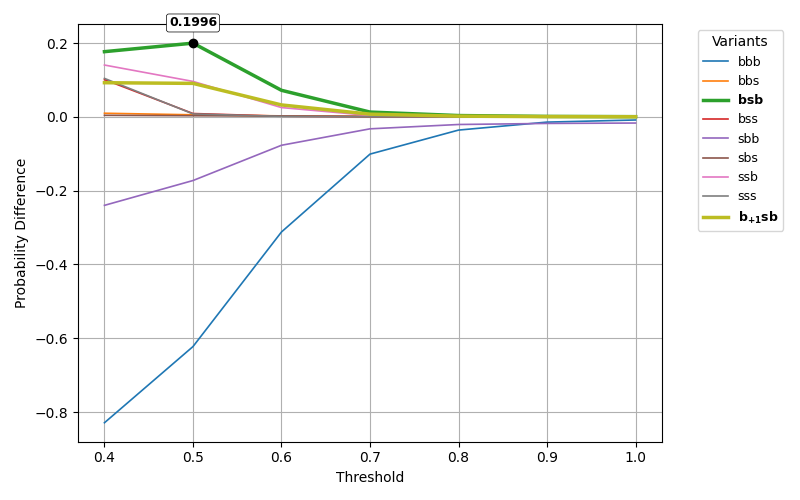} %
    \small
    \caption{Effect of intervening on the tens circuit with a carry originating from tens digit position.}
    \label{fig:carry_tens_pos}
\end{figure}

\paragraph{Results.}
Figures \ref{fig:carry_unit_pos} and \ref{fig:carry_tens_pos} show that the \texttt{bbs} variant and the \texttt{bsb} variant, for the unit and tens carry interventions, respectively, exhibit the strongest increase in probability. At the same time, the effect on the result variants \(bb_{+1}s\) and \(b_{+1}sb\) is significantly smaller.
This provides further evidence that digit-position-specific arithmetic circuits operate largely independently of one another and suggests that carry information is likely determined and processed by separate mechanisms, rather than being embedded in the digit-position circuits.

\subsection{MLP Similarity Reflects Task Overlap}
We investigate how similar digit-specific MLP activations are for prompts sharing the same digit-level arithmetic task at a given position (e.g., $154+635$ and $137+611$ share the same hundreds digit addition: $1+6$). Specifically, we compute pairwise cosine similarity between activations of MLP neurons in the digit-position circuit \(d\), for samples with the same subcomputation at digit position \(d\). 
As a baseline, we compute cosine similarity in these neurons across 5000 random sample pairs.

\begin{table}[ht]
\small
\centering
\resizebox{0.45\textwidth}{!}{
\begin{tabular}{|l|c|c|c|c|c|c|}
\hline
\multicolumn{1}{|l|}{Layer} &
\multicolumn{2}{c|}{Unit (\(t=0.6\))} &
\multicolumn{2}{c|}{Tens (\(t=0.5\))} &
\multicolumn{2}{c|}{Hundreds (\(t=0.9\))} \\
\cline{2-7}
& Sim & Random &
  Sim & Random &
  Sim & Random \\
&&(mean ± sd)&&(mean ± sd)&&(mean ± sd) \\
\hline
15 & \textbf{0.84} & 0.72 ± 0.08 & \textbf{0.83} & 0.73 ± 0.08 & \textbf{0.87} & 0.70 ± 0.11 \\
16 & \textbf{0.81} & 0.68 ± 0.07 & \textbf{0.78} & 0.66 ± 0.07 & \textbf{0.82} & 0.62 ± 0.11 \\
17 & \textbf{0.76} & 0.59 ± 0.08 & \textbf{0.73} & 0.59 ± 0.09 & \textbf{0.76} & 0.51 ± 0.14 \\
18 & \textbf{0.73} & 0.54 ± 0.11 & \textbf{0.72} & 0.58 ± 0.11 & \textbf{0.77} & 0.52 ± 0.15 \\
19 & \textbf{0.77} & 0.60 ± 0.10 & \textbf{0.75} & 0.61 ± 0.10 & \textbf{0.79} & 0.54 ± 0.16 \\
20 & \textbf{0.78} & 0.64 ± 0.09 & \textbf{0.77} & 0.63 ± 0.10 & \textbf{0.79} & 0.54 ± 0.15 \\
21 & \textbf{0.72} & 0.52 ± 0.12 & \textbf{0.70} & 0.50 ± 0.15 & \textbf{0.74} & 0.46 ± 0.18 \\
22 & \textbf{0.68} & 0.49 ± 0.14 & \textbf{0.68} & 0.46 ± 0.16 & \textbf{0.73} & 0.41 ± 0.21 \\
23 & \textbf{0.68} & 0.48 ± 0.12 & \textbf{0.67} & 0.47 ± 0.13 & \textbf{0.72} & 0.43 ± 0.20 \\
24 & \textbf{0.58} & 0.34 ± 0.15 & \textbf{0.53} & 0.34 ± 0.15 & \textbf{0.52} & 0.31 ± 0.27 \\
\hline
\end{tabular}
}
\caption{Pairwise cosine similarity of digit-position-specific MLP sub-updates on the same digit subtask. Includes a random-pair baseline for comparison.}
\label{tab:cosine_addition}
\end{table}

As expected, Table~\ref{tab:cosine_addition} shows consistently high within-label similarity across digit positions. For example, in layer 15, pairs sharing the same unit-digit computation have an average similarity of 0.84 in unit-digit MLP neurons.
This provides additional evidence for digit position-specific processing in mid-late MLP neurons.

\section{Structured Arithmetic Processing or a Bag of Heuristics?}
We are interested in how digit-position specific arithmetic processing fits in with previous findings that mathematical processing is solved by a sparse set of ``heuristic'' neurons that respond to specific operand or result patterns \cite{nikankin2024bagofheuristics}. These heuristics include, for example, results in a specific numerical range (e.g., results between 200 and 290), or results that share arithmetic properties (e.g., results congruent to \( 4 \mod 5 \)). 

We hypothesize that such heuristics may in part be fragments of digit-wise circuits, e.g., a neuron responding to results \( \equiv 4 \mod 5 \) may reflect unit-digit sensitivity, while neurons responding to result ranges (e.g., 200–290) may reflect sensitivity to higher digit positions.

To qualitatively test this idea, we perform a small-scale exploratory analysis. In the spirit of \citet{nikankin2024bagofheuristics} we examine digit-specific activation patterns (instead of analyzing neurons across full operand values) to look for heuristic patterns in individual MLP neurons. We analyze the 20 highest Fisher Score MLP neurons per digit circuit (units, tens, hundreds) per layer across layers 15–24 in LLaMA 3 8B. We then generate digit-wise heatmaps for each neuron, plotting the mean activation as a function of the digit values in operand 1 and operand 2, for each digit position.

\begin{figure}[ht]
    \centering
    \includegraphics[width=0.44\textwidth]{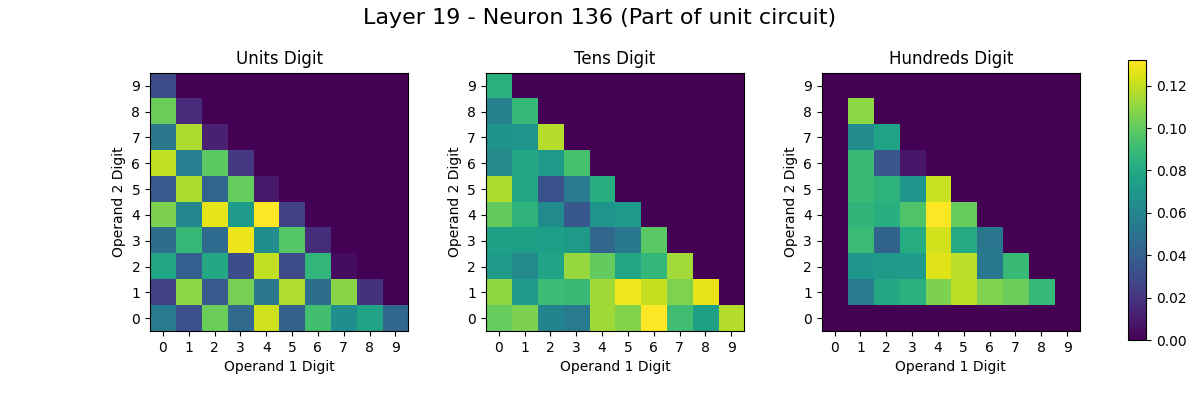} %
    \small
    \caption{Neuron \(N_{19, 136}\) (in unit circuit) implements parity heuristic, i.e., result is \(0 \mod 2\).}
    \label{fig:heurisit_parity}
\end{figure}

\begin{figure}[ht]
    \centering
    \includegraphics[width=0.44\textwidth]{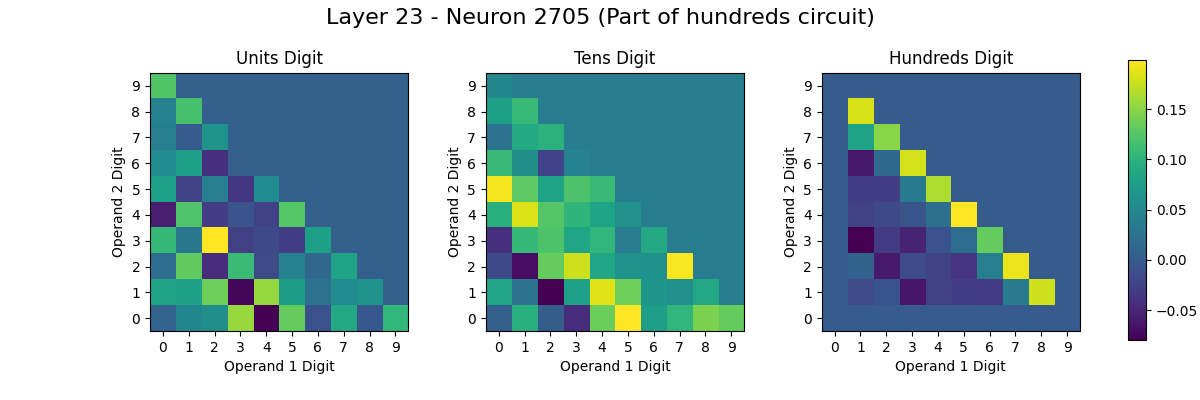} %
    \small
    \caption{Neuron \(N_{23, 2705}\) (in hundreds circuit) implements a result range heuristic (result in range 900 - 999), i.e., hundred digit in result is \(9\).}
    \label{fig:heurisit_resultrange}
\end{figure}

Our small qualitative study shows that many high-Fisher-score neurons exhibit highly structured and interpretable patterns, which mirror the kinds of heuristics described by \citet{nikankin2024bagofheuristics}: 
Certain neurons (Figures \ref{fig:heurisit_op1} and \ref{fig:heurisit_op2}) respond to specific operand digit values, while others are sensitive to certain result ranges (Figure \ref{fig:heurisit_resultrange}), result digits (Figure \ref{fig:heurisit_resultunit}), or result parity (Figure \ref{fig:heurisit_parity})\footnote{All plots in \url{https://github.com/tbaeumel/transformer-digit-arithmetic/tree/main/Digit-Heuristics_Top_Digit-Circuit_Neurons}}.

Importantly, the type of heuristic implemented by a neuron is typically aligned with the digit circuit that neuron belongs to, such that a range 900-999 result heuristic is implemented by a neuron within the hundreds-digit circuit, and the \(0 \mod 2\) result heuristic neuron is located within the unit-digit circuit.

While these are preliminary results they suggest that heuristic neurons may reflect digit-specific processing embedded within broader modular arithmetic circuits. In this view, heuristic neurons may be a manifestation of modular arithmetic circuits, rather than an alternative to them: they may be single-neuron approximations of low-level arithmetic subroutines tied to individual digit positions. While our analysis is qualitative and limited in scope, it highlights that compositional mechanisms and heuristic rules need not be mutually exclusive.
\section{Related Work}
\label{sec:related-work}

\textit{Mechanistic Interpretability.} Mechanistic interpretability (MI) aims to reverse-engineer the internal mechanisms of language models by analyzing weights and components. In transformer LLMs, a \emph{circuit} refers to a small set of interconnected components (e.g., MLPs or attention heads) that collectively perform a specific computation \cite{olah2020zoom, elhage2021mathematical}. Causal mediation techniques such as activation patching \cite{vig_investigating_2020, meng_locating_2023} enable demonstrating the causal role of these circuits during generation. In this work, we identify digit-position-specific arithmetic circuits and verify their causal contribution through targeted interventions.

\textit{Arithmetic Reasoning Interpretability.} Recent studies have begun to uncover how LLMs tackle arithmetic tasks \cite{stolfo_mechanistic_2023, nikankin2024bagofheuristics, zhang2024interpretingimprovinglargelanguage, lindsey2025biology, Baeumel2025TheLL, zhou2024transformersachievelengthgeneralization}. For example, \citet{lindsey2025biology} describe a dual-pathway mechanism in a model, where one pathway estimates the rough magnitude of results and the other generates the precise unit digit. \citet{nikankin2024bagofheuristics} argue that LLMs rely on a “bag of heuristics” rather than a single coherent algorithm, with individual neurons implementing simple heuristics like “result \% 5 = 0.” \citet{stolfo_mechanistic_2023} and others have used circuit analysis to reveal internal processing during arithmetic, while \citet{deng2024language} suggest LLMs mainly perform symbolic pattern recognition, not true numerical computation. \citet{kantamneni2025languagemodelsusetrigonometry} recently proposed that LLMs represent numbers as generalized helices and perform addition using a “Clock” algorithm \cite{nanda_progress_2023}. Despite this progress, a unified understanding of how LLMs perform basic arithmetic remains open. 

\textit{Number Representation.} Understanding how LLMs represent numbers internally has attracted considerable attention. \citet{levy2024language} show that probes on LLM hidden states fail to recover exact numeric values directly, but succeed in recovering each digit in base 10, indicating digt-wise representation in of numbers in LLMs. This extends findings by \citet{gould2023successor}, who showed LLMs encode numeric values modulo 10. Related work \cite{zhu-etal-2025-language, marjieh2025number} suggests number representations can blend string-like and numerical forms or be encoded linearly. Our work is inspired by these representation-focused studies and shows that numbers are not only represented digit digit but also processed this way in simple arithmetic tasks.

\section{Conclusion}

We identify and validate digit-position-specific arithmetic circuits in LLMs. Through targeted interventions, we demonstrate that these circuits are not only highly specific and modular but also causally involved in generating individual digit outputs.
 
Our results suggest LLMs perform simple arithmetic digit by digit, with distinct circuits operating independently at the units, tens, and hundreds positions.
This modular arithmetic structure is compatible with prior findings that LLMs use heuristic pathways and distributed neuron groups for arithmetic \citep{nikankin2024bagofheuristics}, however our results show LLMs solve arithmetic tasks in a far more organized way than previously thought.

\section*{Limitations}
While our results provide strong evidence for digit-position-specific arithmetic circuits in LLMs, our analysis is limited to addition and subtraction. More complex operations, such as multiplication and division, are not addressed in this work. Extending our framework to these tasks is an important direction for future work.
We also leave the analysis of circuits responsible for composing digit-level results into final outputs to future work.

Our analysis focuses on MLP layers, which contain the neurons that directly control digit outputs. This captures the core computation but does not account for the role of attention heads or other residual stream components.

Finally, we use Fisher Score to identify neurons involved in arithmetic, as it is simple and effective across model scales. However, other methods—such as gradient-based attribution—may yield more precise circuits. Despite this, our identified circuits are causal, interpretable, and robust, demonstrating a strong proof of concept for neuron-level circuit discovery in language models.

\section*{Acknowledgements}
We thank Cennet Oguz for her helpful feedback on the paper draft. This work has been supported by the German Ministry of Education and Research (BMBF) as part of the project TRAILS (01IW24005).

\bibliography{anthology, anthology1, anthology2, custom, custom1}

\appendix
\section{Implementation Details}
For our intervention experiments, we use the pyvene library \cite{wu-etal-2024-pyvene} to perform interchange interventions, where 
we intervene at a specific layer on a specific module at the last token position of a base prompt, with the corresponding activation from a source prompt (\textit{Interchange Intervention}).
\section{Results on Single Digit Tokenization Models}
\label{app:single-digit}
We present results from Gemma 2 9B, a single-digit tokenization model, to investigate whether it also exhibits digit-position-specific processing pathways.

\paragraph{Intuition: Single- vs Multi-digit Numeric Tokenization.}
Before presenting results on Gemma 2 9B, we clarify why arithmetic processing may differ in models that tokenize numbers into single digits (e.g., ``3'', ``4'', ``7'') versus a single multi-digit token (e.g., ``347''). 

The key distinction lies in the model's output, not its input. A multi-digit tokenization model, such as those in the Llama 3 family, tokenizes an input like ``347 + 231 = '' as [347, +, 231, =], and it is trained to generate a full multi-digit output token, such as ``578''. In contrast, a single-digit tokenization model, such as from the Gemma 2 family, tokenizes the same input as [3, 4, 7, +, 2, 3, 1, =] and can only generate single-digit output tokens. It will therefore output ``5'', not ``578''. 

As a result, the two models learn to solve fundamentally different arithmetic tasks. Multi-digit models must solve the entire arithmetic expression in a single forward pass, as they need the full answer immediately. Single-digit models, on the other hand, only need to generate the next digit, deferring the rest of the arithmetic task to future forward passes.

Therefore, when we expect single-digit tokenization models to contain different processing paths compared to multi-digit tokenization models, it is because these models have to learn completely different tasks during pretraining to become sufficient at solving arithmetics. 

\paragraph{Localizing Digit-Specific Neurons.}

\begin{figure*}[t!]
    \centering

    \begin{subfigure}[t]{\textwidth}
        \centering
        \begin{subfigure}[b]{0.32\textwidth}
            \centering
            \includegraphics[width=\textwidth]{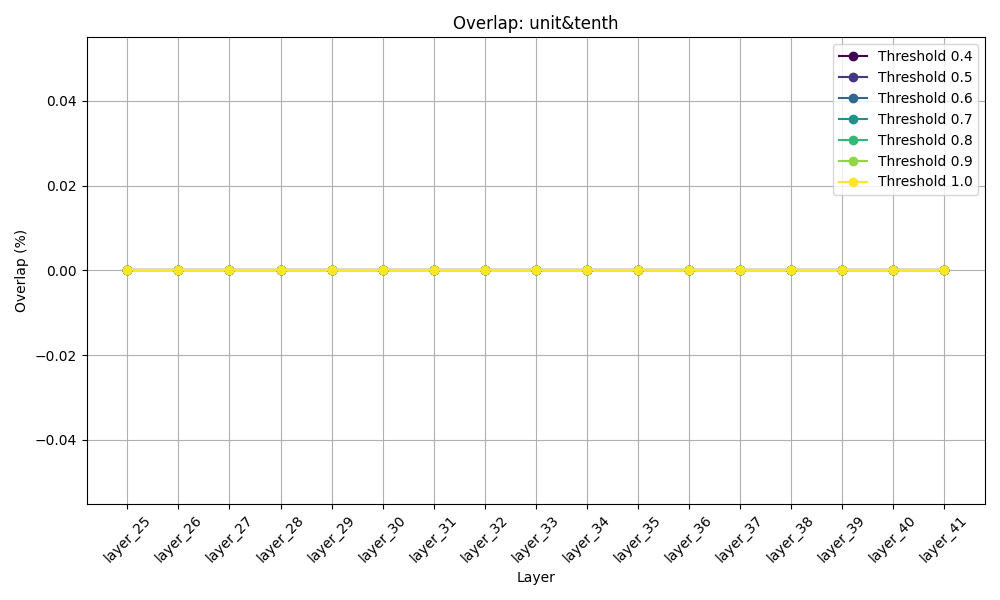}
            \caption{Unit and Tens}
            \label{fig:sub_overlap_unit_tenth_add_gemma}
        \end{subfigure}
        \hfill
        \begin{subfigure}[b]{0.32\textwidth}
            \centering
            \includegraphics[width=\textwidth]{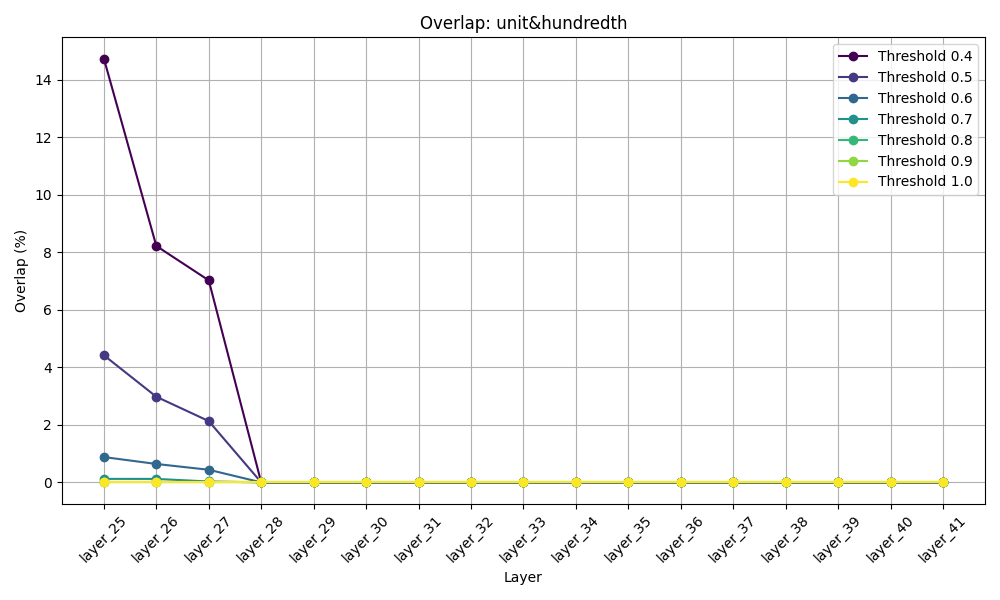}
            \caption{Unit and Hundreds}
            \label{fig:sub_overlap_unit_hundredth_add_gemma}
        \end{subfigure}
        \hfill
        \begin{subfigure}[b]{0.32\textwidth}
            \centering
            \includegraphics[width=\textwidth]{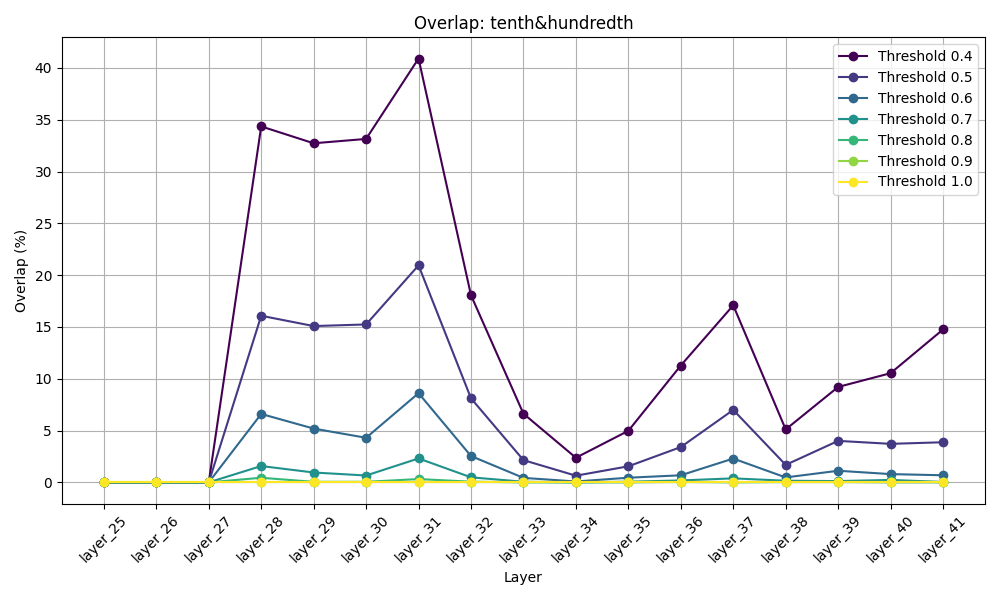}
            \caption{Tens and Hundreds}
            \label{fig:sub_overlap_tenth_hundredth_add_gemma}
        \end{subfigure}
        \caption{\textit{Circuit Overlap}: Overlap in neurons (\%) between digit-position circuits.}
        \label{fig:sub_row_overlap_add_gemma}
    \end{subfigure}

    \vspace{1em}

    \begin{subfigure}[t]{\textwidth}
        \centering
        \begin{subfigure}[b]{0.32\textwidth}
            \centering
            \includegraphics[width=\textwidth]{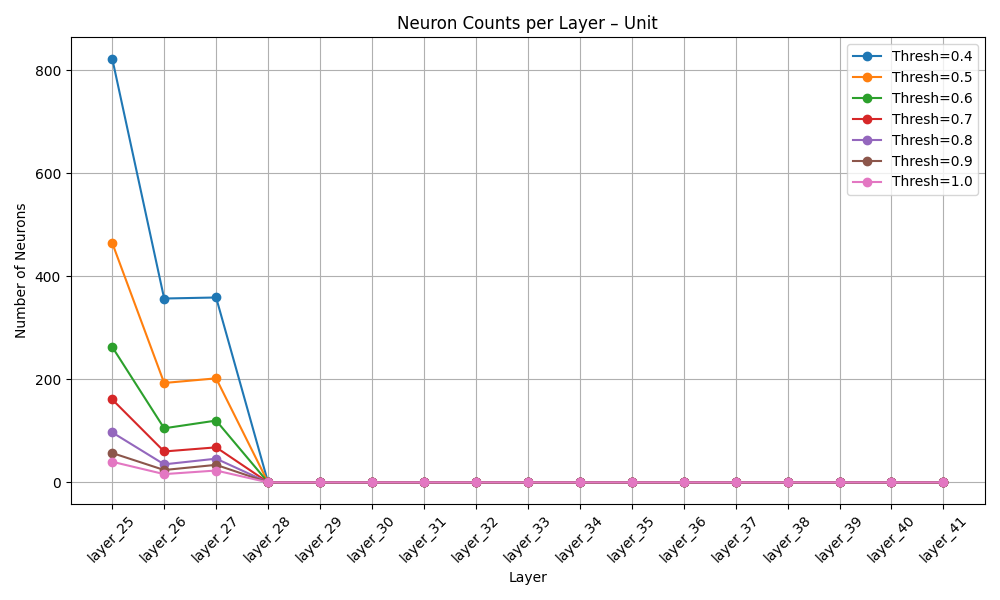}
            \caption{Unit}
            \label{fig:sub_counts_unit_add_gemma}
        \end{subfigure}
        \hfill
        \begin{subfigure}[b]{0.32\textwidth}
            \centering
            \includegraphics[width=\textwidth]{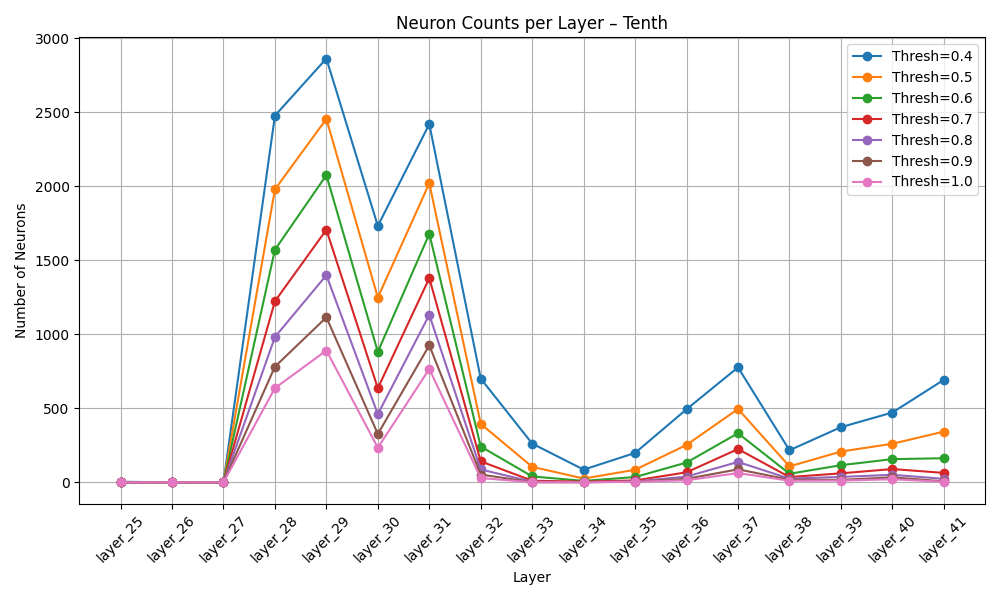}
            \caption{Tens}
            \label{fig:sub_counts_tenth_add_gemma}
        \end{subfigure}
        \hfill
        \begin{subfigure}[b]{0.32\textwidth}
            \centering
            \includegraphics[width=\textwidth]{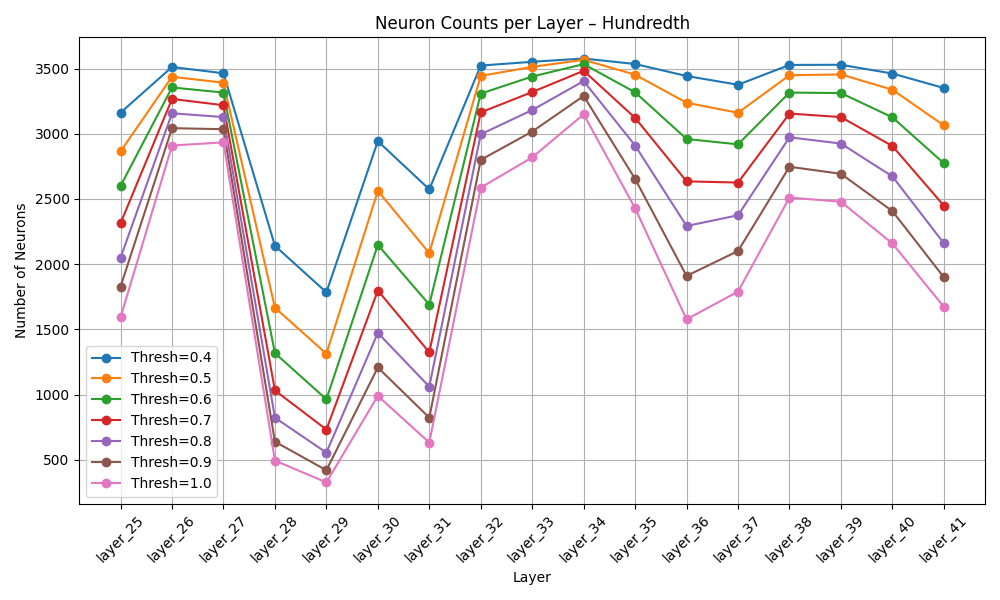}
            \caption{Hundreds}
            \label{fig:sub_counts_hundredth_add_gemma}
        \end{subfigure}
        \caption{\textit{Circuit Size}: Number of neurons per layer in digit-position circuits (Hidden size = 3584)}
        \label{fig:sub_row_counts_add_gemma}
    \end{subfigure}

    \caption{Gemma 2 9B, \(D_{add}\): Circuit statistics across digit positions and thresholds.}
    \label{fig:add_gemma9b_stats}
\end{figure*}

\begin{figure*}[t!]
    \centering

    \begin{subfigure}[t]{\textwidth}
        \centering
        \begin{subfigure}[b]{0.32\textwidth}
            \centering
            \includegraphics[width=\textwidth]{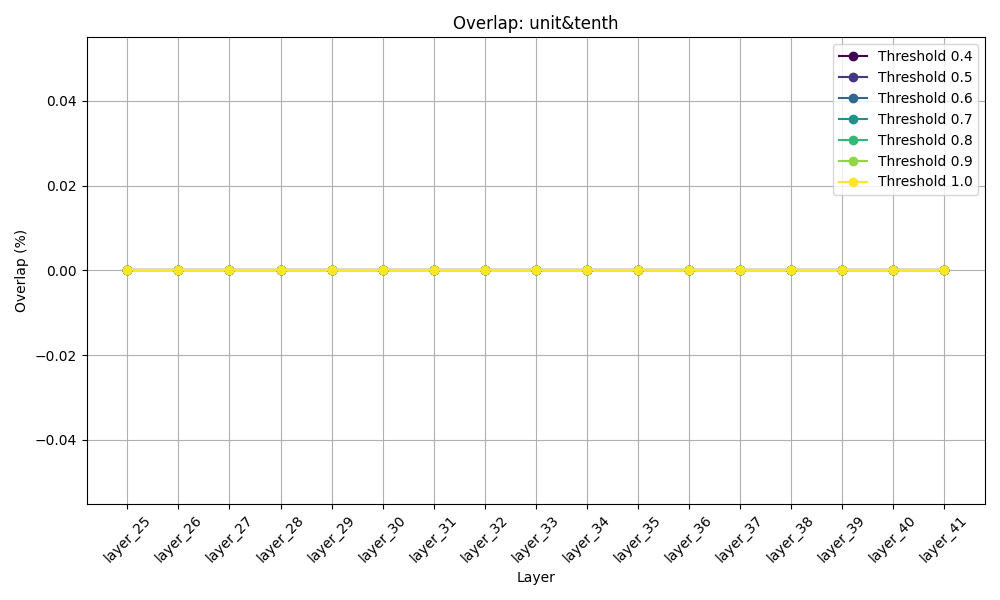}
            \caption{Unit and Tens}
            \label{fig:sub_overlap_unit_tenth_sub_gemma}
        \end{subfigure}
        \hfill
        \begin{subfigure}[b]{0.32\textwidth}
            \centering
            \includegraphics[width=\textwidth]{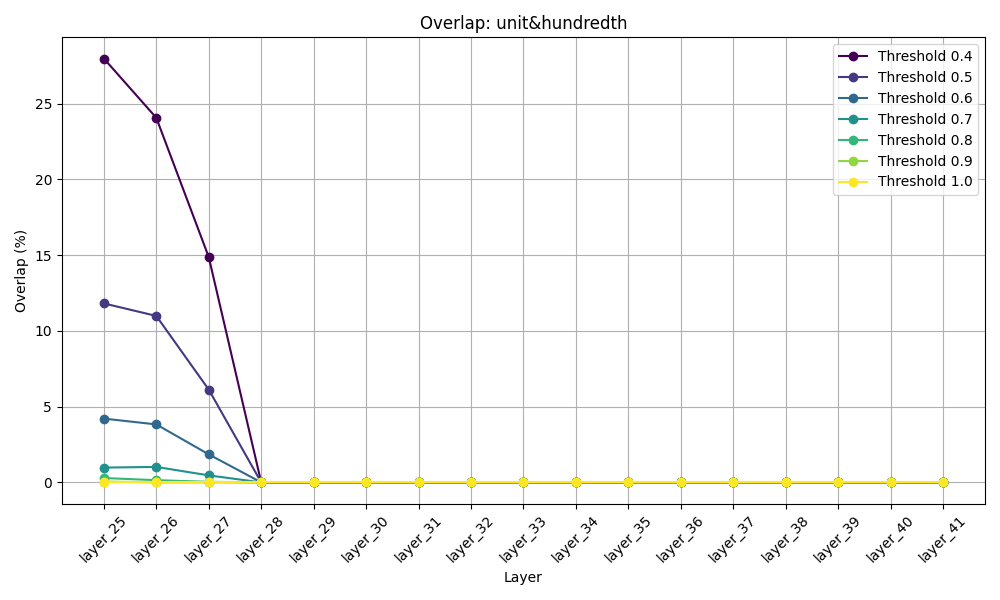}
            \caption{Unit and Hundreds}
            \label{fig:sub_overlap_unit_hundredth_sub_gemma}
        \end{subfigure}
        \hfill
        \begin{subfigure}[b]{0.32\textwidth}
            \centering
            \includegraphics[width=\textwidth]{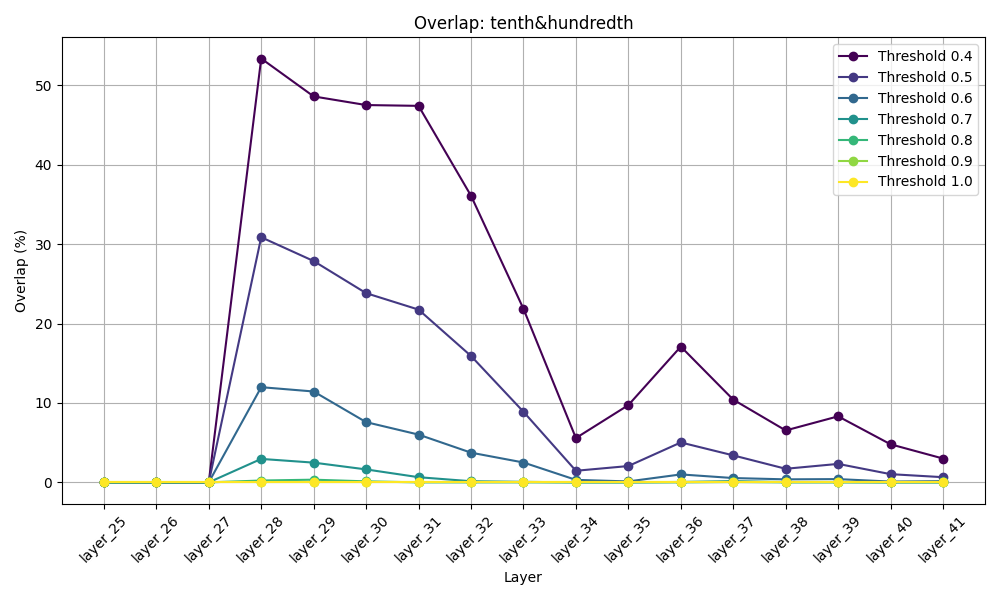}
            \caption{Tens and Hundreds}
            \label{fig:sub_overlap_tenth_hundredth_sub_gemma}
        \end{subfigure}
        \caption{\textit{Circuit Overlap}: Overlap in neurons (\%) between digit-position circuits.}
        \label{fig:sub_row_overlap_sub_gemma}
    \end{subfigure}

    \vspace{1em}

    \begin{subfigure}[t]{\textwidth}
        \centering
        \begin{subfigure}[b]{0.32\textwidth}
            \centering
            \includegraphics[width=\textwidth]{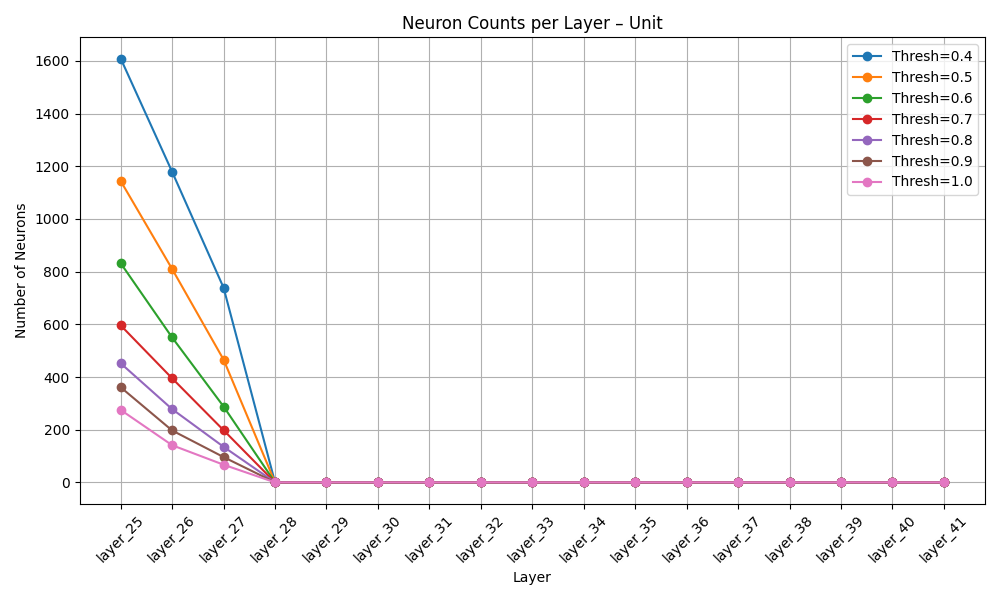}
            \caption{Unit}
            \label{fig:sub_counts_unit_sub_gemma}
        \end{subfigure}
        \hfill
        \begin{subfigure}[b]{0.32\textwidth}
            \centering
            \includegraphics[width=\textwidth]{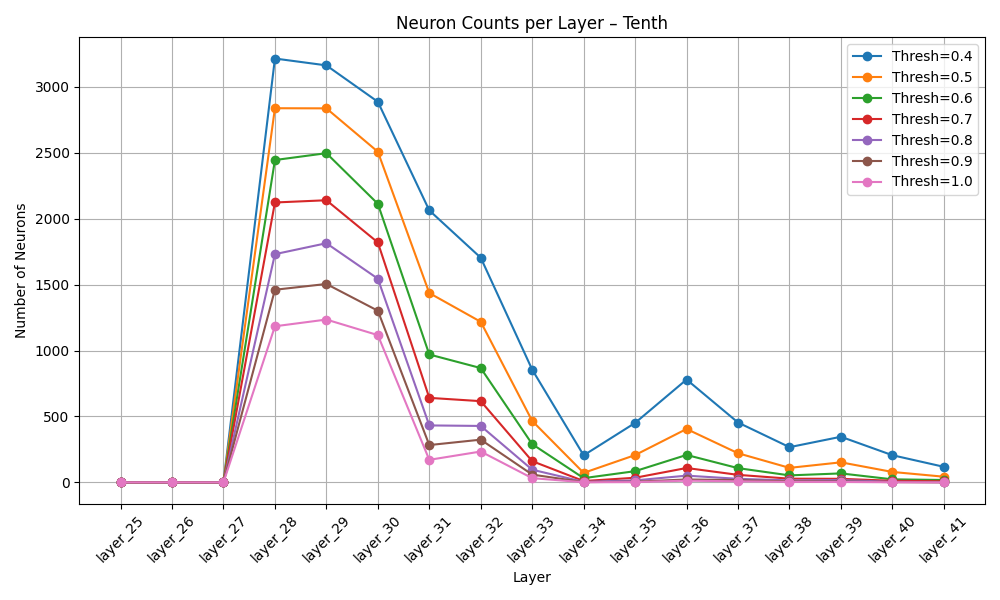}
            \caption{Tens}
            \label{fig:sub_counts_tenth_sub_gemma}
        \end{subfigure}
        \hfill
        \begin{subfigure}[b]{0.32\textwidth}
            \centering
            \includegraphics[width=\textwidth]{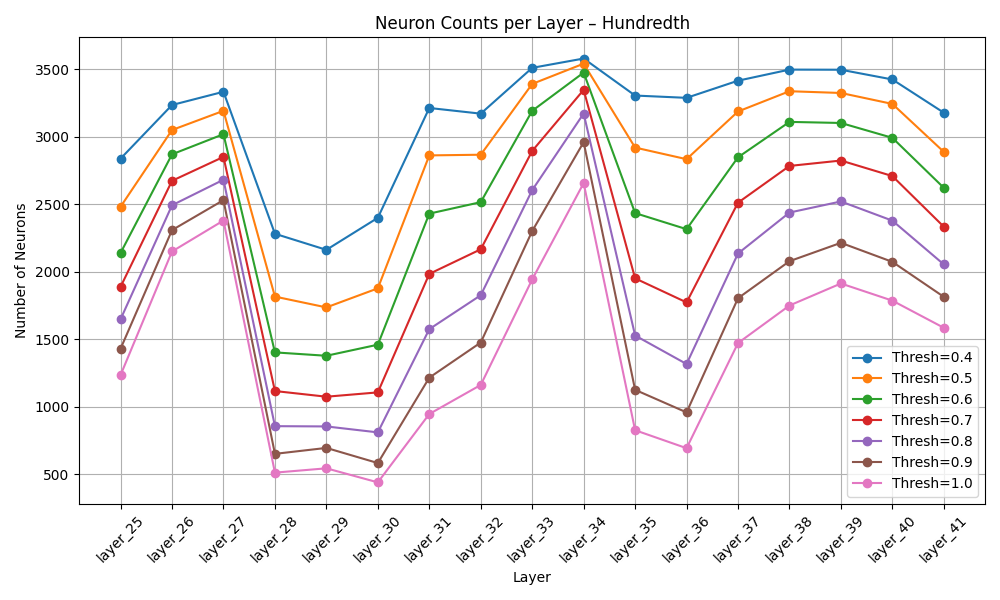}
            \caption{Hundreds}
            \label{fig:sub_counts_hundredth_sub_gemma}
        \end{subfigure}
        \caption{\textit{Circuit Size}: Number of neurons per layer in digit-position circuits (Hidden size = 3584).}
        \label{fig:sub_row_counts_sub_gemma}
    \end{subfigure}

    \caption{Gemma 2 9B, \(D_{sub}\): Circuit statistics across digit positions and thresholds.}
    \label{fig:sub_gemma9b_circuits}
\end{figure*}

After the injection of operand information (Layer 28; Figures \ref{fig:AttGemma9BSub} and \ref{fig:AttGemma9BAdd} in Appendix \ref{app:injectionLayers}) we observe the following pattern for both addition and subtraction (Figures \ref{fig:add_gemma9b_stats} and \ref{fig:sub_gemma9b_circuits}):

The hundreds circuit is large in size (spanning ~70-90\% of all MLP neurons per layer on average) and present all the way to the final layer. The tens circuit is available and large in size in layers 28 to 31, but then drastically diminishes in size. A unit circuit is non-existent, as evident by the absence of any neurons in the circuit, meaning that no neuron is sensitive to the subtask in the unit position. 

Based on the intuition given above and previous work that finds LLMs internally generate one result digit more than needed for their current generation to account for carry bits, but no further digits \cite{Baeumel2025TheLL}, we conclude the following: Single-digit tokenization models also have digit-position-specific arithmetic circuits, which consist of one dominant circuit responsible for predicting the digit needed for the output generation, and one smaller and shorter circuit which may be responsible for determining whether a carry-bit influences the generation.

\paragraph{Intervention.}
Although somewhat trivial, we intervene on the hundreds digit circuit to observe the probability of the hundreds base digit \texttt{b} and hundreds source digit \texttt{s}. Table \ref{tab:gemmaresults} shows that the intervention is highly effective in changing the model's prediction, which is expected given that the hundreds digit circuit spans the majority of MLP neurons, making the intervention highly invasive.

\begin{table}[h!]
\small
\centering
\begin{tabular}{|c|c|c|c|}
\hline
\(\mathrm{o}\) &\(t*\) & b & s\\
\multicolumn{1}{|c|}{} & 
\multicolumn{1}{|c|}{} & 
\multicolumn{2}{c|}{\textbf{\(\Delta p\)} (after - before).} \\
\hline
\multirow{5}{*}{\(+\)} & \(0.6\) & -92.61\% & +91.55\% \\
                       & \(0.7\) & -92.28\% & +91.26\% \\
                       & \(0.8\) & -91.93\% & +90.93\% \\
                       & \(0.9\) & -91.54\% & +90.55\% \\
                       & \(1.0\) & -91.25\% & +90.34\% \\
\hline
\multirow{5}{*}{\(-\)} & \(0.6\) & -92.6\% & +91.84\% \\
                       & \(0.7\) & -92.32\% & +91.54\% \\
                       & \(0.8\) & -91.76\% & +90.95\% \\
                       & \(0.9\) & -90.52\% & +89.63\% \\
                       & \(1.0\) & -88.05\% & +86.92\% \\
\hline
\end{tabular}
\caption{Intervention on the hundreds digit circuit in Gemma 2 9B on \(D_{add, op2}\) and \(D_{sub, op2}\). We report the change in prediction probabilities in percentage points (effect size) for result variants after intervention.}
\label{tab:gemmaresults}
\end{table}

\clearpage
\newpage
\section{Layers Involved in Arithmetic}
\label{app:injectionLayers}

Following \citet{stolfo_mechanistic_2023}, we focus our search for digit-position-specific arithmetic circuits on MLP layers that could plausibly contribute to producing arithmetic output. We identify the earliest layer at which operand information has been propagated to the residual stream at the final token, as earlier layers lack relevant input context and are unlikely to contribute to the result generation. We therefore restrict our search for arithmetic circuits to layers at which operand information has been propagated.

We perform causal interventions on the output of individual modules - particularly each decoder, attention, and MLP block. 
We intervene during the forward pass on a \textit{base} prompt (e.g., \texttt{'347 + 231 = ' }) 
by replacing the activations of a specific module with the activations of the same module during the forward pass on a \textit{source} prompt (e.g., \texttt{'261 + 512 = '}).
We measure the change in output probabilities for the correct result of the base query (\texttt{bbb}, here, '578') and the result of the source query (\texttt{sss}, here,  '773'). \\

\begin{figure*}[ht]
    \centering
    \begin{subfigure}[b]{0.31\textwidth}
        \centering
        \includegraphics[width=\textwidth]{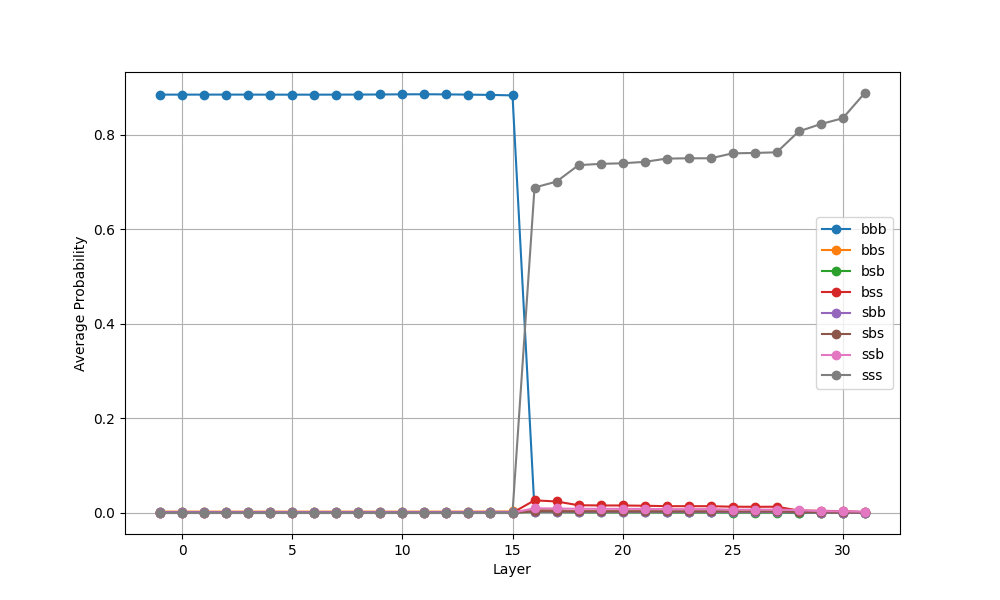}
        \caption{Block Output - \(D_{add, op1}\)}
    \end{subfigure} 
    \hfill
    \begin{subfigure}[b]{0.31\textwidth}
        \centering
        \includegraphics[width=\textwidth]{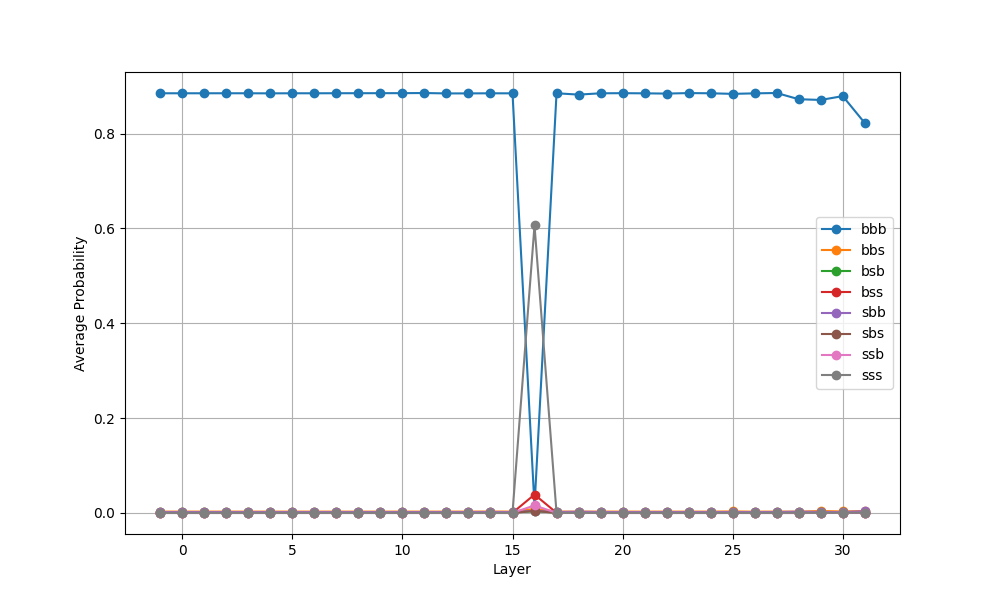}
        \caption{Attention Output - \(D_{add, op1}\)}
    \end{subfigure}
    \hfill
    \begin{subfigure}[b]{0.31\textwidth}
        \centering
        \includegraphics[width=\textwidth]{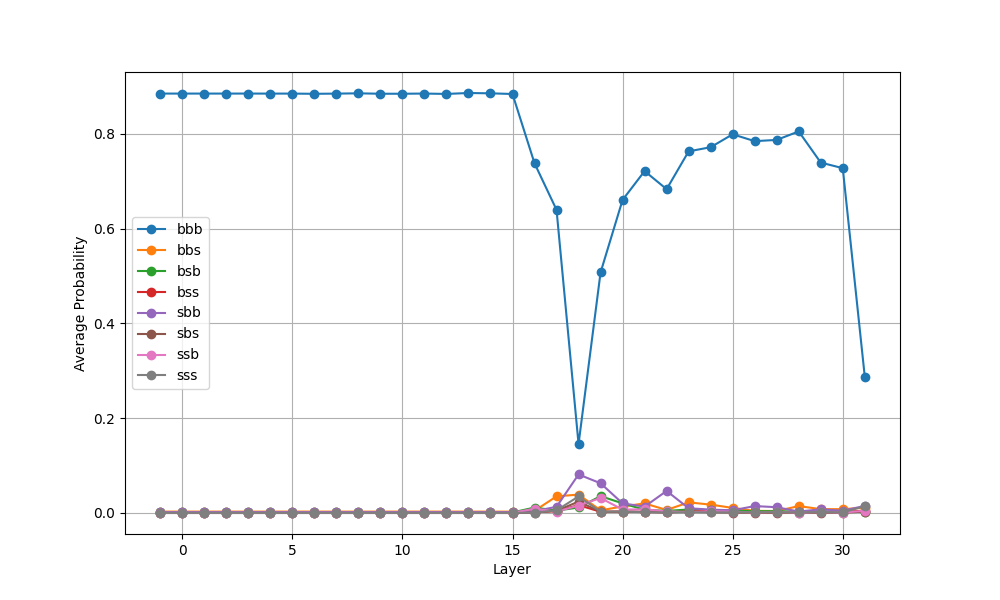} 
        \caption{MLP Output - \(D_{add, op1}\)}
    \end{subfigure}
    \vspace{1em} 
        \begin{subfigure}[b]{0.31\textwidth}
        \centering
        \includegraphics[width=\textwidth]{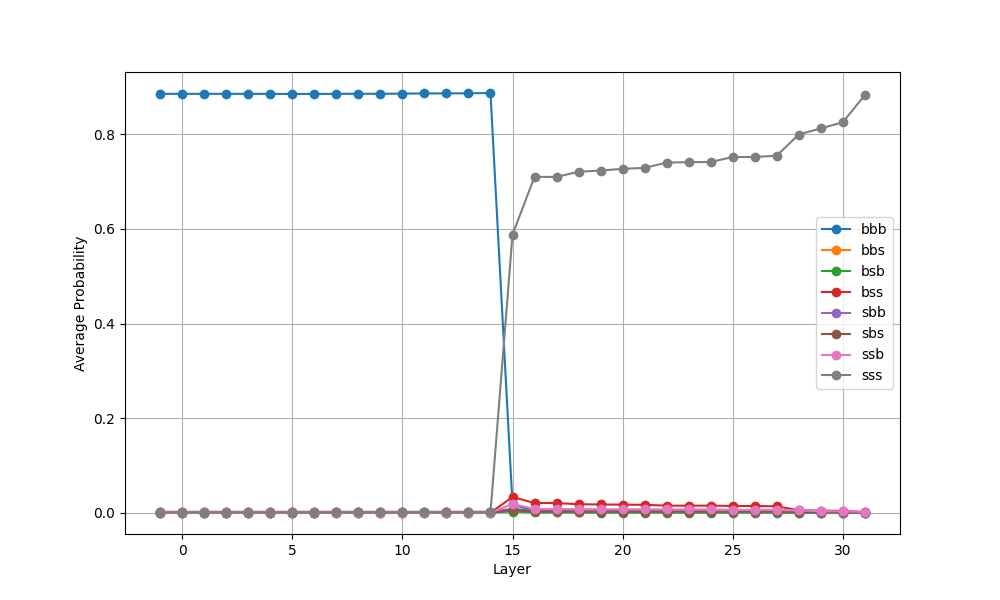}
        \caption{Block Output - \(D_{add, op2}\)}
    \end{subfigure} 
    \hfill
    \begin{subfigure}[b]{0.31\textwidth}
        \centering
        \includegraphics[width=\textwidth]{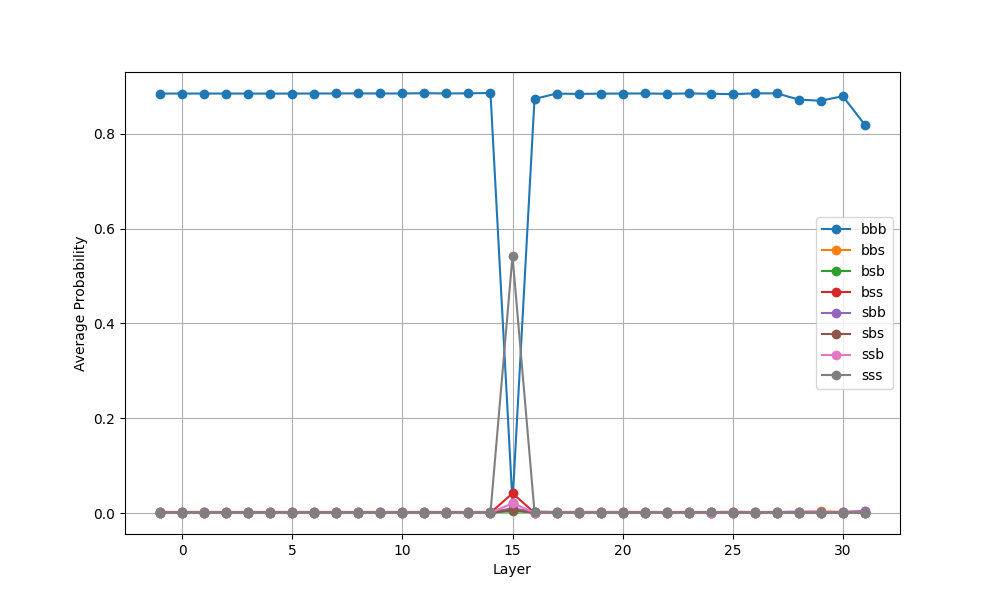}
        \caption{Attention Output - \(D_{add, op2}\)}
    \end{subfigure}
    \hfill
    \begin{subfigure}[b]{0.31\textwidth}
        \centering
        \includegraphics[width=\textwidth]{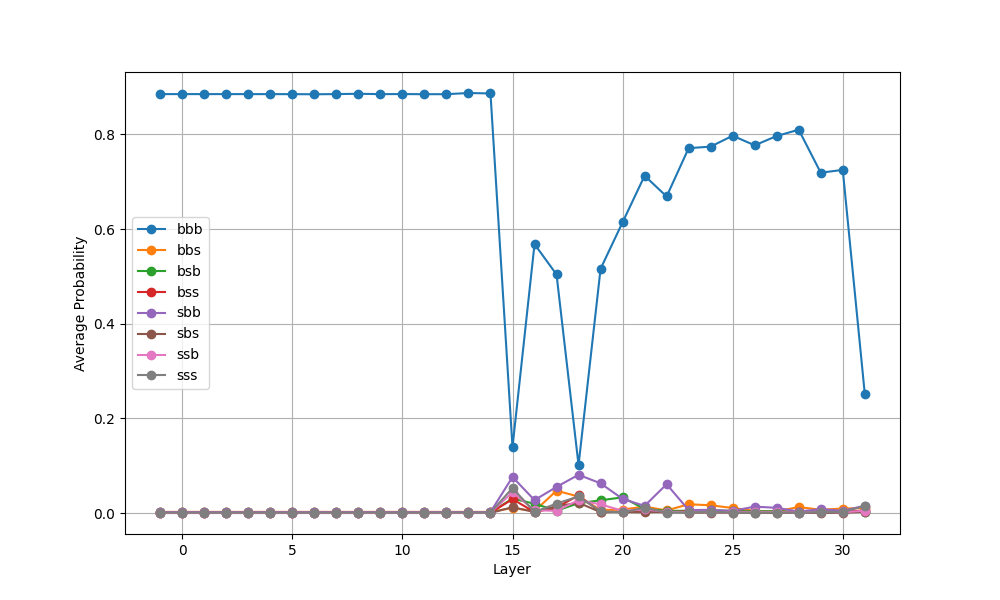} 
        \caption{MLP Output - \(D_{add, op2}\)}
    \end{subfigure}
    \caption{\textbf{LLaMA3-8B}: Probability of result variants after intervention at individual modules of individual layers on Addition Datasets. We see the Operand Injection into the residual stream in Layer 16 for \(D_{add, op1}\), and in Layer 15 for \(D_{add, op2}\).}
    \label{fig:SingleLayers}
\end{figure*}

\begin{figure*}[ht]
    \centering
    \begin{subfigure}[b]{0.35\textwidth}
        \centering
        \includegraphics[width=\textwidth]{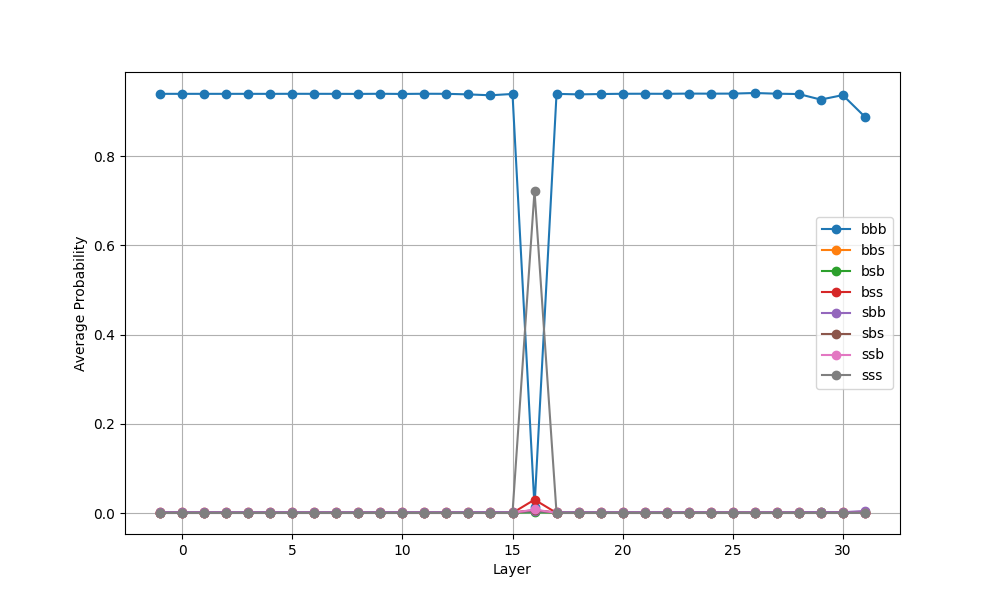}
        \caption{\(D_{sub, op1}\): Operand Injection into the residual stream in Layer 16}
    \end{subfigure}
    \begin{subfigure}[b]{0.35\textwidth}
        \centering
        \includegraphics[width=\textwidth]{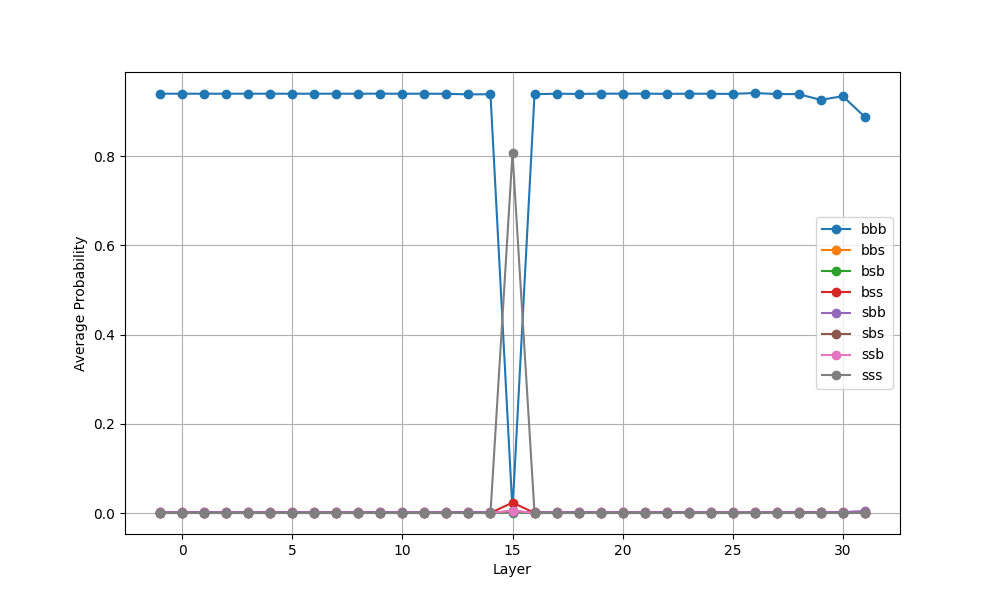}
        \caption{\(D_{sub, op2}\): Operand Injection into the residual stream in Layer 15}
    \end{subfigure}
    \caption{\textbf{LLaMA3-8B}: Probability of result variants after intervention at attention modules of individual layers on Subtraction Datasets.}
    \label{fig:AttLlama8BSub}
\end{figure*}

\begin{figure*}[ht]
    \centering
    \begin{subfigure}[b]{0.35\textwidth}
        \centering
        \includegraphics[width=\textwidth]{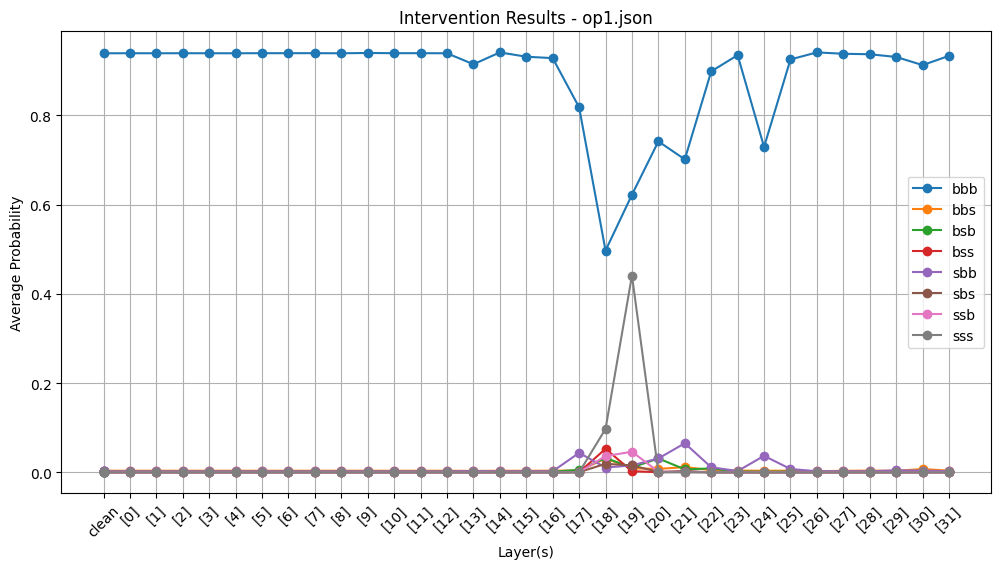}
        \caption{\(D_{add, op1}\): Operand Injection into the residual stream starting in Layer 17}
    \end{subfigure}
    \begin{subfigure}[b]{0.35\textwidth}
        \centering
        \includegraphics[width=\textwidth]{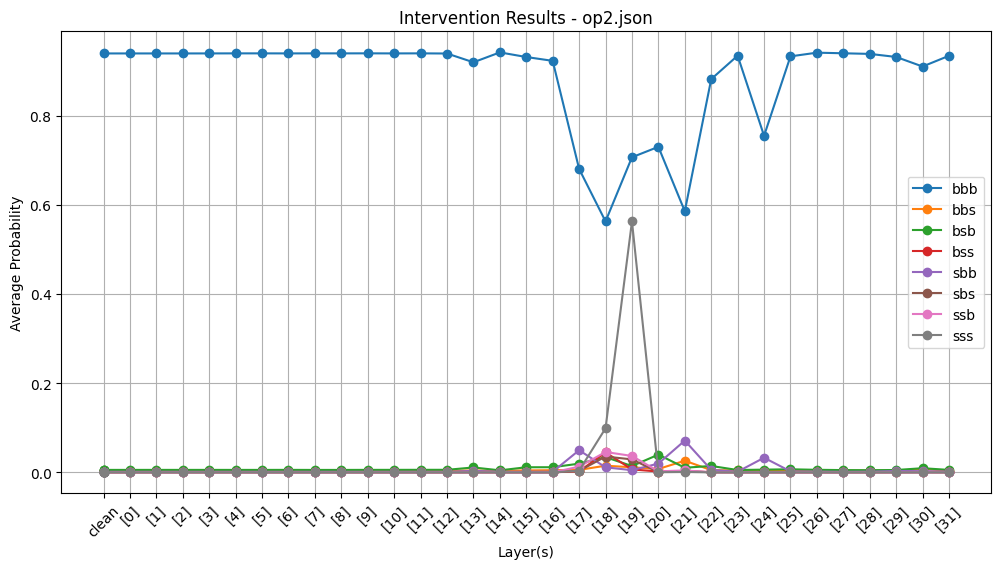}
        \caption{\(D_{add, op2}\): Operand Injection into the residual stream starting in Layer 17}
    \end{subfigure}
    \caption{\textbf{Olmo 2 7B}: Probability of result variants after intervention at attention modules of individual layers on Addition Datasets.}
    \label{fig:AttOlmAdd}
\end{figure*}

\begin{figure*}[ht]
    \centering
    \begin{subfigure}[b]{0.35\textwidth}
        \centering
        \includegraphics[width=\textwidth]{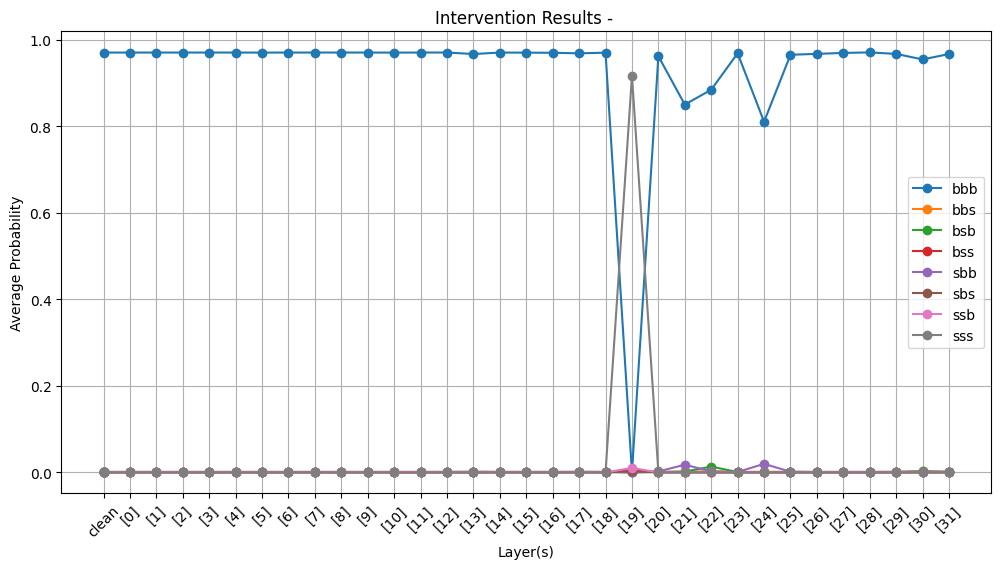}
        \caption{\(D_{sub, op1}\): Operand Injection into the residual stream in Layer 19}
    \end{subfigure}
    \begin{subfigure}[b]{0.35\textwidth}
        \centering
        \includegraphics[width=\textwidth]{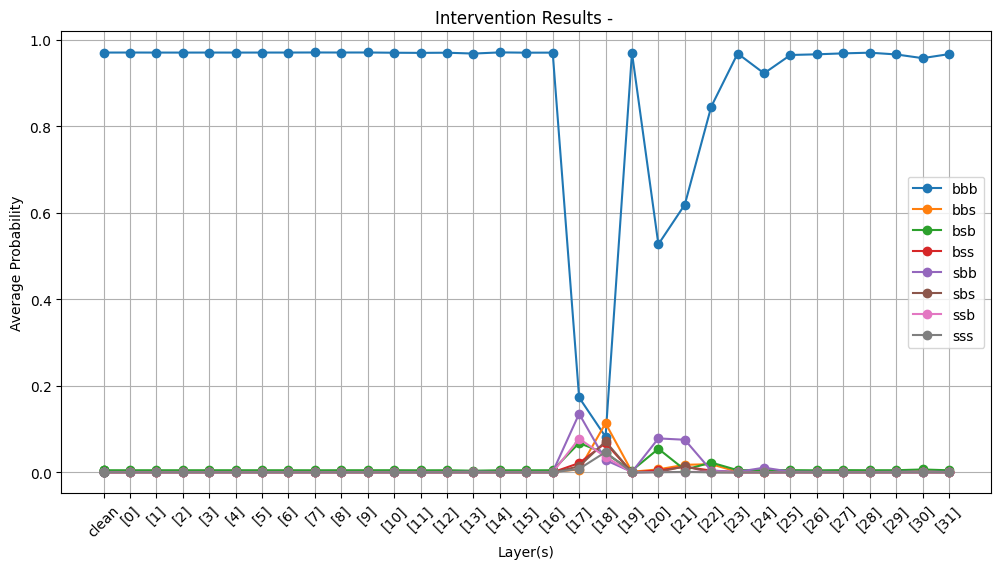}
        \caption{\(D_{sub, op2}\): Operand Injection into the residual stream starting in Layer 17}
    \end{subfigure}
    \caption{\textbf{Olmo 2 7B}: Probability of result variants after intervention at attention modules of individual layers on Subtraction Datasets.}
    \label{fig:AttOlmSub}
\end{figure*}

\begin{figure*}[ht]
    \centering
    \begin{subfigure}[b]{0.35\textwidth}
        \centering
        \includegraphics[width=\textwidth]{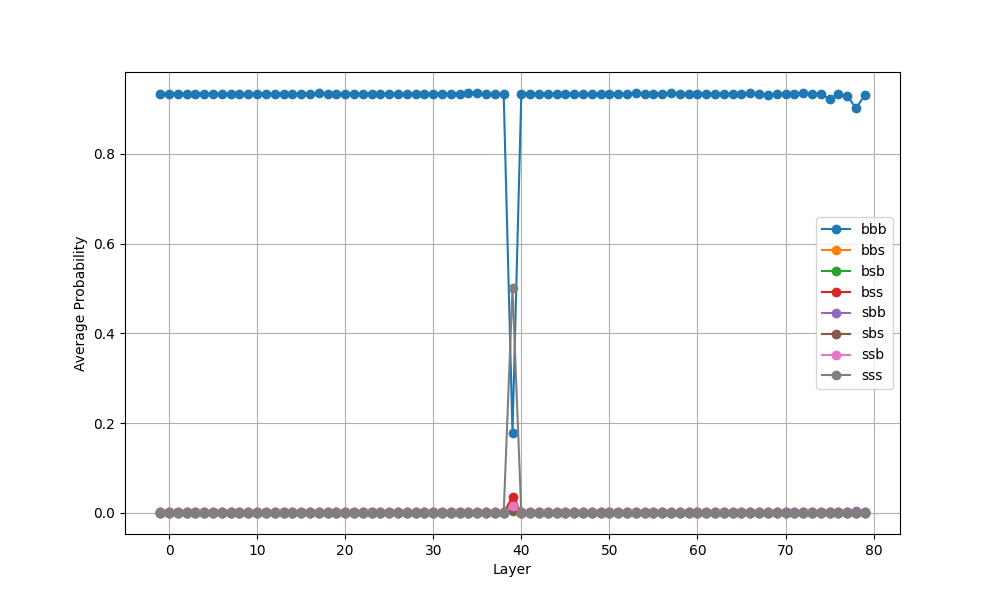}
        \caption{\(D_{add, op1}\): Operand Injection into the residual stream in Layer 39}
    \end{subfigure}
    \begin{subfigure}[b]{0.35\textwidth}
        \centering
        \includegraphics[width=\textwidth]{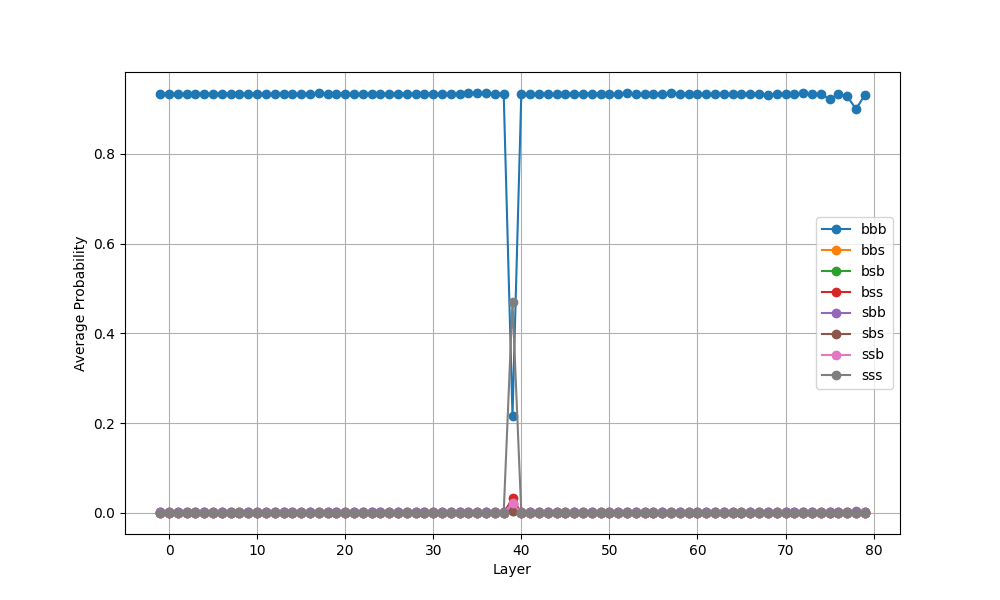}
        \caption{\(D_{add, op2}\): Operand Injection into the residual stream in Layer 39}
    \end{subfigure}
    \caption{\textbf{LLaMA3-70B}: Probability of result variants after intervention at attention modules of individual layers on Addition Datasets.}
    \label{fig:Att70BAdd}
\end{figure*}

\begin{figure*}[ht]
    \centering
    \begin{subfigure}[b]{0.35\textwidth}
        \centering
        \includegraphics[width=\textwidth]{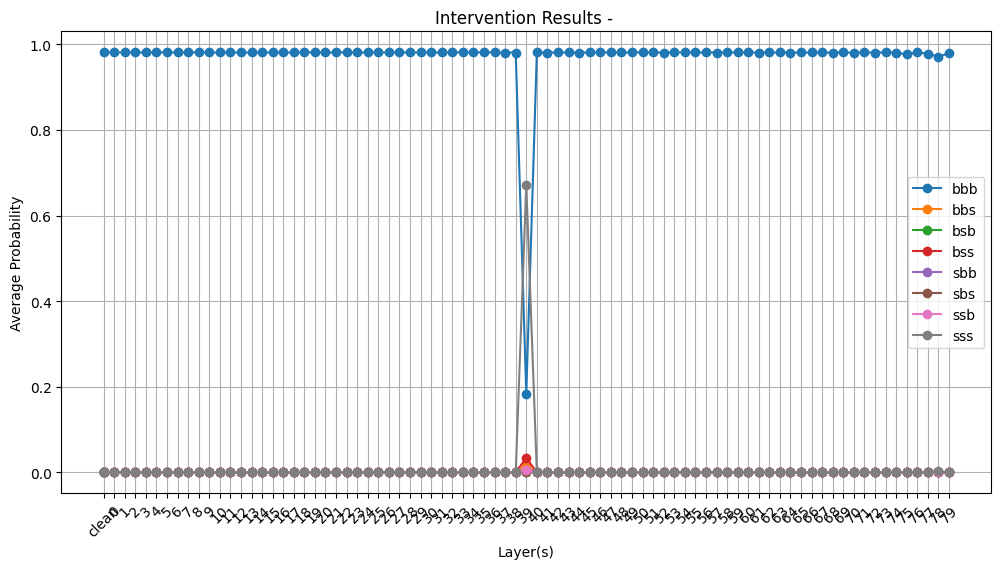}
        \caption{\(D_{sub, op1}\): Operand Injection into the residual stream in Layer 39}
    \end{subfigure}
    \begin{subfigure}[b]{0.35\textwidth}
        \centering
        \includegraphics[width=\textwidth]{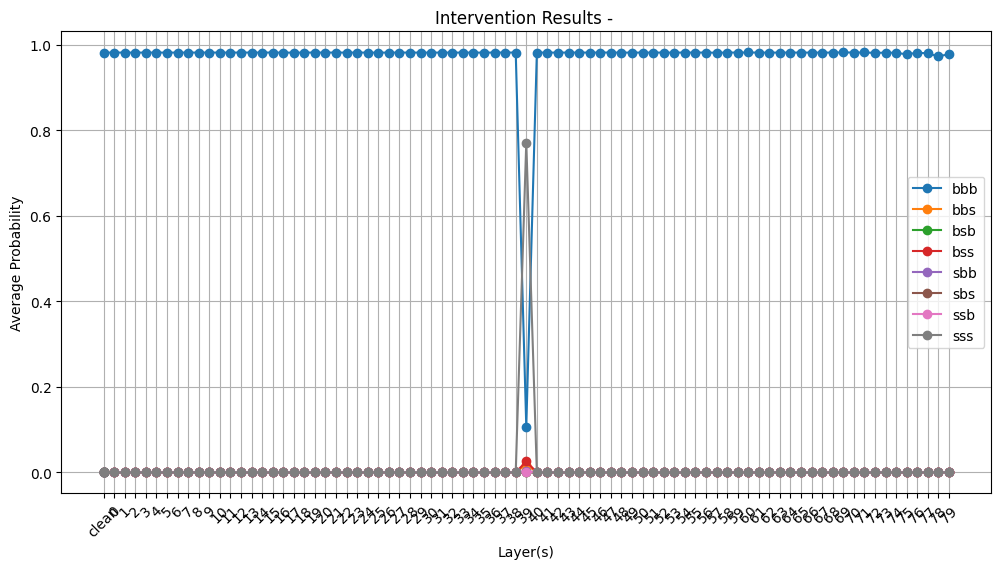}
        \caption{\(D_{sub, op2}\): Operand Injection into the residual stream in Layer 39}
    \end{subfigure}
    \caption{\textbf{LLaMA3-70B}: Probability of result variants after intervention at attention modules of individual layers on Subtraction Datasets.}
    \label{fig:Att70BSub}
\end{figure*}

\begin{figure*}[ht]
    \centering
    \begin{subfigure}[b]{0.35\textwidth}
        \centering
        \includegraphics[width=\textwidth]{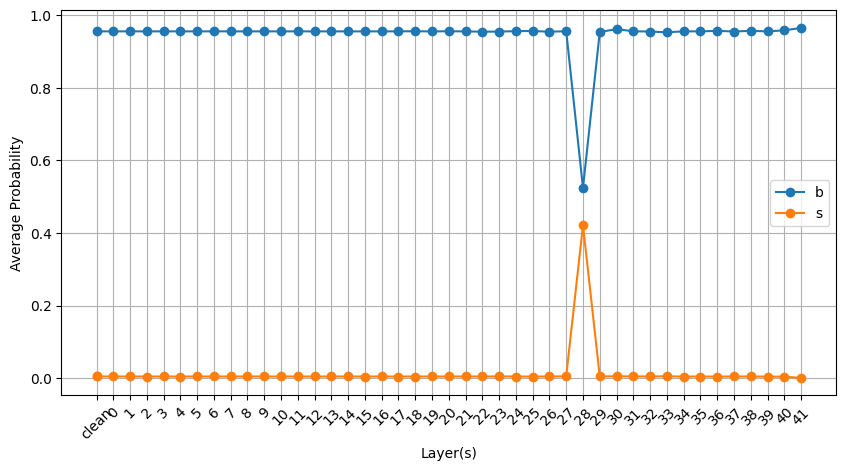}
        \caption{\(D_{add, op1}\): Operand Injection into the residual stream in Layer 28}
    \end{subfigure}
    \begin{subfigure}[b]{0.35\textwidth}
        \centering
        \includegraphics[width=\textwidth]{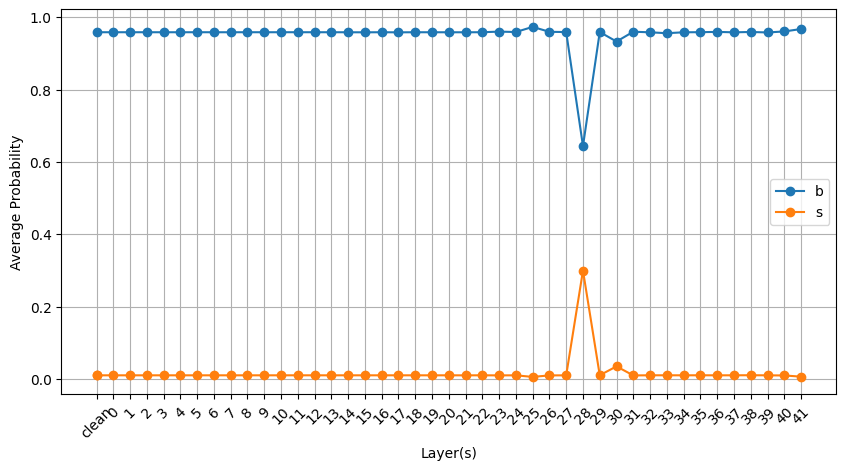}
        \caption{\(D_{add, op2}\): Operand Injection into the residual stream in Layer 28}
    \end{subfigure}
    \caption{\textbf{Gemma 2 9B}: Probability of result variants after intervention at attention modules of individual layers on Addition Datasets.}
    \label{fig:AttGemma9BAdd}
\end{figure*}

\begin{figure*}[ht]
    \centering
    \begin{subfigure}[b]{0.35\textwidth}
        \centering
        \includegraphics[width=\textwidth]{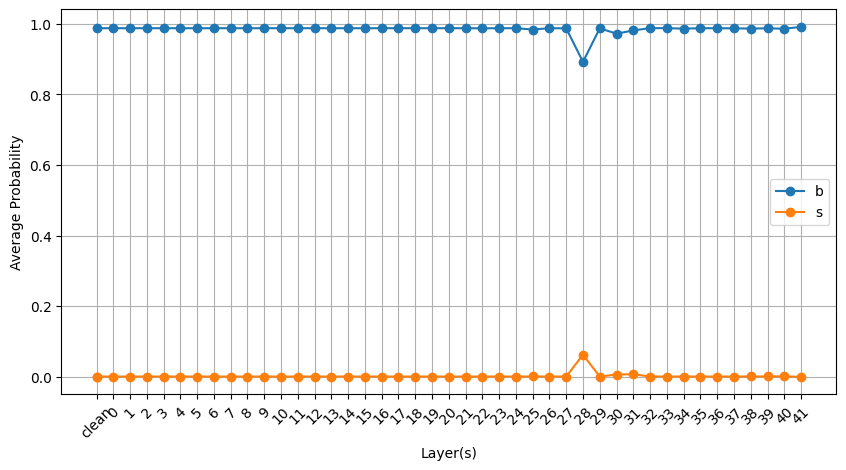}
        \caption{\(D_{sub, op1}\): Minimal Effect of Operand Injection into the residual stream in Layer 28}
    \end{subfigure}
    \begin{subfigure}[b]{0.35\textwidth}
        \centering
        \includegraphics[width=\textwidth]{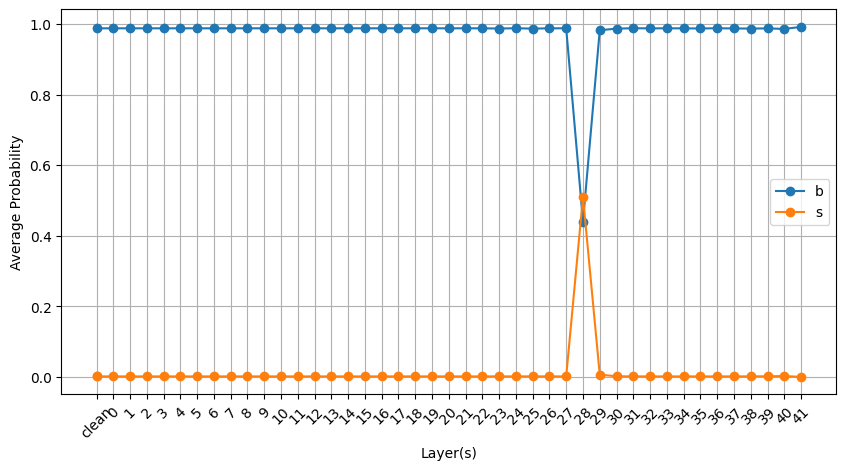}
        \caption{\(D_{sub, op2}\): Operand Injection into the residual stream in Layer 28}
    \end{subfigure}
    \caption{\textbf{Gemma 2 9B}: Probability of result variants after intervention at attention modules of individual layers on Subtraction Datasets.}
    \label{fig:AttGemma9BSub}
\end{figure*}

Figures \ref{fig:SingleLayers} to \ref{fig:AttGemma9BSub}, reveal that intervention on specific attention modules dramatically increases the probability of \texttt{sss} in all models. This suggests that these attention modules are responsible for injecting operand information into the residual stream at the final token position, which is consistent with prior work \cite{stolfo_mechanistic_2023}.

We thus look for digit-wise addition neurons in the MLP layers following the operand injection into the residual stream. For Llama 3 8B for instance, this happens at layer 16 for \(D_{add, op1}\) and layer 15 for \(D_{add, op2}\) respectively (Figure~\ref{fig:SingleLayers}).

\clearpage
\section{Fisher Score for Digit-Sensitivity of Neurons.}
\label{app:FisherFormal}

We measure how sensitive individual MLP neurons are to digit-level structure in arithmetic prompts, using a Fisher Score that quantifies class discriminability.

Let:

\begin{itemize} [itemsep=2pt, topsep=2pt]
  \item \( x \in \mathcal{X} \): an input prompt (e.g., \texttt{"157 + 431 = "}),
  \item \( i \): index of a neuron in an MLP layer,
  \item \( d \in \{\text{hundred}, \text{ten}, \text{unit}\} \): a fixed digit position in the operands (e.g., the "tens" digit),
  \item \( c \in \{00, 01, \dots, 99\} \): class label for the digit pair at position \( d \), formed by concatenating\footnote{We choose to concatenate digit pairs, as a way to express the arithmetic digit level subtasks as a class label} the digit from each operand (e.g., \( c = 71 \) for \( d = unit\) and input prompt \(34\textbf{7} + 23\textbf{1} =\)),
  \item \( \mathcal{X}_{c,d} \subseteq \mathcal{X} \):  input prompts with digit pair \( c \) at position \( d \) (e.g., \( \mathcal{X}_{71,unit} = \{ 34\textbf{7} + 23\textbf{1} =,  21\textbf{7} + 65\textbf{1} =, ...\}\)),
  \item \( a_i(x) \in \mathbb{R} \): the activation of neuron \( i \) when processing input \( x \).
\end{itemize}

\vspace{1em}
Define:

\begin{itemize} [itemsep=2pt, topsep=2pt]
  \item \textbf{Mean activation for class \( c \) at position \( d \):}
  \[
  \mu_{i,c,d} = \frac{1}{|\mathcal{X}_{c,d}|} \sum_{x \in \mathcal{X}_{c,d}} a_i(x)
  \]
  \item \textbf{Variance of activations for class \( c \) at position \( d \):}
  \[
  \sigma^2_{i,c,d} = \frac{1}{|\mathcal{X}_{c,d}|} \sum_{x \in \mathcal{X}_{c,d}} \left(a_i(x) - \mu_{i,c,d} \right)^2
  \]
  \item \textbf{Overall mean activation across all classes at position \( d \):}
  \[
  \mu_{i,d} = \frac{1}{\sum_c |\mathcal{X}_{c,d}|} \sum_c |\mathcal{X}_{c,d}| \mu_{i,c,d}
  \]
\end{itemize}

Then, the \textbf{Fisher Score} for neuron \( i \) with respect to digit position \( d \) is:

\[
F_{i,d} = \frac{\sum_{c} |\mathcal{X}_{c,d}| \left(\mu_{i,c,d} - \mu_{i,d}\right)^2}{\sum_{c} |\mathcal{X}_{c,d}| \sigma^2_{i,c,d}}
\]

\vspace{1em}
\noindent\textbf{Intuition.} The Fisher Score quantifies how much neuron \( i \)'s activation varies \emph{between} digit-pair classes (numerator), relative to how much it varies \emph{within} each class (denominator).
A high \( F_{i,d} \) indicates that neuron \( i \)'s activation reliably distinguishes between different digit values at position \( d \), while remaining relatively insensitive to other input variations. This implies that neuron \( i \) likely participates in a digit-position-specific addition subcircuit.

\newpage
\clearpage
\section{Statistics on Arithmetic Circuits}
\label{app:circuit_statistics}

\begin{figure*}[t!]
    \centering

    \begin{subfigure}[t]{\textwidth}
        \centering
        \begin{subfigure}[b]{0.32\textwidth}
            \centering
            \includegraphics[width=\textwidth]{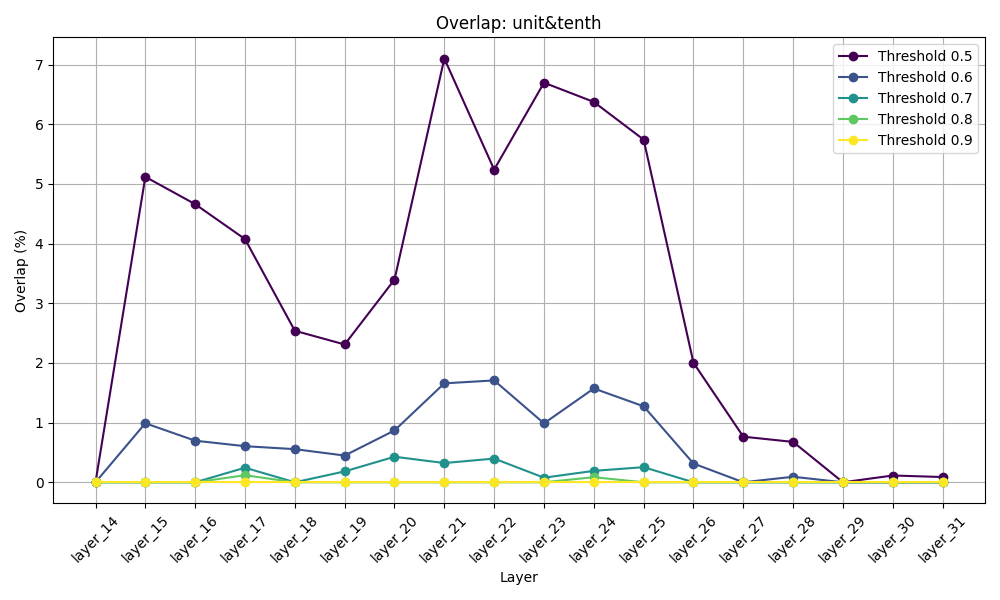}
            \caption{Unit and Tens}
            \label{fig:sub_overlap_unit_tenth_add_llama8}
        \end{subfigure}
        \hfill
        \begin{subfigure}[b]{0.32\textwidth}
            \centering
            \includegraphics[width=\textwidth]{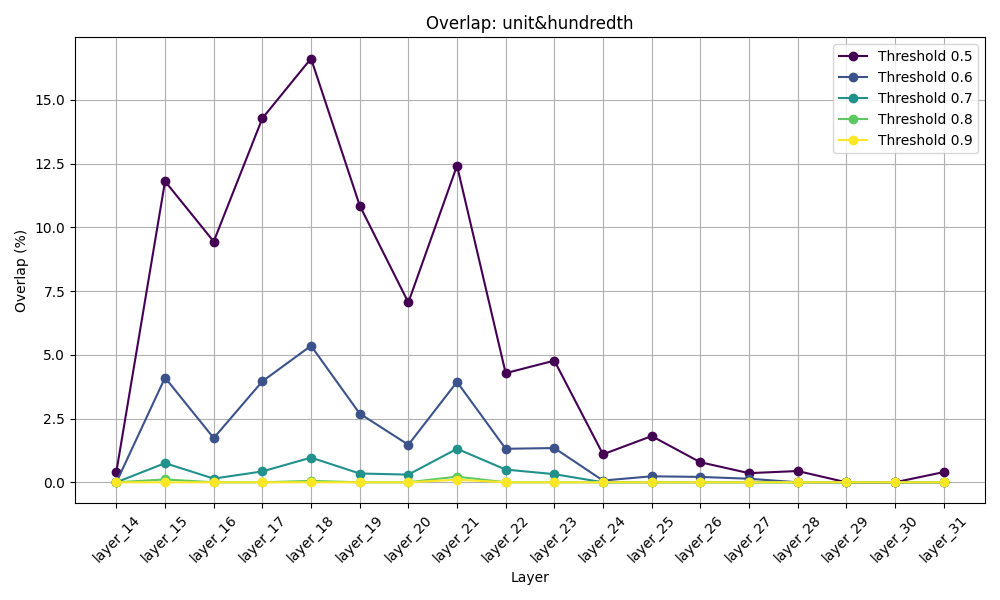}
            \caption{Unit and Hundreds}
            \label{fig:sub_overlap_unit_hundredth_add_llama8}
        \end{subfigure}
        \hfill
        \begin{subfigure}[b]{0.32\textwidth}
            \centering
            \includegraphics[width=\textwidth]{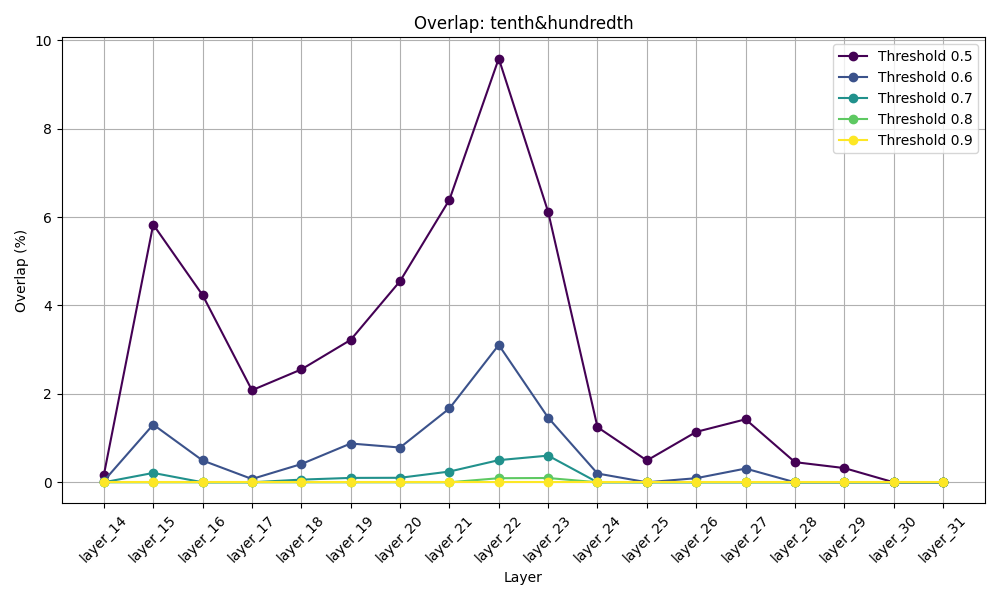}
            \caption{Tens and Hundreds}
            \label{fig:sub_overlap_tenth_hundredth_add_llama8}
        \end{subfigure}
        \caption{\textit{Circuit Overlap}: Overlap in neurons (\%) between digit-position circuits.}
        \label{fig:sub_row_overlap_add_llama8}
    \end{subfigure}

    \vspace{1em}

    \begin{subfigure}[t]{\textwidth}
        \centering
        \begin{subfigure}[b]{0.32\textwidth}
            \centering
            \includegraphics[width=\textwidth]{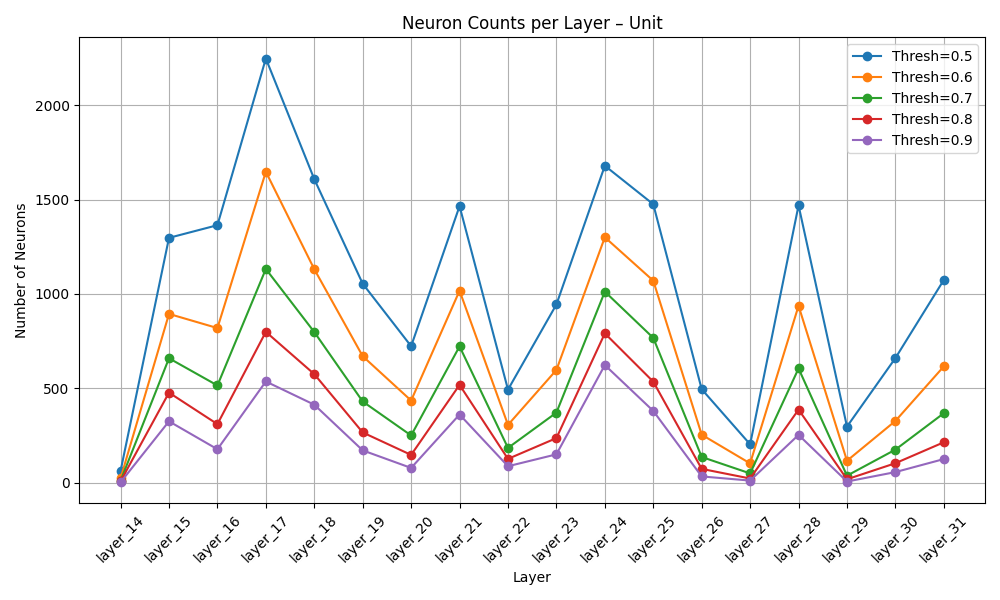}
            \caption{Unit}
            \label{fig:sub_counts_unit_add_llama8}
        \end{subfigure}
        \hfill
        \begin{subfigure}[b]{0.32\textwidth}
            \centering
            \includegraphics[width=\textwidth]{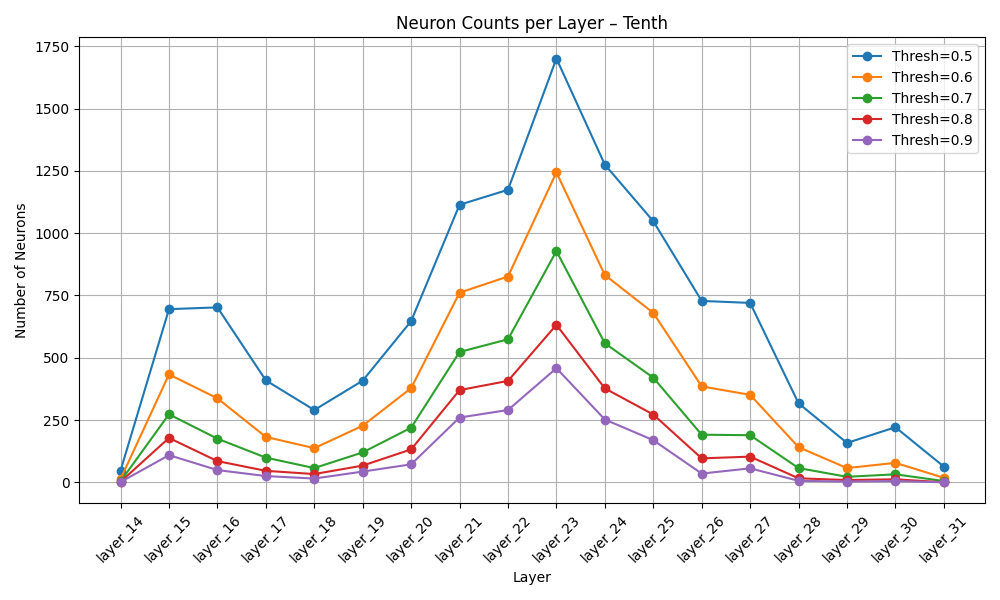}
            \caption{Tens}
            \label{fig:sub_counts_tenth_add_llama8}
        \end{subfigure}
        \hfill
        \begin{subfigure}[b]{0.32\textwidth}
            \centering
            \includegraphics[width=\textwidth]{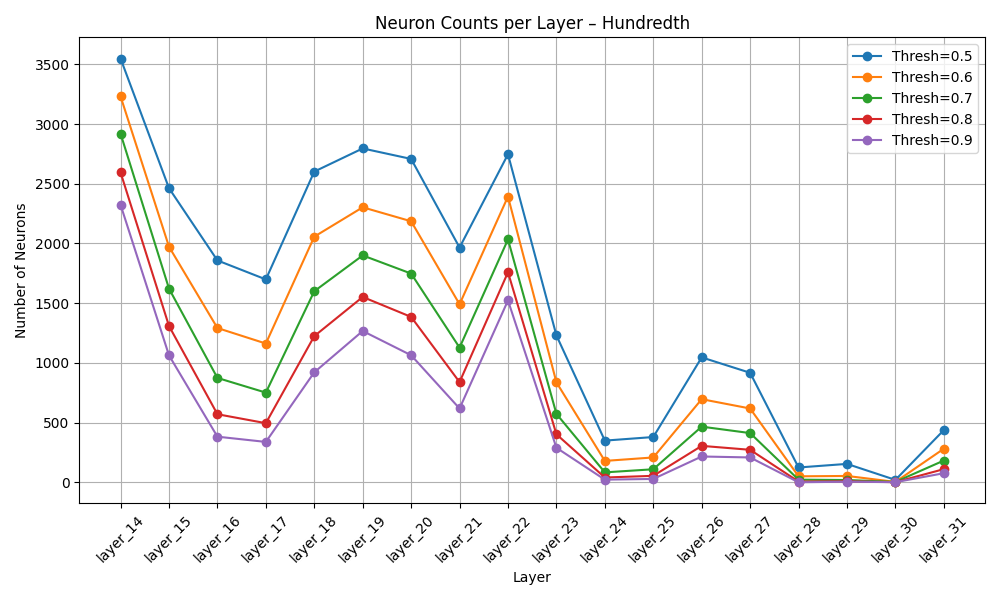}
            \caption{Hundreds}
            \label{fig:sub_counts_hundredth_add_llama8}
        \end{subfigure}
        \caption{\textit{Circuit Size}: Number of neurons per layer in digit-position circuits (Hidden size = 4096).}
        \label{fig:sub_row_counts_add_llama8}
    \end{subfigure}

    \vspace{1em}

    \begin{subfigure}[t]{\textwidth}
        \centering
        \begin{subfigure}[b]{0.32\textwidth}
            \centering
            \includegraphics[width=\textwidth]{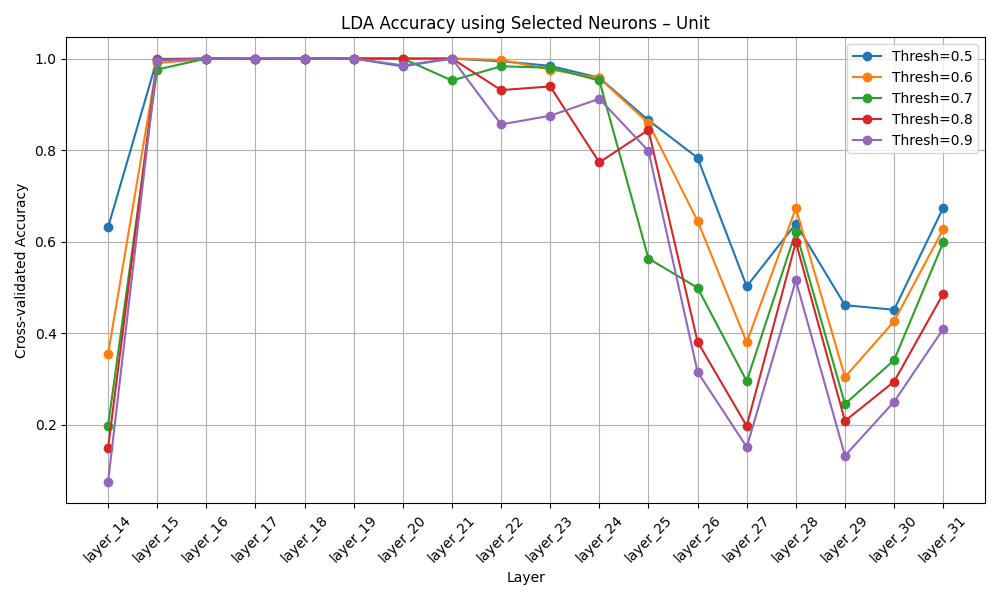}
            \caption{Unit}
            \label{fig:sub_suff_unit_add_llama8}
        \end{subfigure}
        \hfill
        \begin{subfigure}[b]{0.32\textwidth}
            \centering
            \includegraphics[width=\textwidth]{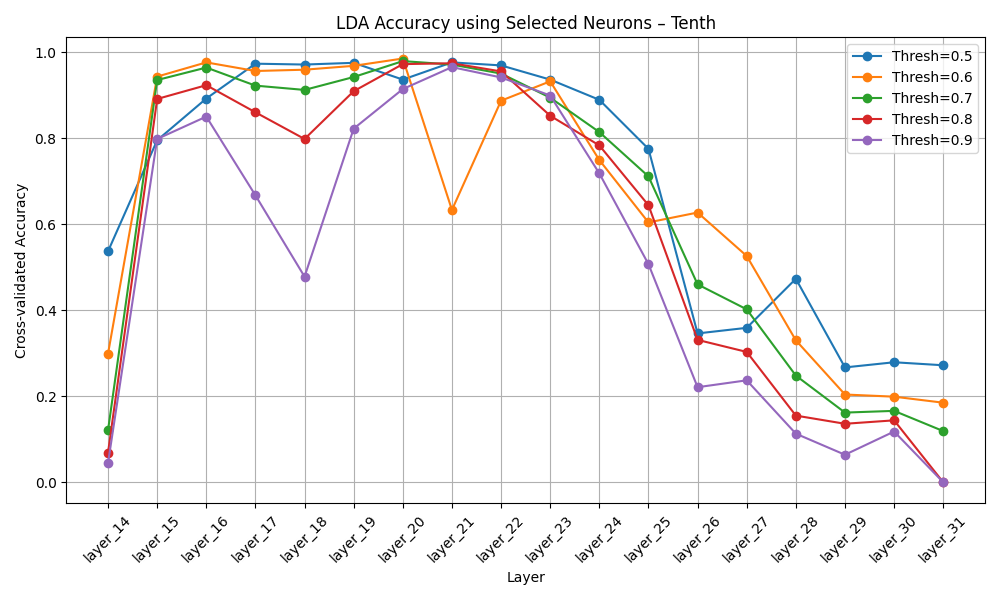}
            \caption{Tens}
            \label{fig:sub_suff_tenth_add_llama8}
        \end{subfigure}
        \hfill
        \begin{subfigure}[b]{0.32\textwidth}
            \centering
            \includegraphics[width=\textwidth]{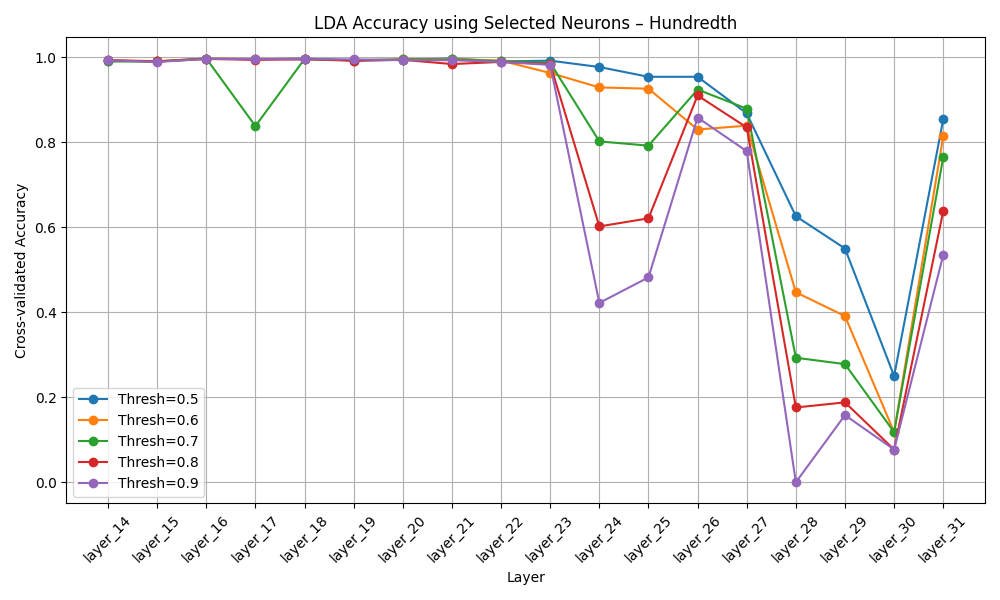}
            \caption{Hundreds}
            \label{fig:sub_suff_hundredth_add_llama8}
        \end{subfigure}
        \caption{\textit{Circuit Sufficiency}: Sufficiency of digit-position circuit vs. full LDA}
        \label{fig:sub_row_sufficiency_add_llama8}
    \end{subfigure}

    \caption{Llama 3 8B, \(D_{add}\): Circuit statistics across digit positions and thresholds.}
    \label{fig:llama8_circuit_stats}
\end{figure*}

\textbf{\textit{Circuit Size.}}
We find that digit-specific circuits are relatively 'wide' (Figure \ref{fig:llama8_circuit_stats}). For the respective best thresholds \(t*\)\footnote{See Section \ref{sec:intervention}: 0.5 for unit, 0.6 for tens, 0.9 for hundreds}, the average number of MLP neurons per layer responsible for one of the digit-position specific circuits is 60.3\% of all MLP neurons. 
In other words, almost two-thirds of the MLP neurons in relevant layers are digit-position specific, when Llama 3 8B solves addition tasks.

\textbf{\textit{Circuit Overlap.}}
If arithmetic tasks are indeed solved in a digit-wise manner, then neuron groups responsible for generating different digit position results should be distinct. To assess the distinctiveness of different digit-position circuits, we examine the overlap of member neurons across digit position circuits. 
We find that neuron sets for different digit positions are largely distinct (Figure \ref{fig:llama8_circuit_stats}).
In fact, for higher thresholds (>= 0.7) neural overlap between circuits is largely negligible (<2\%). 
The distinctiveness of digit-position specific arithmetic circuits provides a first indication of digit-positional modularity in LLM arithmetic.

\textbf{\textit{Circuit Sufficiency.}}
To evaluate whether Fisher-identified neuron groups are sufficient to represent digit-specific arithmetic subtasks, we perform a Linear Discriminant Analysis (LDA)-based classification test. For each layer \(l\) and digit position \(d\), we train two LDA classifiers to predict the digit-pair label \(c\): one using the full MLP activation vector, and one using a reduced representation containing only neurons above a Fisher importance threshold.

Figure~\ref{fig:llama8_circuit_stats} compares classification accuracy between the full and reduced settings. We find that performance remains high in the reduced setting—especially in middle layers (15–24)—indicating that the selected neurons are sufficient to support digit-position classification.

Sufficiency is strongest for the unit and hundreds digits, and at lower thresholds, also for the tens digit. Stable accuracy across thresholds suggests that unit and hundreds information is more redundantly or compactly encoded, while the tens digit may require more neurons. A drop in later layers suggests that digit-specific information becomes less localized in deeper MLP layers.

\begin{figure*}[t!]
    \centering

    \begin{subfigure}[t]{\textwidth}
        \centering
        \begin{subfigure}[b]{0.32\textwidth}
            \centering
            \includegraphics[width=\textwidth]{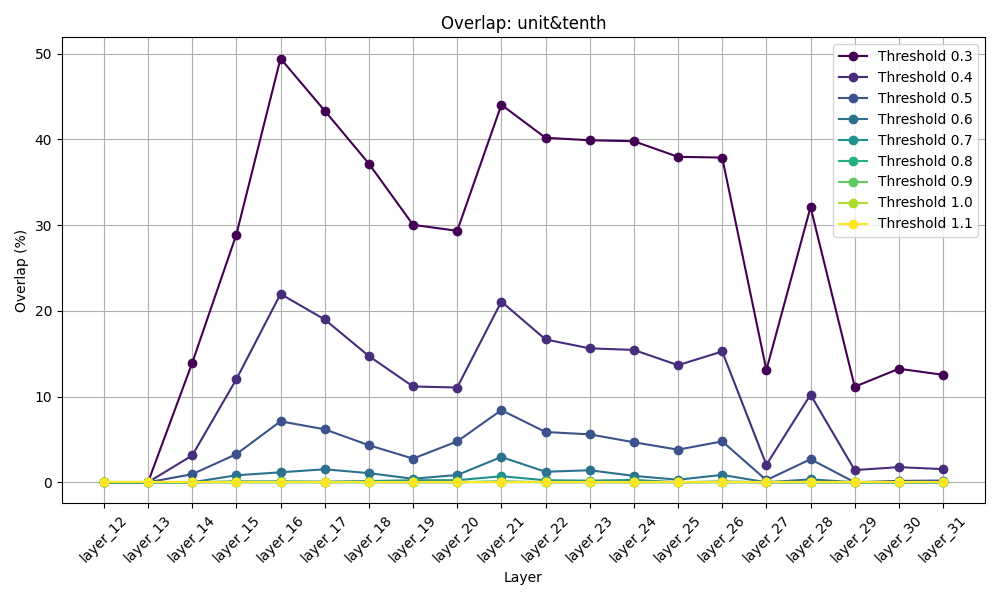}
            \caption{Unit and Tens}
            \label{fig:sub_overlap_unit_tenth_sub_llama8}
        \end{subfigure}
        \hfill
        \begin{subfigure}[b]{0.32\textwidth}
            \centering
            \includegraphics[width=\textwidth]{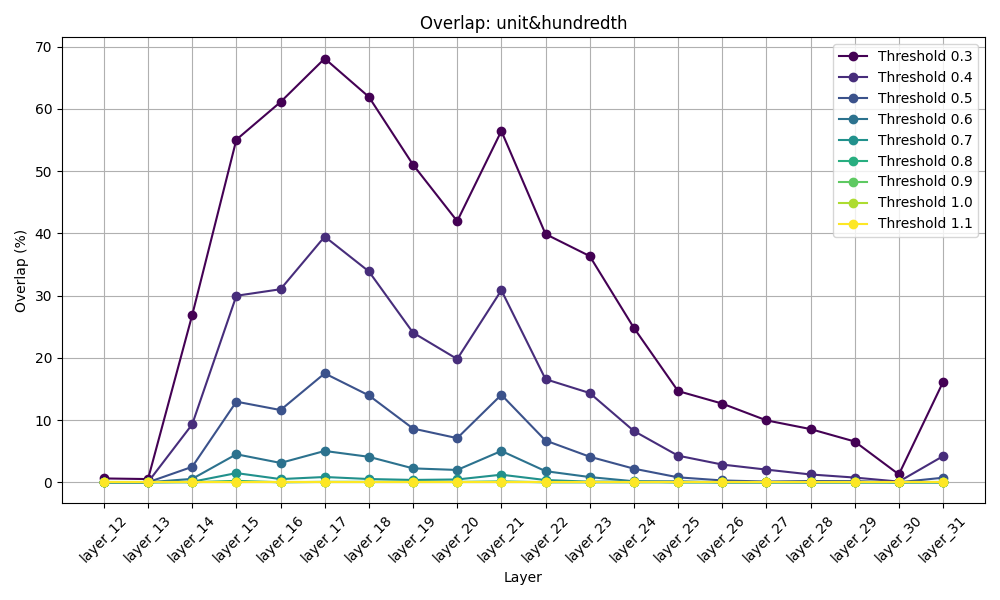}
            \caption{Unit and Hundreds}
            \label{fig:sub_overlap_unit_hundredth_sub_llama8}
        \end{subfigure}
        \hfill
        \begin{subfigure}[b]{0.32\textwidth}
            \centering
            \includegraphics[width=\textwidth]{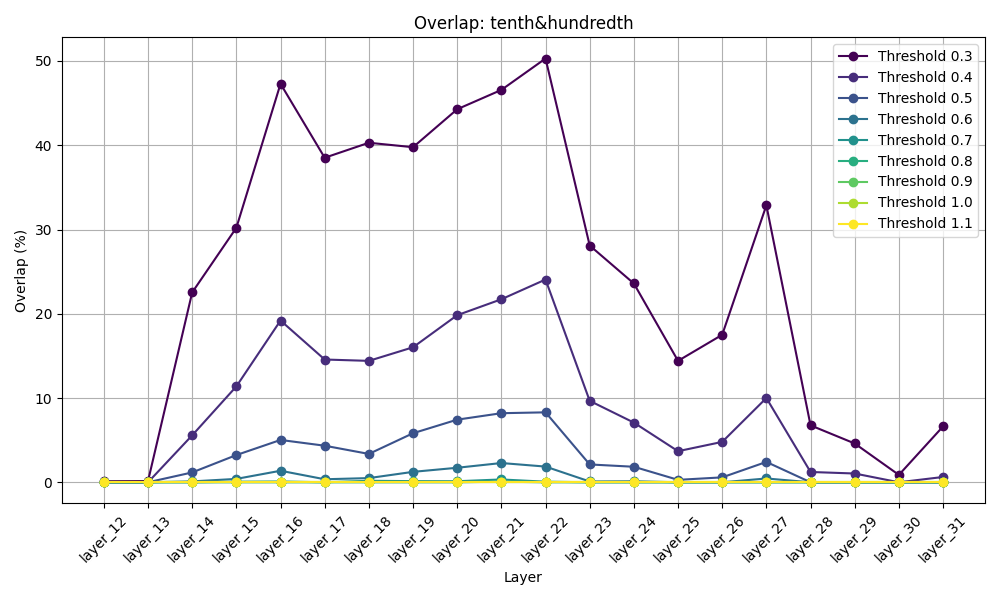}
            \caption{Tens and Hundreds}
            \label{fig:sub_overlap_tenth_hundredth_sub_llama8}
        \end{subfigure}
        \caption{\textit{Circuit Overlap}: Overlap in neurons (\%) between digit-position circuits.}
        \label{fig:sub_row_overlap_sub_llama8}
    \end{subfigure}

    \vspace{1em}

    \begin{subfigure}[t]{\textwidth}
        \centering
        \begin{subfigure}[b]{0.32\textwidth}
            \centering
            \includegraphics[width=\textwidth]{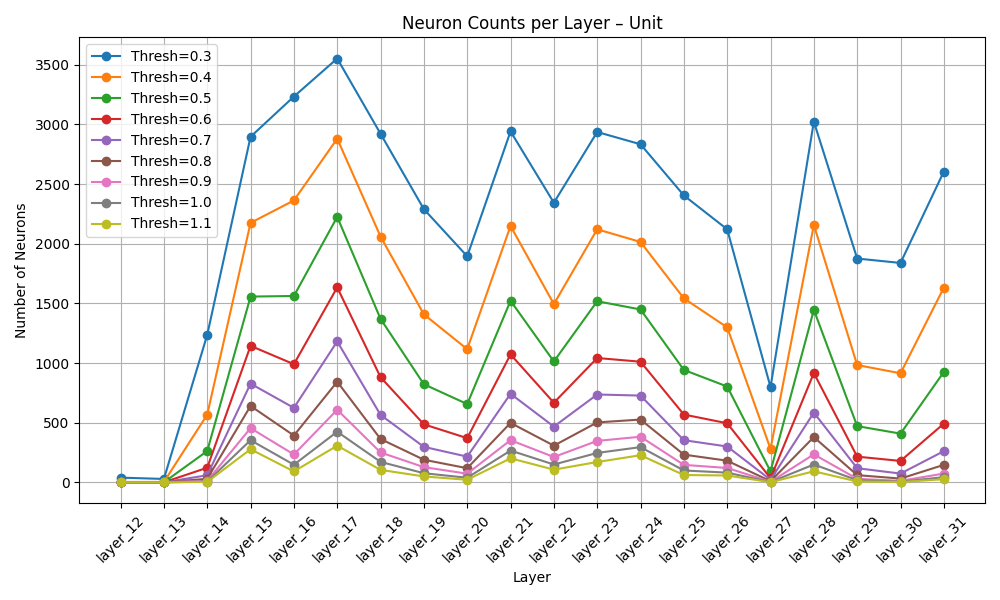}
            \caption{Unit}
            \label{fig:sub_counts_unit_sub_llama8}
        \end{subfigure}
        \hfill
        \begin{subfigure}[b]{0.32\textwidth}
            \centering
            \includegraphics[width=\textwidth]{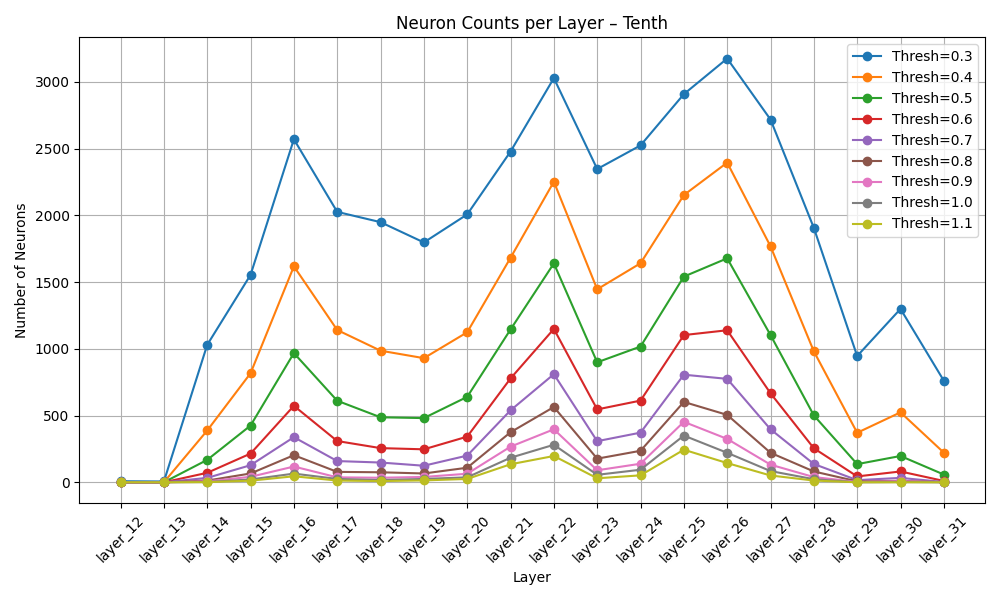}
            \caption{Tens}
            \label{fig:sub_counts_tenth_sub_llama8}
        \end{subfigure}
        \hfill
        \begin{subfigure}[b]{0.32\textwidth}
            \centering
            \includegraphics[width=\textwidth]{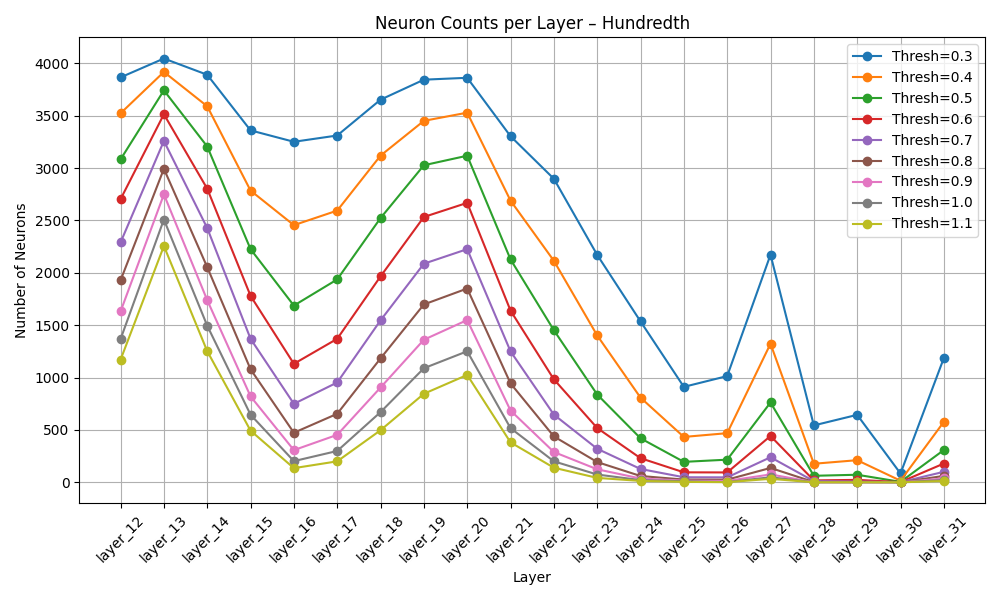}
            \caption{Hundreds}
            \label{fig:sub_counts_hundredth_sub_llama8}
        \end{subfigure}
        \caption{\textit{Circuit Size}: Number of neurons per layer in digit-position circuits (Hidden size = 4096).}
        \label{fig:sub_row_counts_sub_llama8}
    \end{subfigure}

    \vspace{1em}

    \begin{subfigure}[t]{\textwidth}
        \centering
        \begin{subfigure}[b]{0.32\textwidth}
            \centering
            \includegraphics[width=\textwidth]{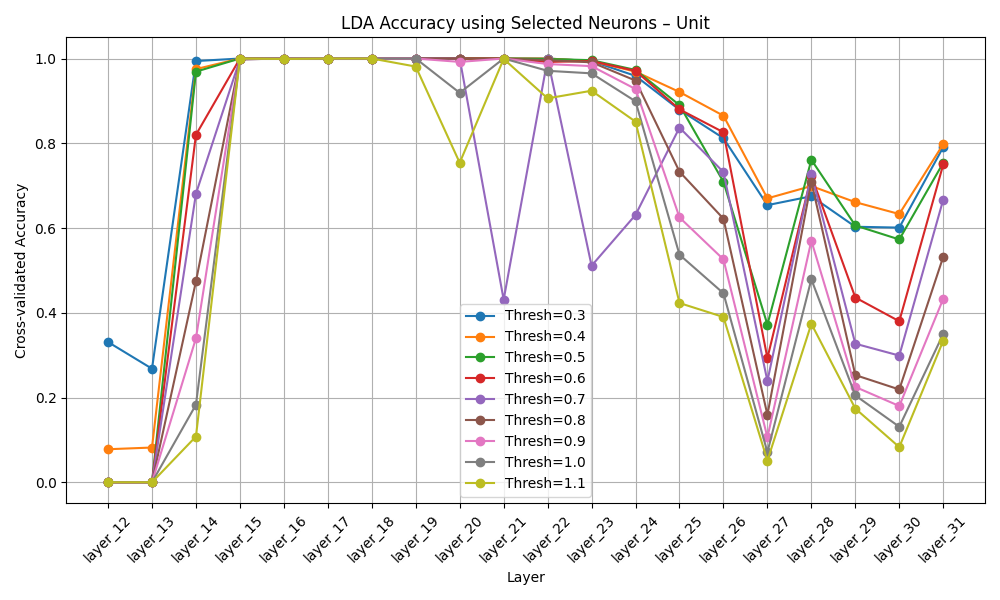}
            \caption{Unit}
            \label{fig:sub_suff_unit_sub_llama8}
        \end{subfigure}
        \hfill
        \begin{subfigure}[b]{0.32\textwidth}
            \centering
            \includegraphics[width=\textwidth]{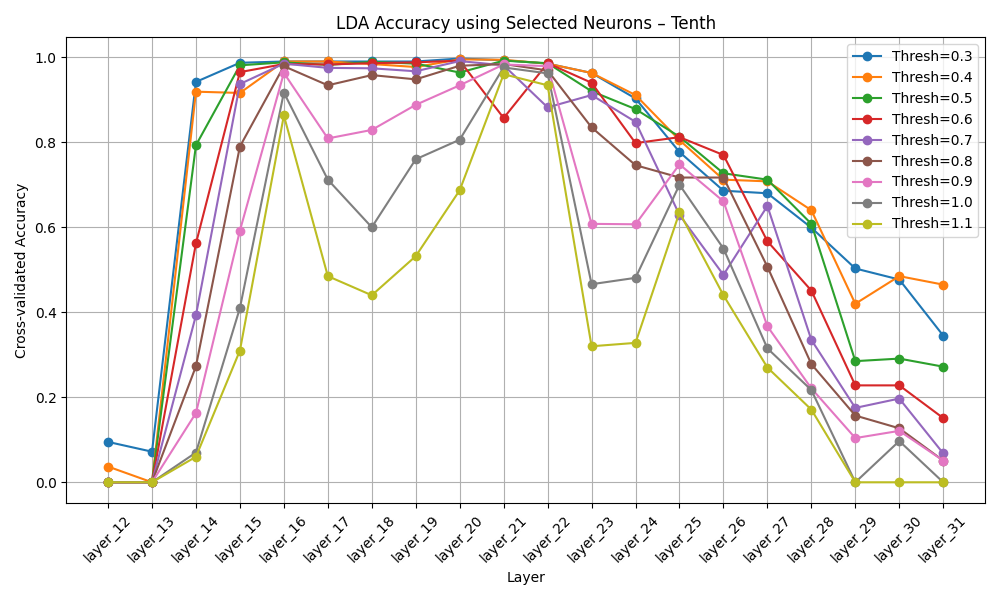}
            \caption{Tens}
            \label{fig:sub_suff_tenth_sub_llama8}
        \end{subfigure}
        \hfill
        \begin{subfigure}[b]{0.32\textwidth}
            \centering
            \includegraphics[width=\textwidth]{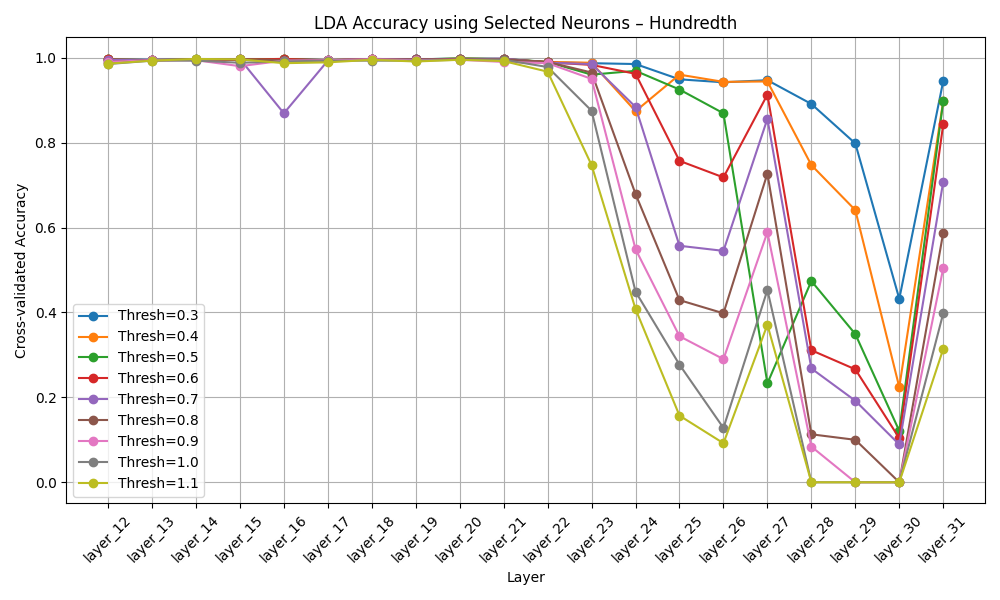}
            \caption{Hundreds}
            \label{fig:sub_suff_hundredth_sub_llama8}
        \end{subfigure}
        \caption{\textit{Circuit Sufficiency}: Sufficiency of digit-position circuit vs. full LDA}
        \label{fig:sub_row_sufficiency_sub_llama8}
    \end{subfigure}

    \caption{Llama 3 8B, \(D_{sub}\): Circuit statistics across digit positions and thresholds.}
    \label{fig:sub_combined_fisher_sub_llama8}
\end{figure*}

\begin{figure*}[h!]
    \centering
    
    \begin{subfigure}[t]{\textwidth}
        \centering
        \begin{subfigure}[b]{0.32\textwidth}
            \centering
            \includegraphics[width=\textwidth]{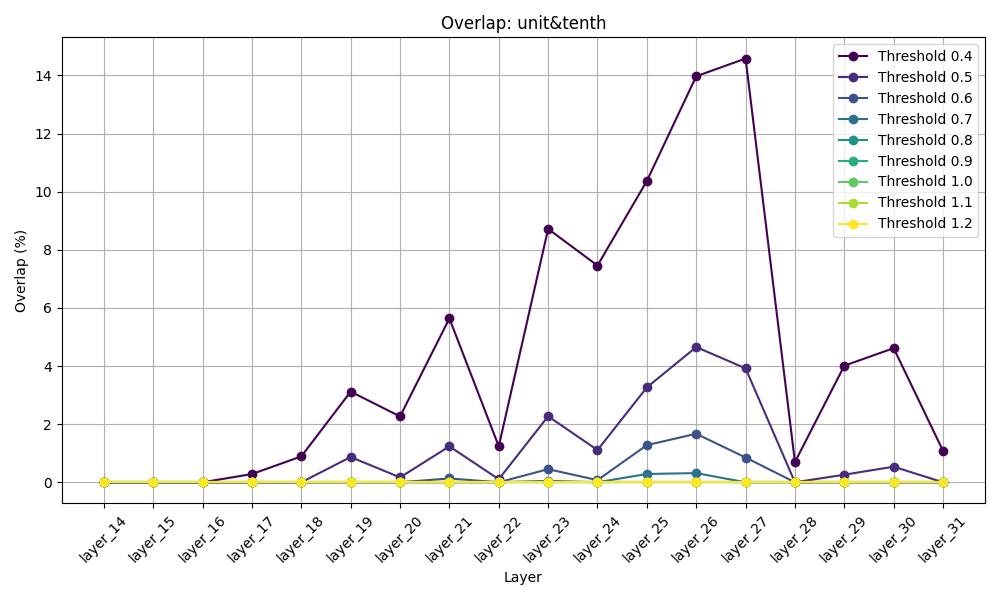}
            \caption{Unit and Tens}
            \label{fig:overlap_unit_tenth_add_olmo}
        \end{subfigure}
        \hfill
        \begin{subfigure}[b]{0.32\textwidth}
            \centering
            \includegraphics[width=\textwidth]{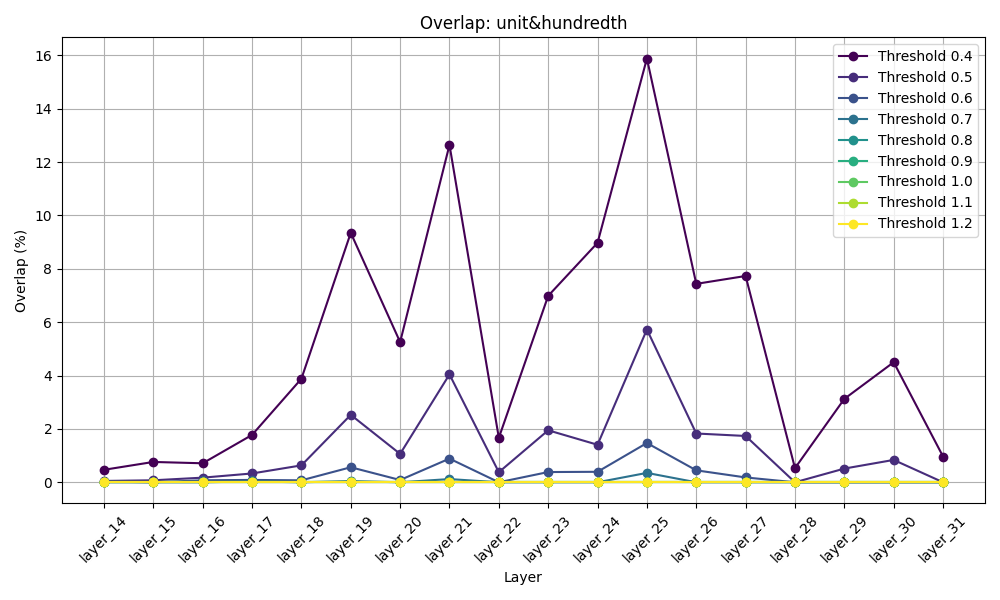}
            \caption{Unit and Hundreds}
            \label{fig:overlap_unit_hundredth_add_olmo}
        \end{subfigure}
        \hfill
        \begin{subfigure}[b]{0.32\textwidth}
            \centering
            \includegraphics[width=\textwidth]{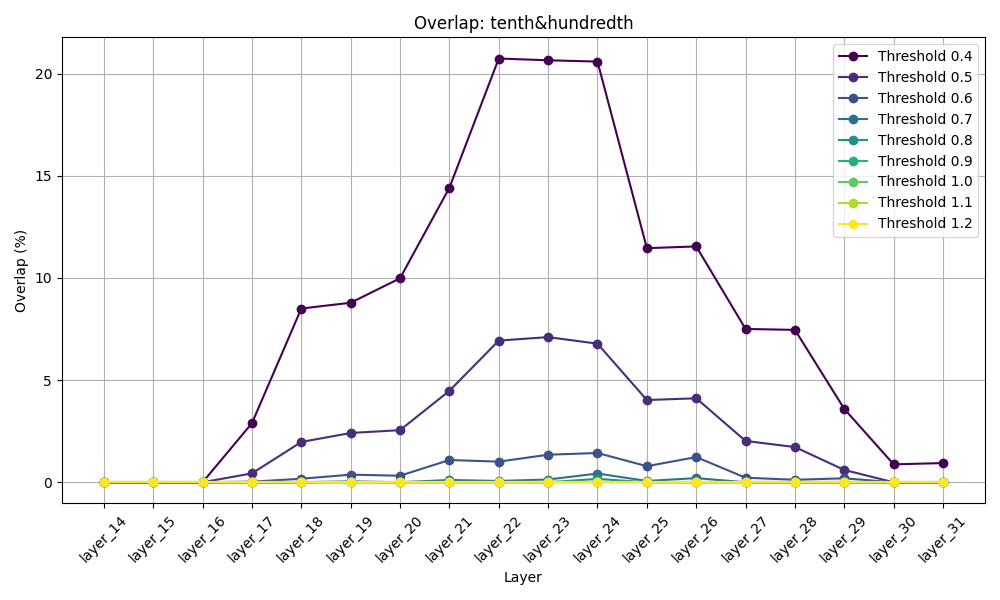}
            \caption{Tens and Hundreds}
            \label{fig:overlap_tenth_hundredth_add_olmo}
        \end{subfigure}
        \caption{\textit{Circuit Overlap}: Overlap in neurons (\%) between digit-position circuits.}
        \label{fig:row_overlap_add_olmo}
    \end{subfigure}
    
    \vspace{1em}

    \begin{subfigure}[t]{\textwidth}
        \centering
        \begin{subfigure}[b]{0.32\textwidth}
            \centering
            \includegraphics[width=\textwidth]{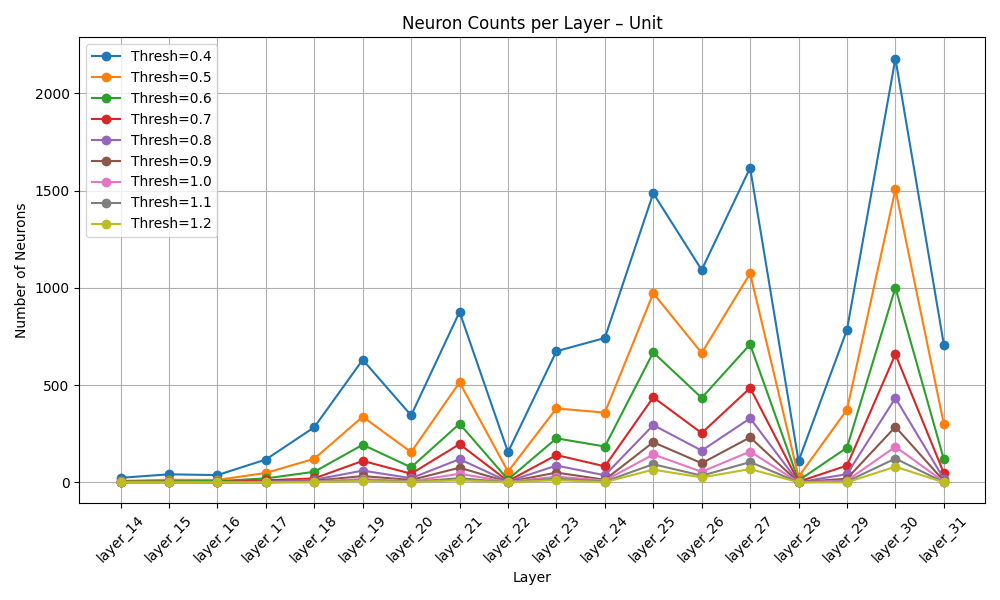}
            \caption{Unit}
            \label{fig:counts_unit_add_olmo}
        \end{subfigure}
        \hfill
        \begin{subfigure}[b]{0.32\textwidth}
            \centering
            \includegraphics[width=\textwidth]{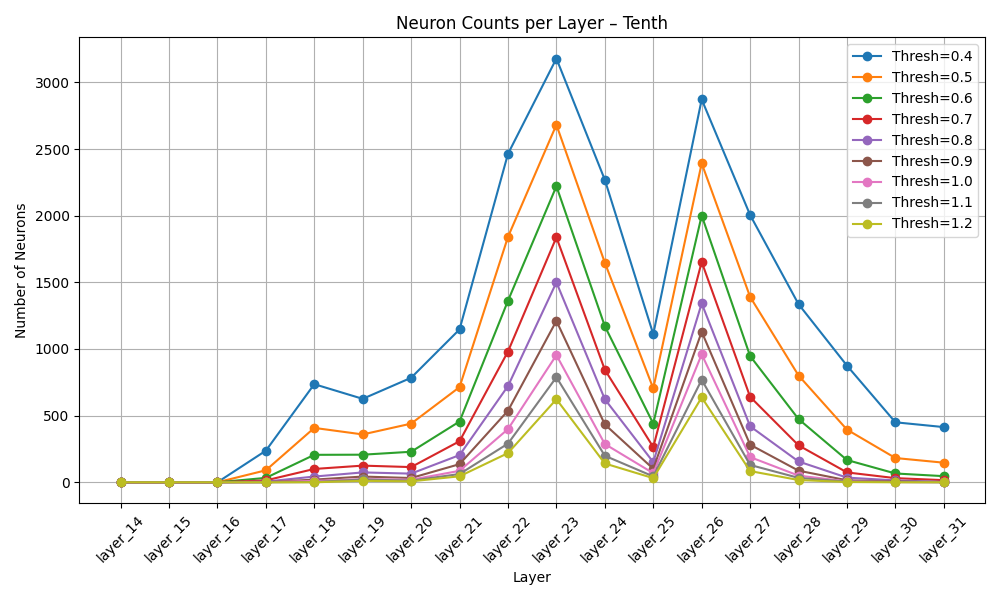}
            \caption{Tens}
            \label{fig:counts_tenth_add_olmo}
        \end{subfigure}
        \hfill
        \begin{subfigure}[b]{0.32\textwidth}
            \centering
            \includegraphics[width=\textwidth]{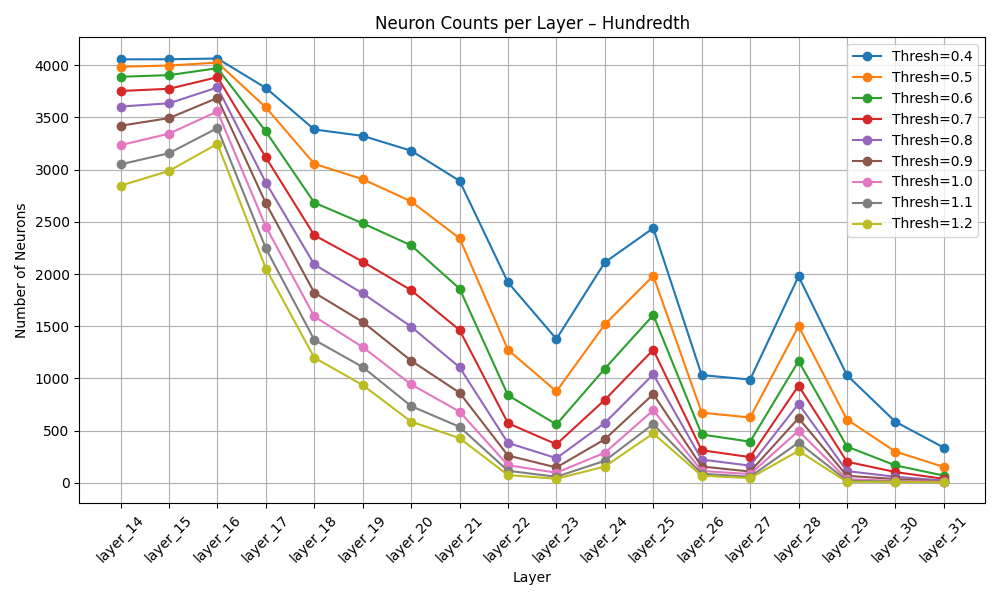}
            \caption{Hundreds}
            \label{fig:counts_hundredth_add_olmo}
        \end{subfigure}
        \caption{\textit{Circuit Size}: Number of neurons per layer in digit-position circuits (Hidden size = 4096).}
        \label{fig:row_counts_add_olmo}
    \end{subfigure}

    \vspace{1em}

    \begin{subfigure}[t]{\textwidth}
        \centering
        \begin{subfigure}[b]{0.32\textwidth}
            \centering
            \includegraphics[width=\textwidth]{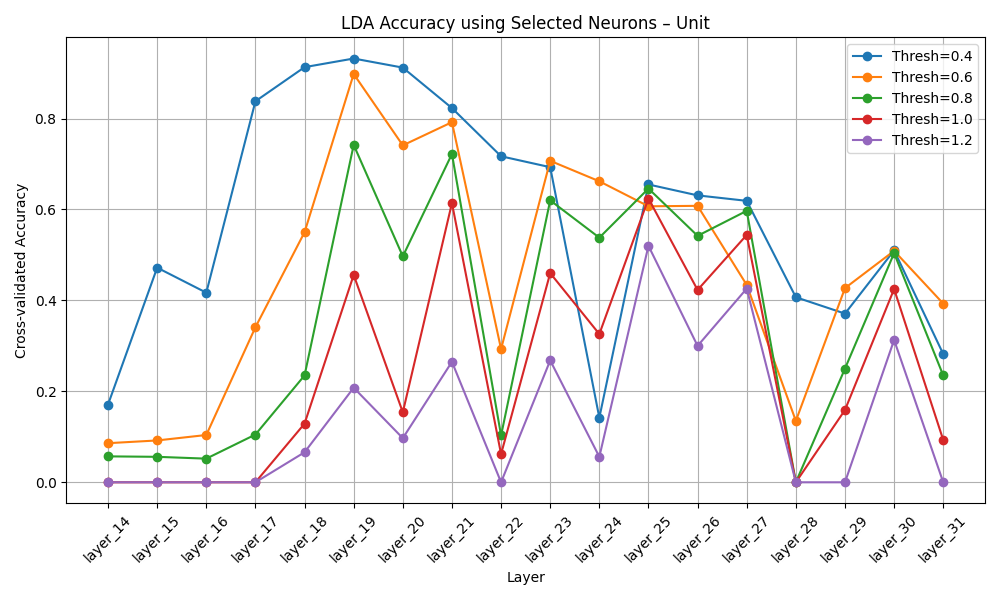}
            \caption{Unit}
            \label{fig:suff_unit_add_olmo}
        \end{subfigure}
        \hfill
        \begin{subfigure}[b]{0.32\textwidth}
            \centering
            \includegraphics[width=\textwidth]{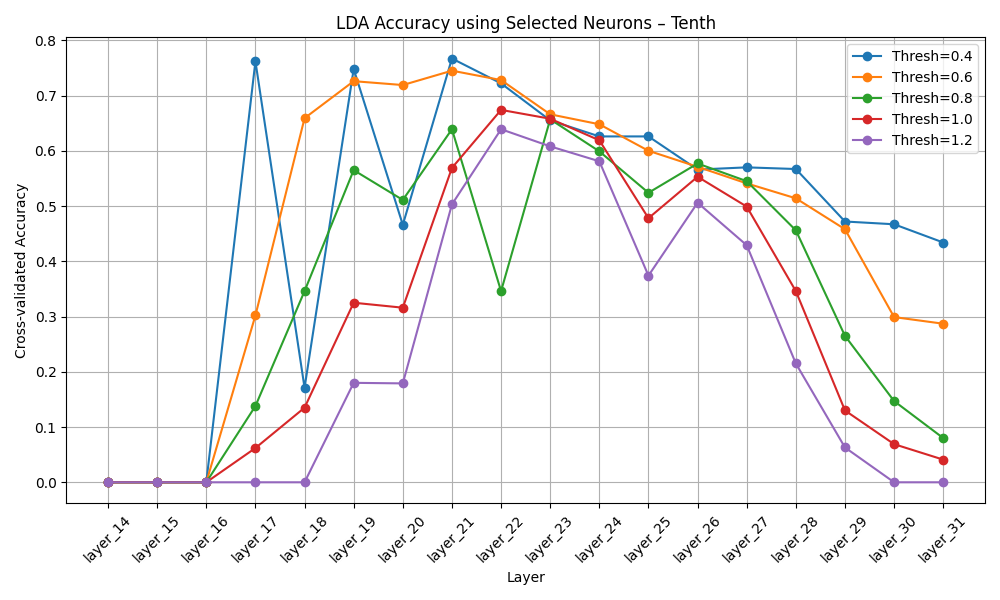}
            \caption{Tens}
            \label{fig:suff_tenth_add_olmo}
        \end{subfigure}
        \hfill
        \begin{subfigure}[b]{0.32\textwidth}
            \centering
            \includegraphics[width=\textwidth]{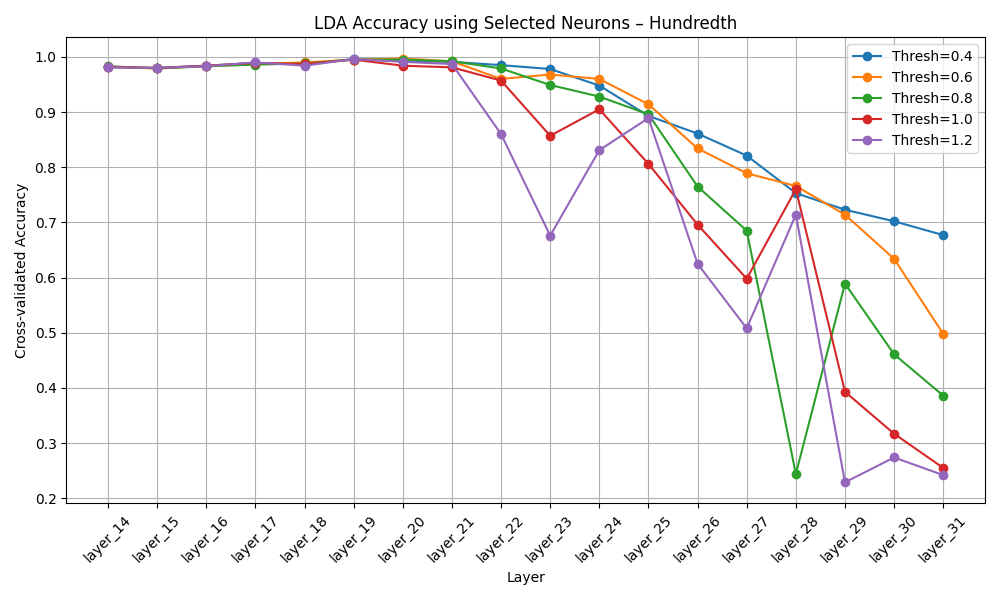}
            \caption{Hundreds}
            \label{fig:suff_hundredth_add_olmo}
        \end{subfigure}
        \caption{\textit{Circuit Sufficiency}: Sufficiency of digit-position circuit vs. full LDA}
        \label{fig:row_sufficiency_add_olmo}
    \end{subfigure}

    \caption{Olmo 2 7B, \(D_{add}\): Circuit statistics across digit positions and thresholds.}
    \label{fig:combined_fisher_add_olmo}
\end{figure*}

\begin{figure*}[h!]
    \centering

    \begin{subfigure}[t]{\textwidth}
        \centering
        \begin{subfigure}[b]{0.32\textwidth}
            \centering
            \includegraphics[width=\textwidth]{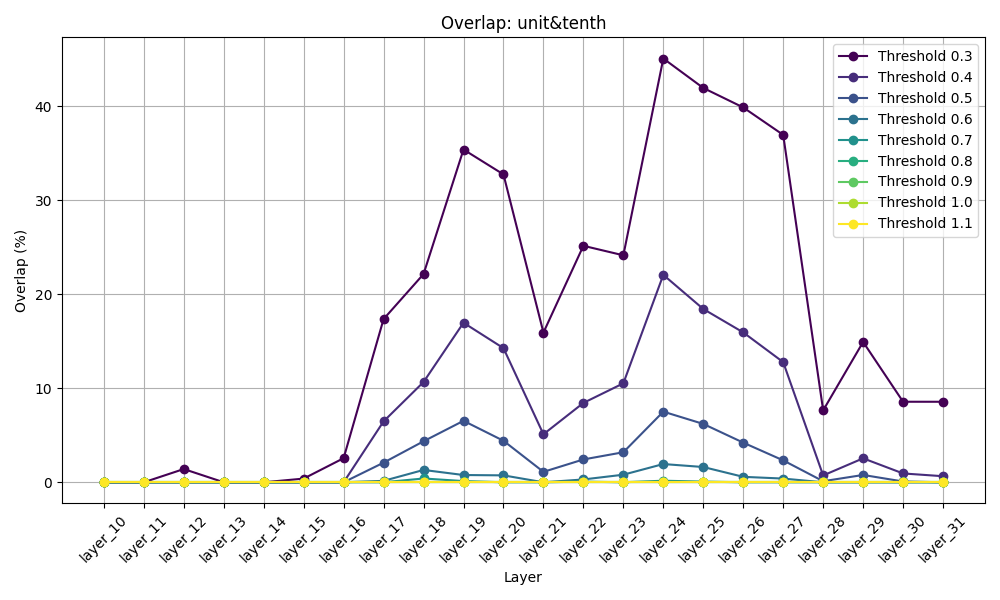}
            \caption{Unit and Tens}
            \label{fig:olmo_sub_overlap_unit_tenth_sub_olmo}
        \end{subfigure}
        \hfill
        \begin{subfigure}[b]{0.32\textwidth}
            \centering
            \includegraphics[width=\textwidth]{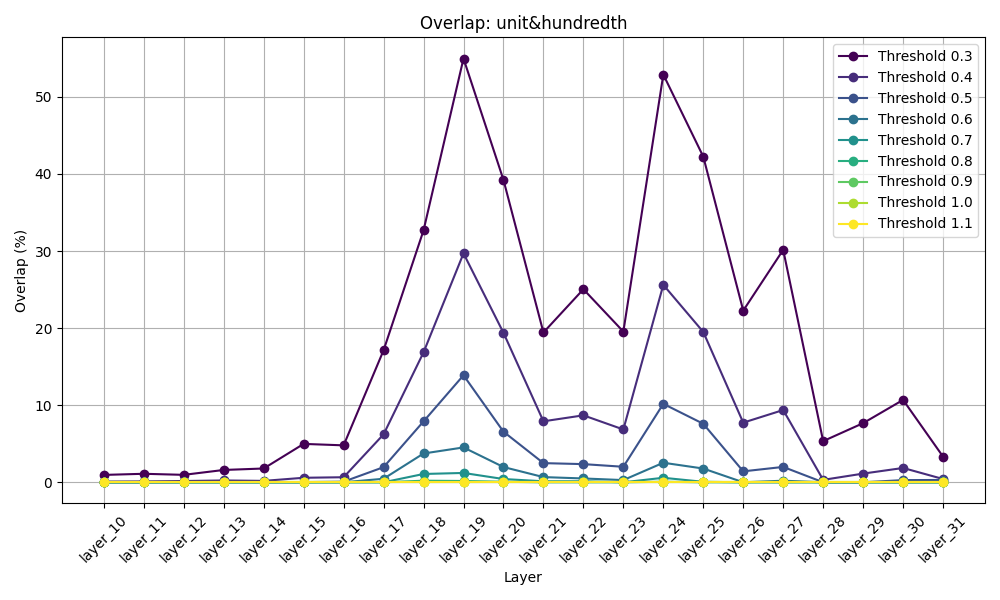}
            \caption{Unit and Hundreds}
            \label{fig:olmo_sub_overlap_unit_hundredth_sub_olmo}
        \end{subfigure}
        \hfill
        \begin{subfigure}[b]{0.32\textwidth}
            \centering
            \includegraphics[width=\textwidth]{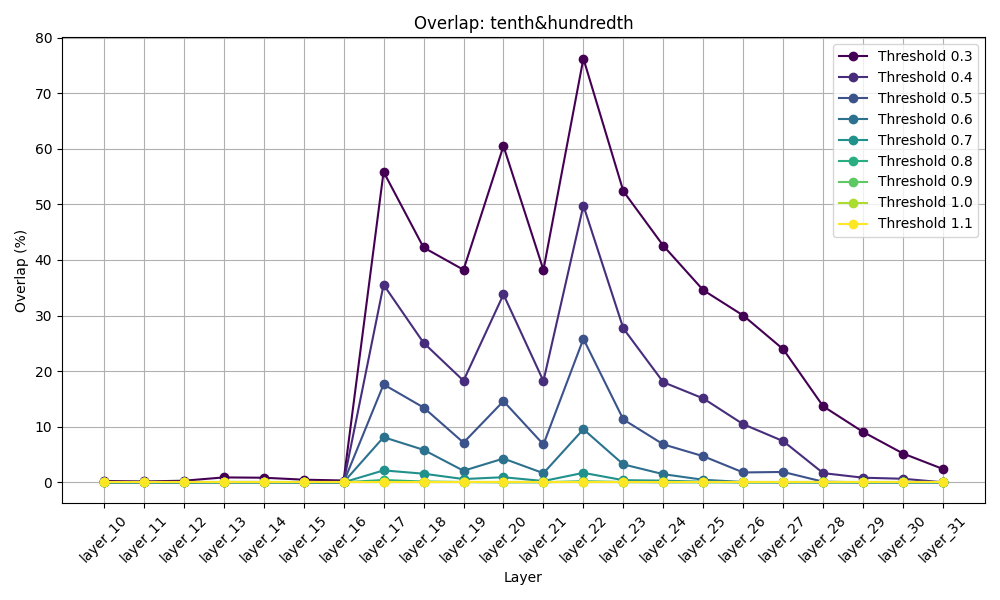}
            \caption{Tens and Hundreds}
            \label{fig:olmo_sub_overlap_tenth_hundredth_sub_olmo}
        \end{subfigure}
        \caption{\textit{Circuit Overlap}: Overlap in neurons (\%) between digit-position circuits.}
        \label{fig:olmo_sub_row_overlap_sub_olmo}
    \end{subfigure}

    \vspace{1em}

    \begin{subfigure}[t]{\textwidth}
        \centering
        \begin{subfigure}[b]{0.32\textwidth}
            \centering
            \includegraphics[width=\textwidth]{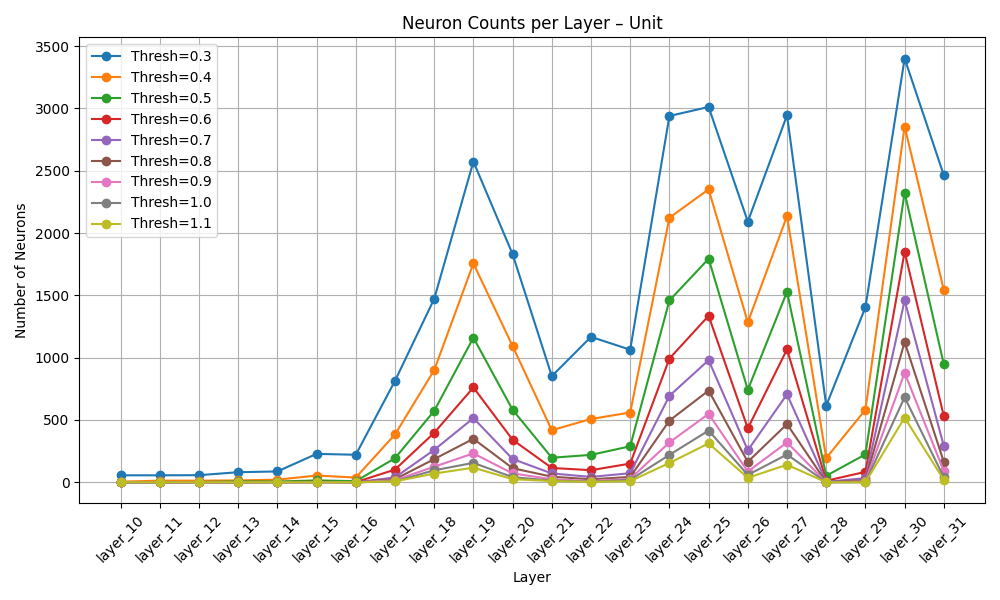}
            \caption{Unit}
            \label{fig:olmo_sub_counts_unit_sub_olmo}
        \end{subfigure}
        \hfill
        \begin{subfigure}[b]{0.32\textwidth}
            \centering
            \includegraphics[width=\textwidth]{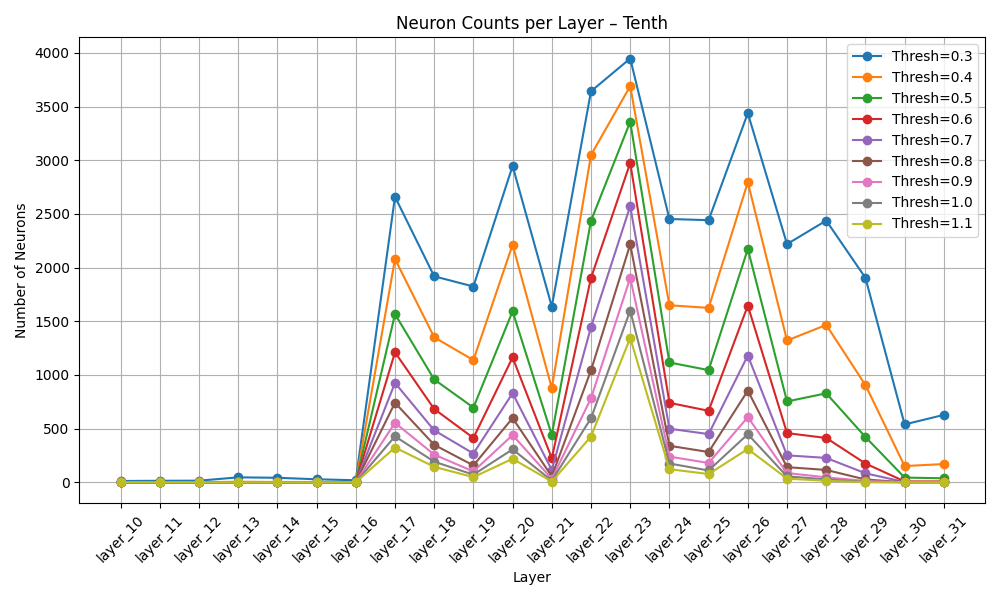}
            \caption{Tens}
            \label{fig:olmo_sub_counts_tenth_sub_olmo}
        \end{subfigure}
        \hfill
        \begin{subfigure}[b]{0.32\textwidth}
            \centering
            \includegraphics[width=\textwidth]{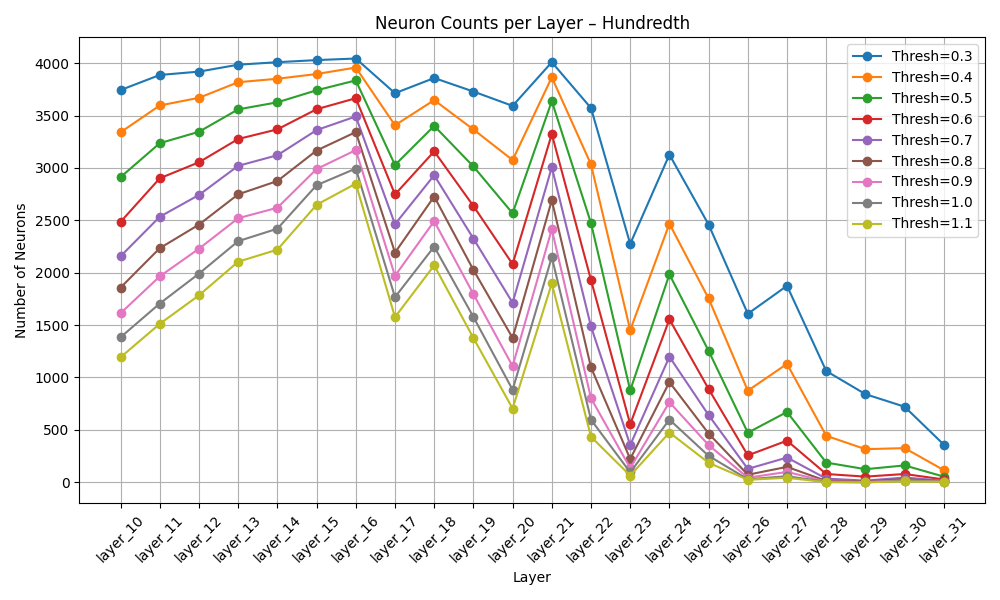}
            \caption{Hundreds}
            \label{fig:olmo_sub_counts_hundredth_sub_olmo}
        \end{subfigure}
        \caption{\textit{Circuit Size}: Number of neurons per layer in digit-position circuits (Hidden size = 4096).}
        \label{fig:olmo_sub_row_counts_sub_olmo}
    \end{subfigure}

    \vspace{1em}

    \begin{subfigure}[t]{\textwidth}
        \centering
        \begin{subfigure}[b]{0.32\textwidth}
            \centering
            \includegraphics[width=\textwidth]{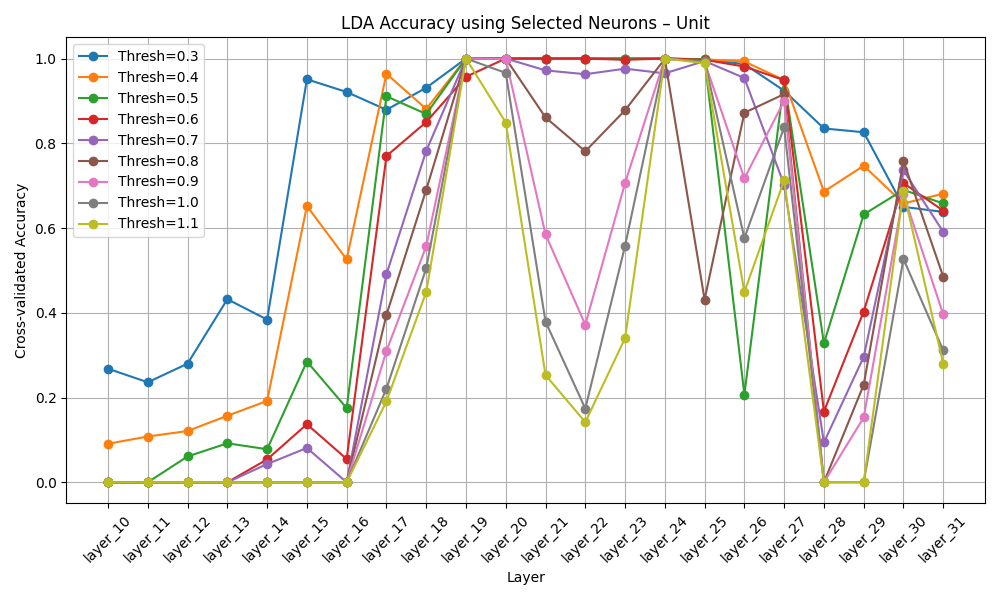}
            \caption{Unit}
            \label{fig:olmo_sub_suff_unit_sub_olmo}
        \end{subfigure}
        \hfill
        \begin{subfigure}[b]{0.32\textwidth}
            \centering
            \includegraphics[width=\textwidth]{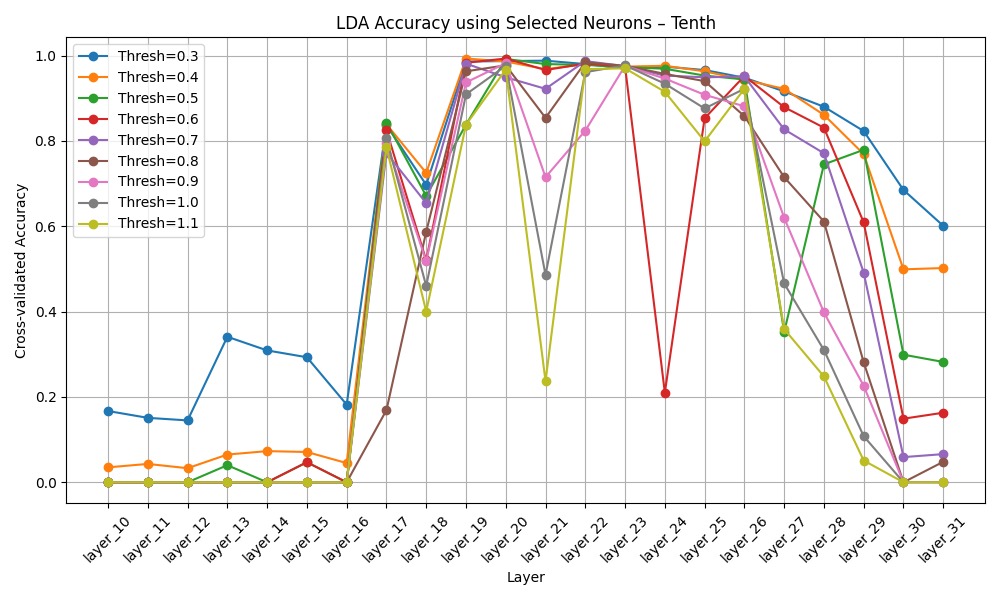}
            \caption{Tens}
            \label{fig:olmo_sub_suff_tenth_sub_olmo}
        \end{subfigure}
        \hfill
        \begin{subfigure}[b]{0.32\textwidth}
            \centering
            \includegraphics[width=\textwidth]{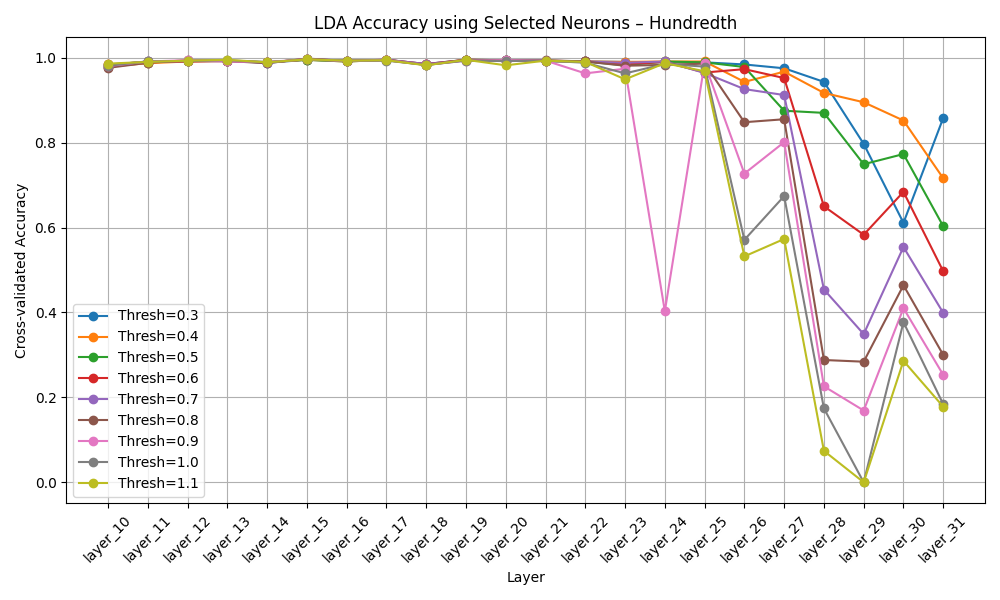}
            \caption{Hundreds}
            \label{fig:olmo_sub_suff_hundredth_sub_olmo}
        \end{subfigure}
        \caption{\textit{Circuit Sufficiency}: Sufficiency of digit-position circuit vs. full LDA}
        \label{fig:olmo_sub_row_sufficiency_sub_olmo}
    \end{subfigure}

    \caption{Olmo 2 7B, \(D_{sub}\): Circuit statistics across digit positions and thresholds.}
    \label{fig:olmo_sub_combined_fisher_sub_olmo}
\end{figure*}

\begin{figure*}[h!]
    \centering

    \begin{subfigure}[t]{\textwidth}
        \centering
        \begin{subfigure}[b]{0.32\textwidth}
            \centering
            \includegraphics[width=\textwidth]{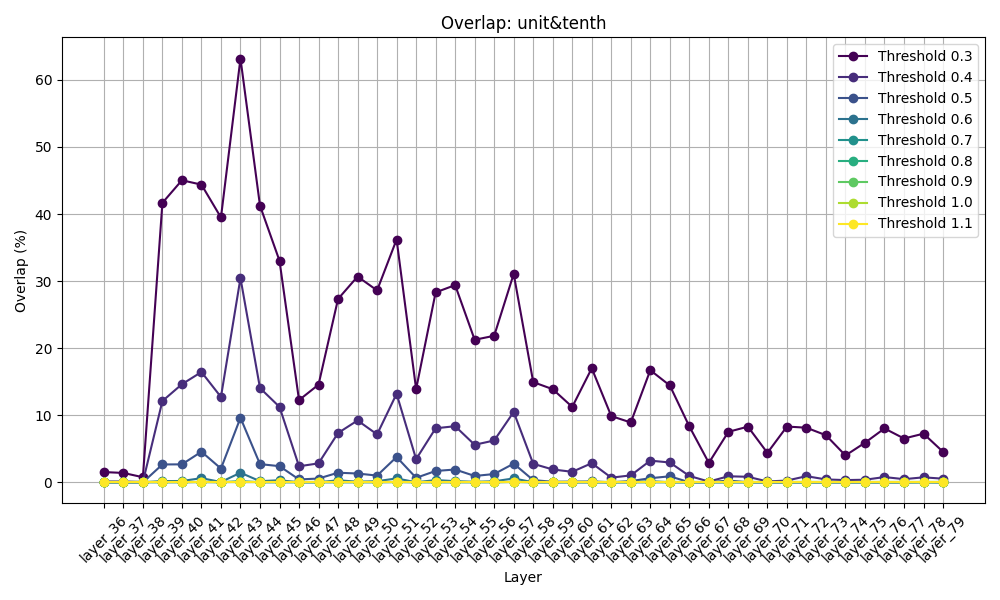}
            \caption{Unit and Tens}
            \label{fig:llama70b_add_overlap_unit_tenth}
        \end{subfigure}
        \hfill
        \begin{subfigure}[b]{0.32\textwidth}
            \centering
            \includegraphics[width=\textwidth]{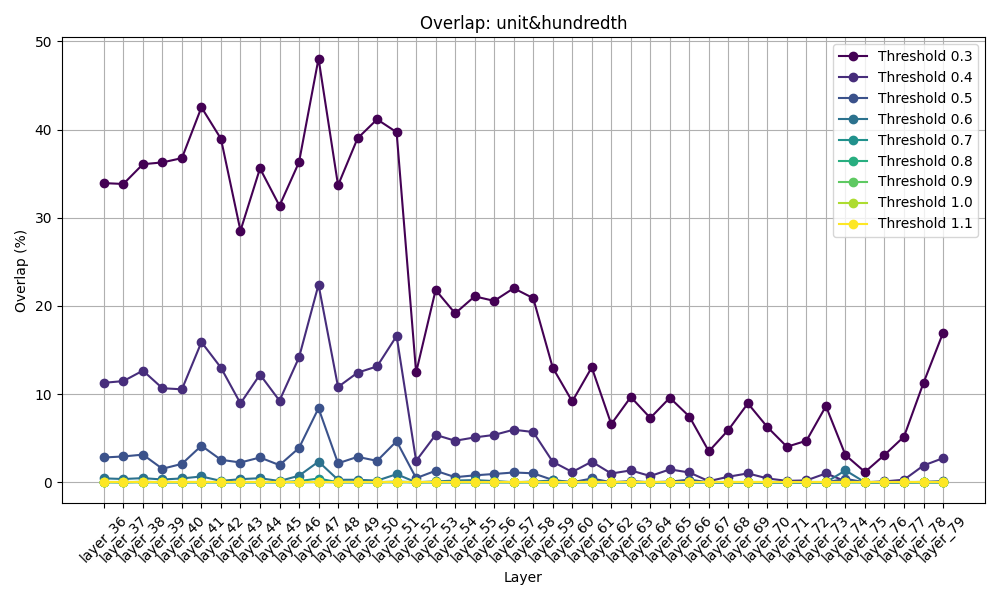}
            \caption{Unit and Hundreds}
            \label{fig:llama70b_add_overlap_unit_hundredth}
        \end{subfigure}
        \hfill
        \begin{subfigure}[b]{0.32\textwidth}
            \centering
            \includegraphics[width=\textwidth]{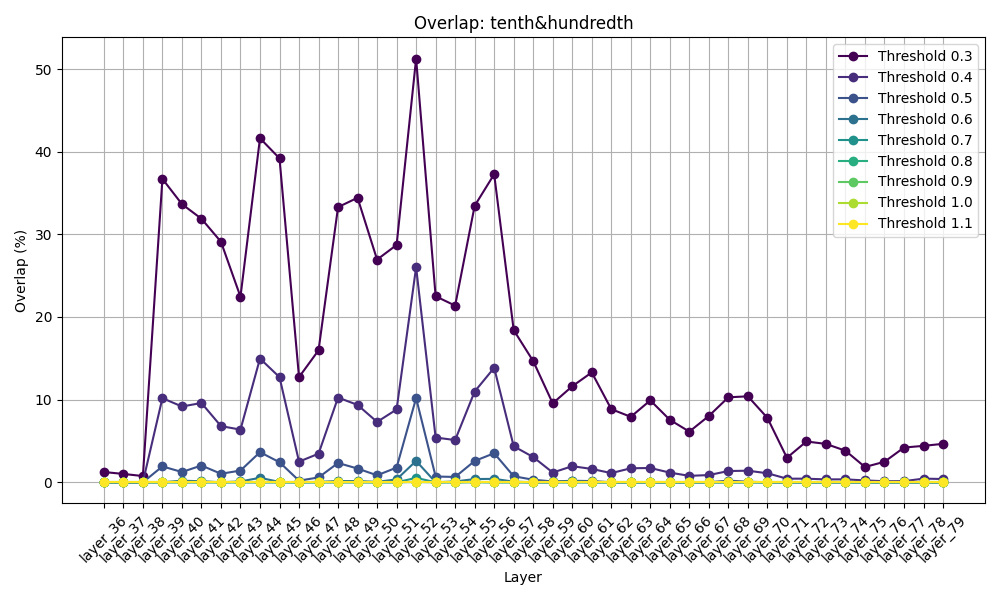}
            \caption{Tens and Hundreds}
            \label{fig:llama70b_add_overlap_tenth_hundredth}
        \end{subfigure}
        \caption{\textit{Circuit Overlap}: Overlap in neurons (\%) between digit-position circuits.}
        \label{fig:llama70b_add_row_overlap}
    \end{subfigure}

    \vspace{1em}

    \begin{subfigure}[t]{\textwidth}
        \centering
        \begin{subfigure}[b]{0.32\textwidth}
            \centering
            \includegraphics[width=\textwidth]{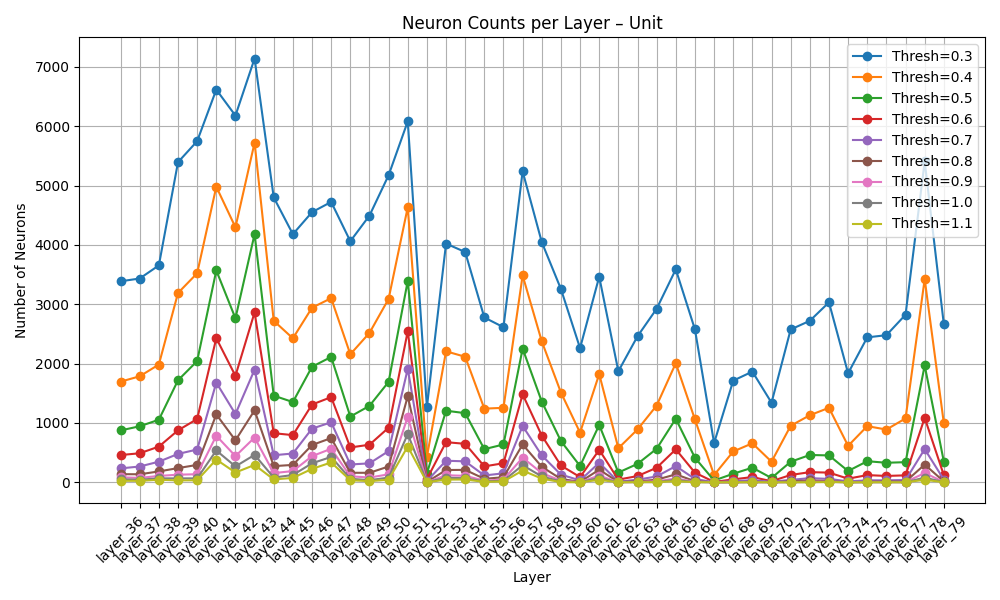}
            \caption{Unit}
            \label{fig:llama70b_add_counts_unit}
        \end{subfigure}
        \hfill
        \begin{subfigure}[b]{0.32\textwidth}
            \centering
            \includegraphics[width=\textwidth]{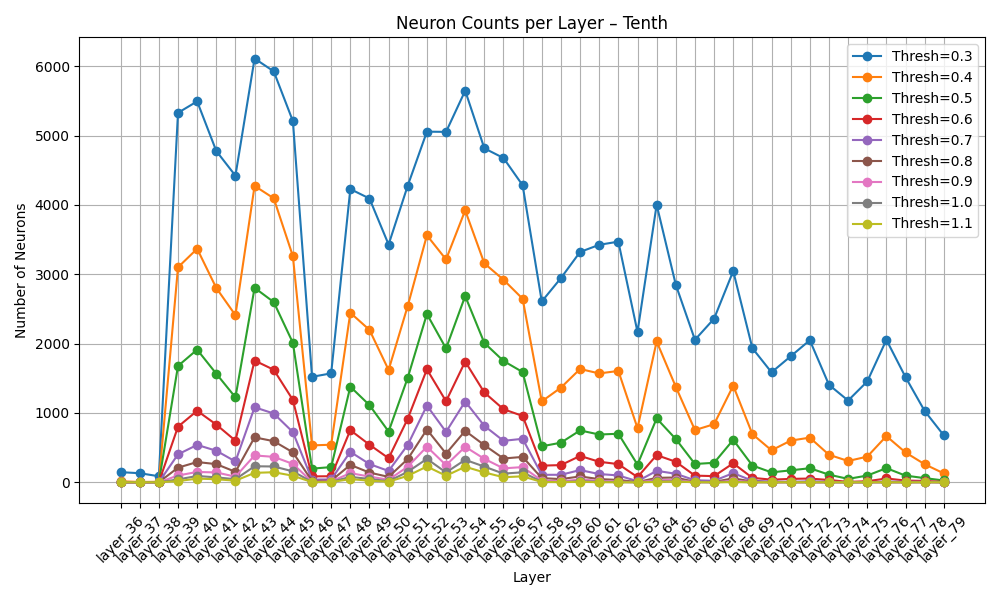}
            \caption{Tens}
            \label{fig:llama70b_add_counts_tenth}
        \end{subfigure}
        \hfill
        \begin{subfigure}[b]{0.32\textwidth}
            \centering
            \includegraphics[width=\textwidth]{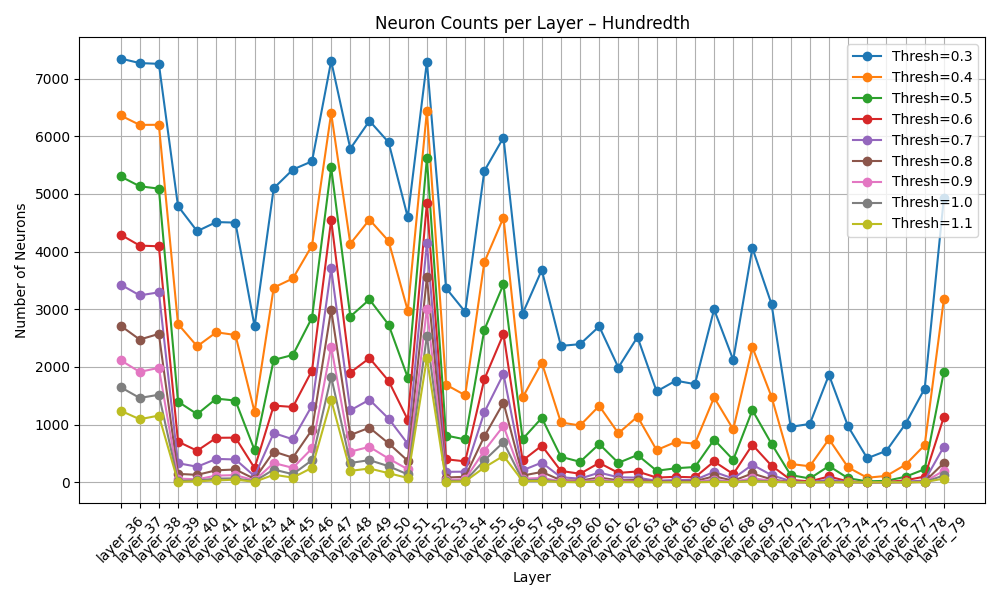}
            \caption{Hundreds}
            \label{fig:llama70b_add_counts_hundredth}
        \end{subfigure}
        \caption{\textit{Circuit Size}: Number of neurons per layer in digit-position circuits (Hidden size = 8192).}
        \label{fig:llama70b_add_row_counts}
    \end{subfigure}

    \vspace{1em}

    \begin{subfigure}[t]{\textwidth}
        \centering
        \begin{subfigure}[b]{0.32\textwidth}
            \centering
            \includegraphics[width=\textwidth]{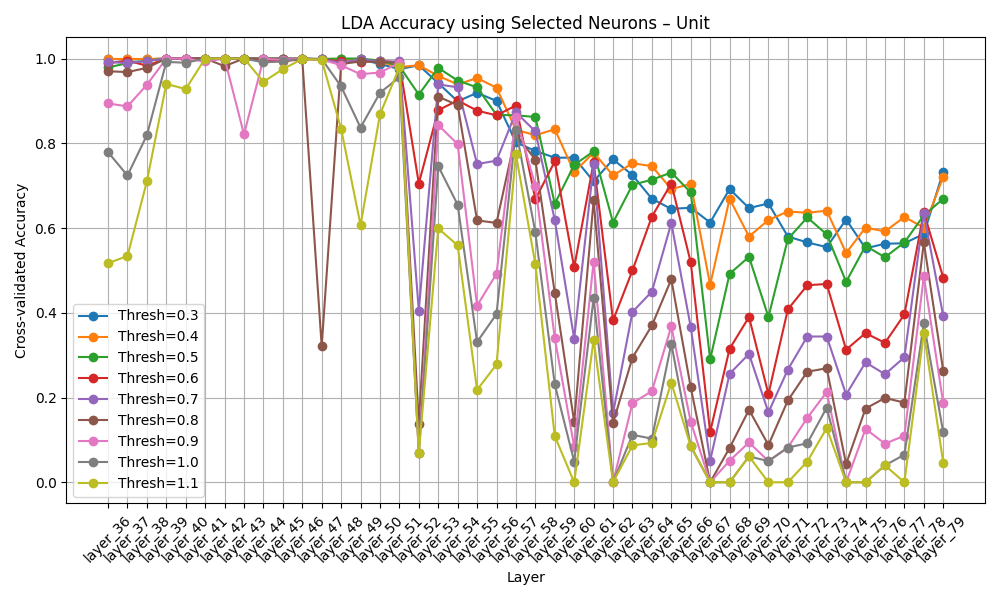}
            \caption{Unit}
            \label{fig:llama70b_add_suff_unit}
        \end{subfigure}
        \hfill
        \begin{subfigure}[b]{0.32\textwidth}
            \centering
            \includegraphics[width=\textwidth]{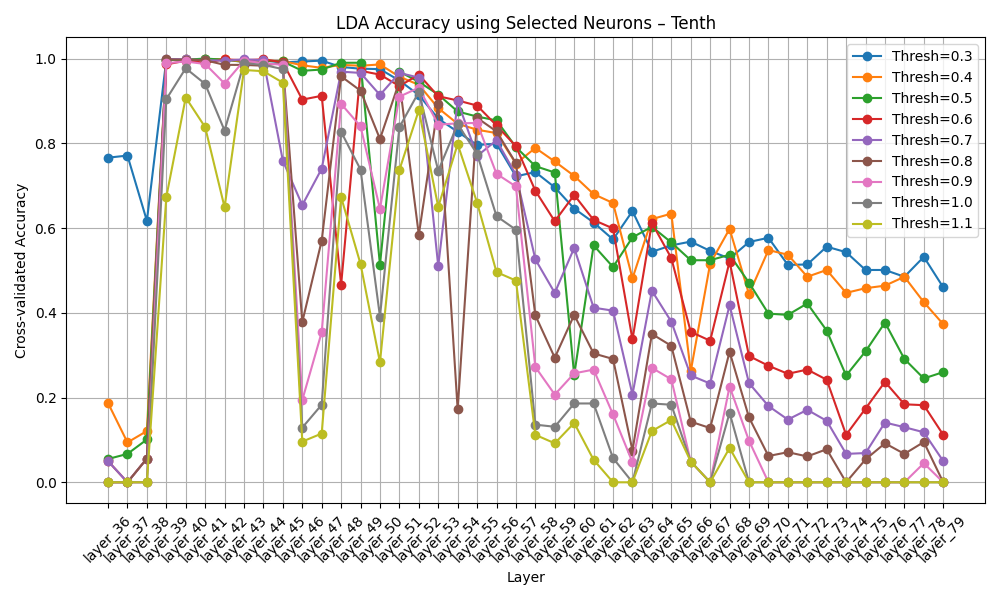}
            \caption{Tens}
            \label{fig:llama70b_add_suff_tenth}
        \end{subfigure}
        \hfill
        \begin{subfigure}[b]{0.32\textwidth}
            \centering
            \includegraphics[width=\textwidth]{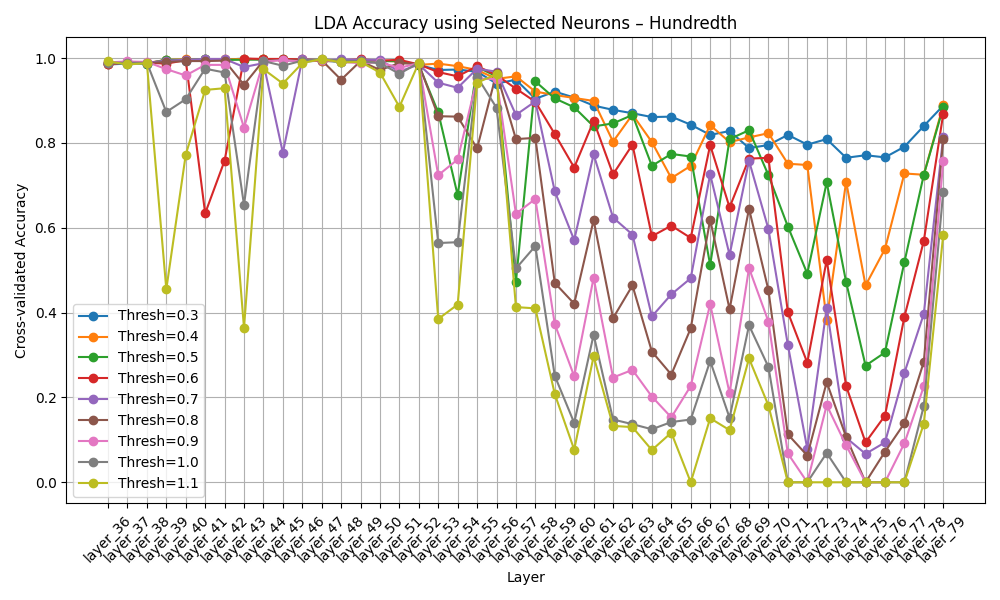}
            \caption{Hundreds}
            \label{fig:llama70b_add_suff_hundredth}
        \end{subfigure}
        \caption{\textit{Circuit Sufficiency}: Sufficiency of digit-position circuit vs. full LDA}
        \label{fig:llama70b_add_row_sufficiency}
    \end{subfigure}

    \caption{Llama 3 70B, \(D_{add}\): Circuit statistics across digit positions and thresholds.}
    \label{fig:llama70b_add_combined_fisher}
\end{figure*}

\begin{figure*}[h!]
    \centering

    \begin{subfigure}[t]{\textwidth}
        \centering
        \begin{subfigure}[b]{0.32\textwidth}
            \centering
            \includegraphics[width=\textwidth]{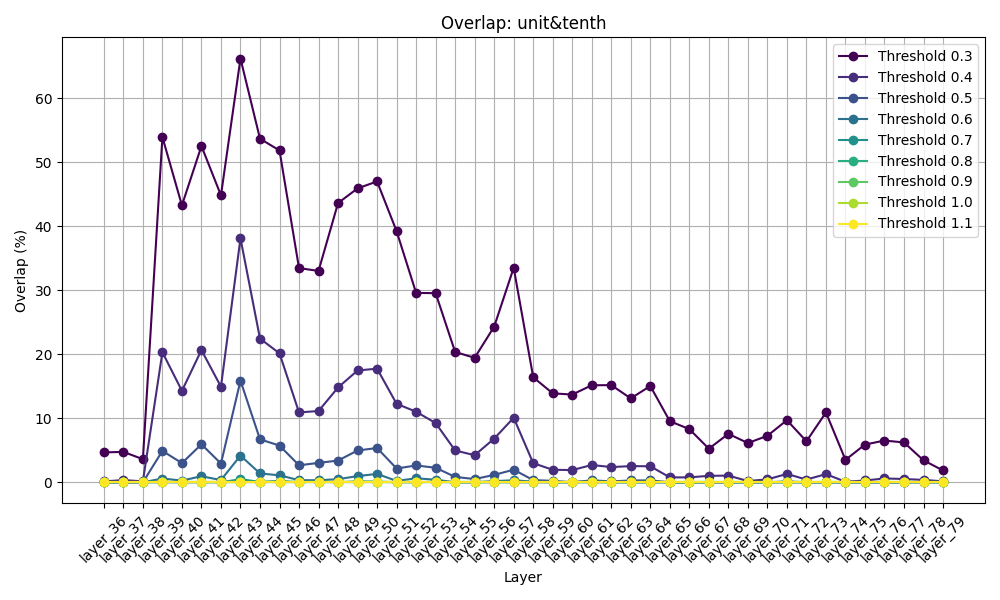}
            \caption{Unit and Tens}
            \label{fig:llama70b_sub_overlap_unit_tenth}
        \end{subfigure}
        \hfill
        \begin{subfigure}[b]{0.32\textwidth}
            \centering
            \includegraphics[width=\textwidth]{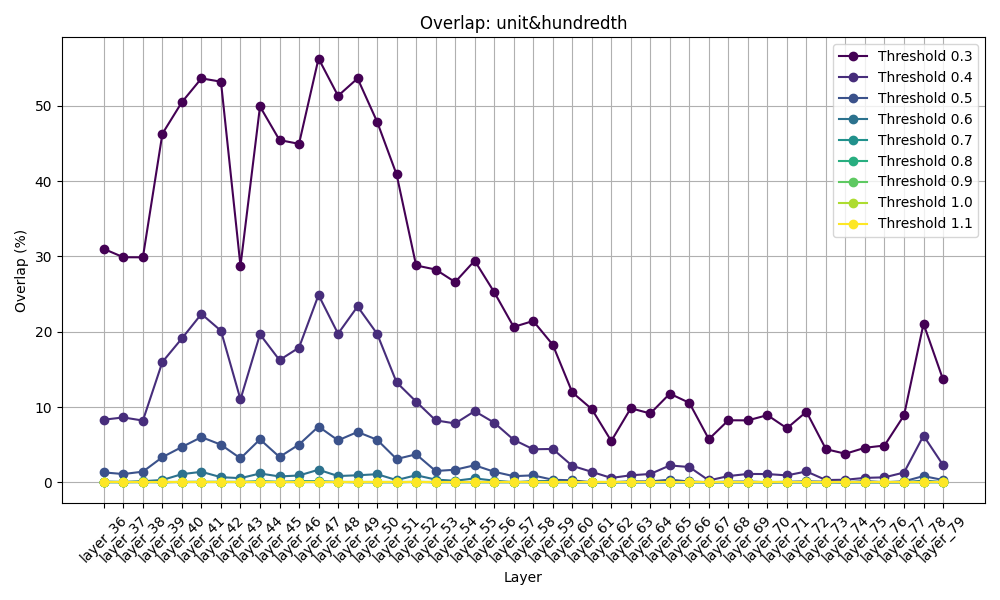}
            \caption{Unit and Hundreds}
            \label{fig:llama70b_sub_overlap_unit_hundredth}
        \end{subfigure}
        \hfill
        \begin{subfigure}[b]{0.32\textwidth}
            \centering
            \includegraphics[width=\textwidth]{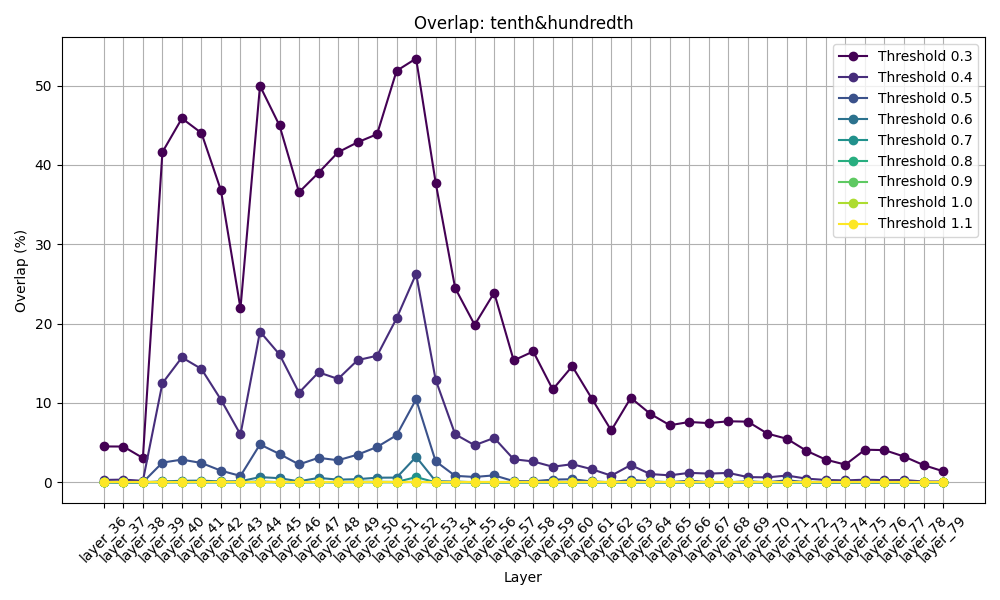}
            \caption{Tens and Hundreds}
            \label{fig:llama70b_sub_overlap_tenth_hundredth}
        \end{subfigure}
        \caption{\textit{Circuit Overlap}: Overlap in neurons (\%) between digit-position circuits.}
        \label{fig:llama70b_sub_row_overlap}
    \end{subfigure}

    \vspace{1em}

    \begin{subfigure}[t]{\textwidth}
        \centering
        \begin{subfigure}[b]{0.32\textwidth}
            \centering
            \includegraphics[width=\textwidth]{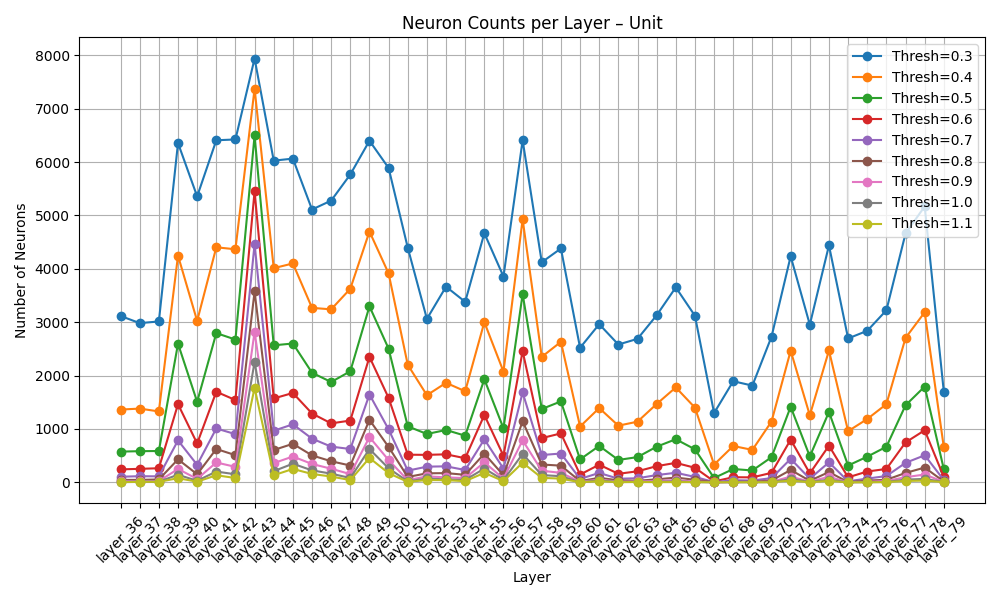}
            \caption{Unit}
            \label{fig:llama70b_sub_counts_unit}
        \end{subfigure}
        \hfill
        \begin{subfigure}[b]{0.32\textwidth}
            \centering
            \includegraphics[width=\textwidth]{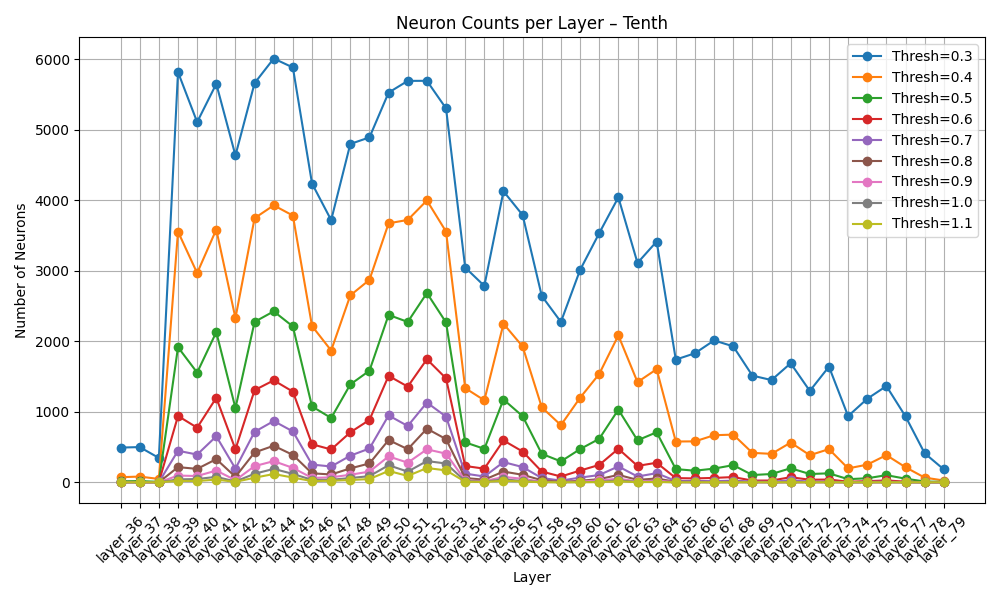}
            \caption{Tens}
            \label{fig:llama70b_sub_counts_tenth}
        \end{subfigure}
        \hfill
        \begin{subfigure}[b]{0.32\textwidth}
            \centering
            \includegraphics[width=\textwidth]{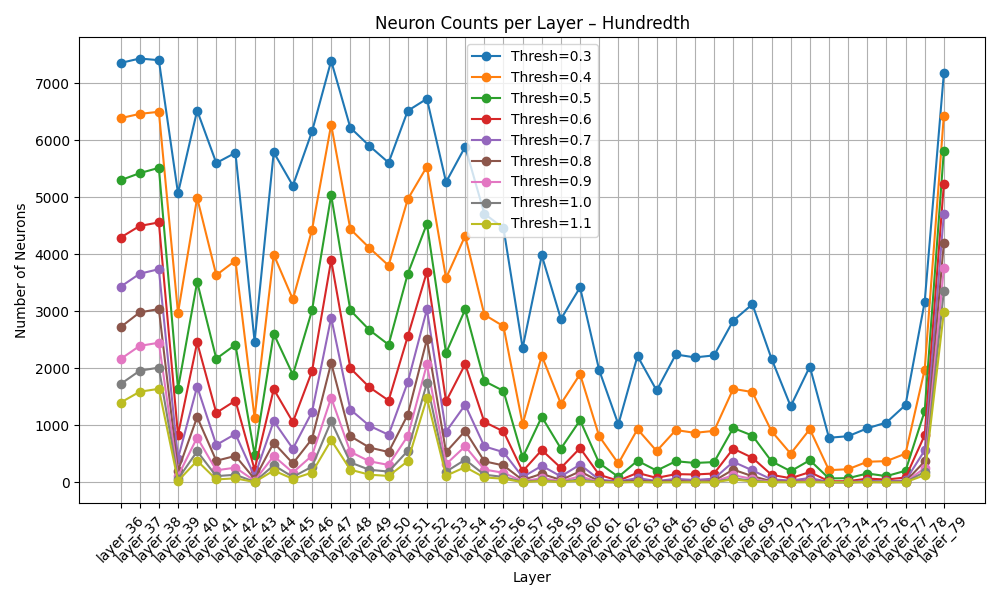}
            \caption{Hundreds}
            \label{fig:llama70b_sub_counts_hundredth}
        \end{subfigure}
        \caption{\textit{Circuit Size}: Number of neurons per layer in digit-position circuits (Hidden size = 8192).}
        \label{fig:llama70b_sub_row_counts}
    \end{subfigure}

    \vspace{1em}

    \begin{subfigure}[t]{\textwidth}
        \centering
        \begin{subfigure}[b]{0.32\textwidth}
            \centering
            \includegraphics[width=\textwidth]{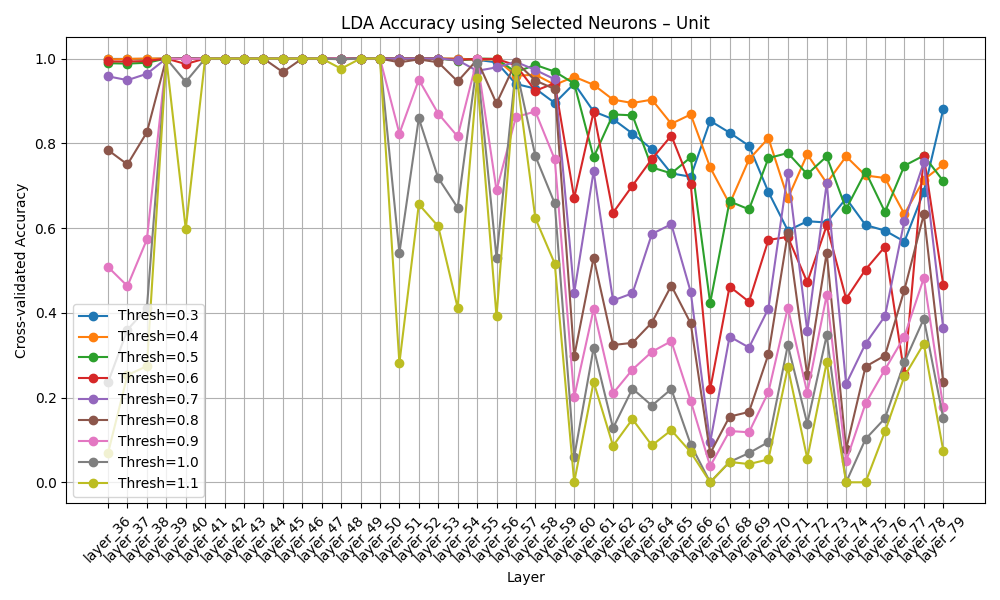}
            \caption{Unit}
            \label{fig:llama70b_sub_suff_unit}
        \end{subfigure}
        \hfill
        \begin{subfigure}[b]{0.32\textwidth}
            \centering
            \includegraphics[width=\textwidth]{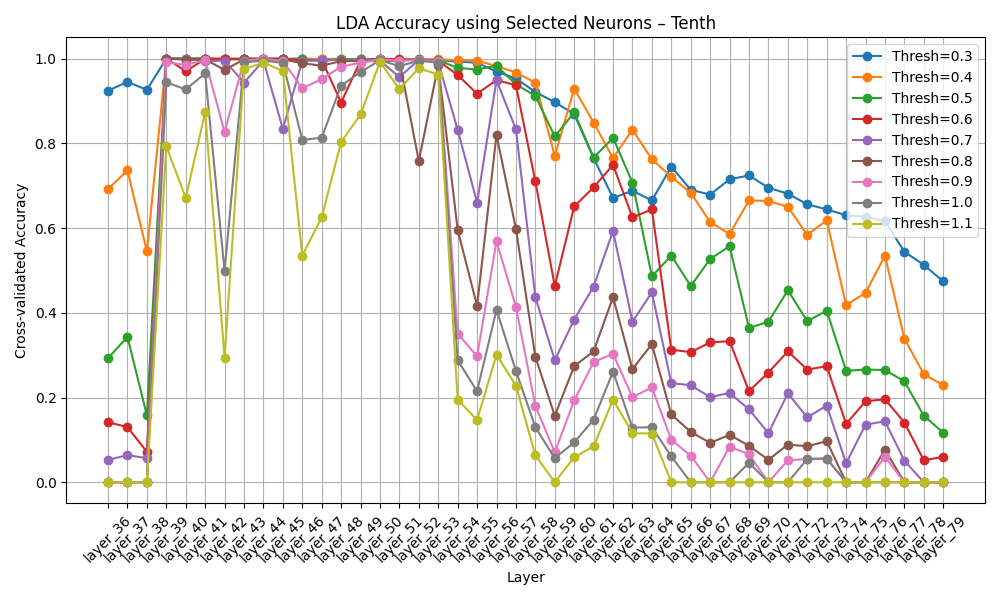}
            \caption{Tens}
            \label{fig:llama70b_sub_suff_tenth}
        \end{subfigure}
        \hfill
        \begin{subfigure}[b]{0.32\textwidth}
            \centering
            \includegraphics[width=\textwidth]{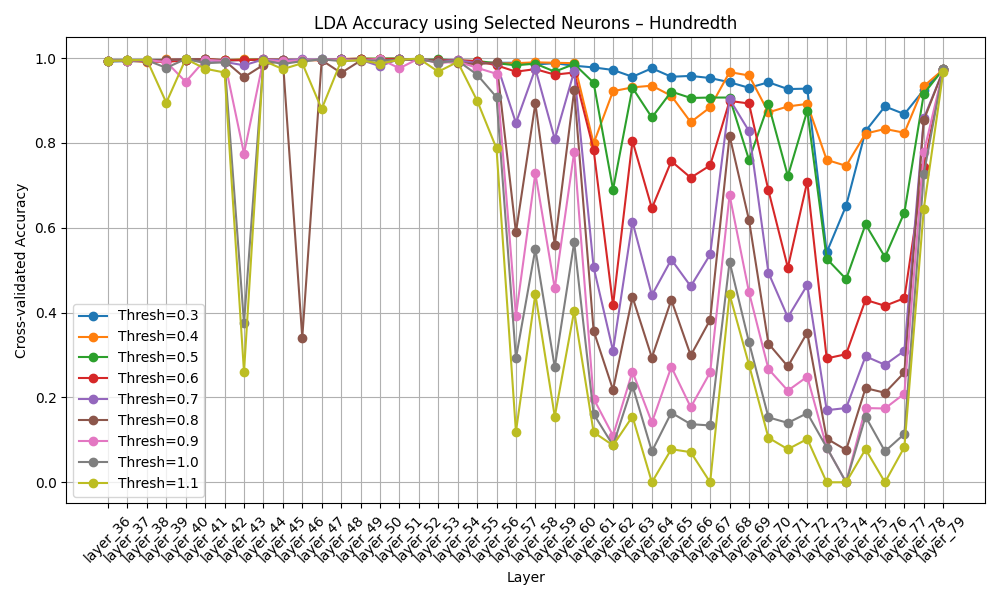}
            \caption{Hundreds}
            \label{fig:llama70b_sub_suff_hundredth}
        \end{subfigure}
        \caption{\textit{Circuit Sufficiency}: Sufficiency of digit-position circuit vs. full LDA}
        \label{fig:llama70b_sub_row_sufficiency}
    \end{subfigure}

    \caption{Llama 3 70B, \(D_{sub}\): Circuit statistics across digit positions and thresholds.}
    \label{fig:llama70b_sub_combined_fisher}
\end{figure*}

\clearpage
\newpage
\begin{figure*}[h!]
\section{Effect of chosen Threshold on Digit-Circuit Intervention}
\label{app:more_models}
    \centering
    \hfill
    \begin{subfigure}[b]{0.32\textwidth}
        \centering
        \includegraphics[width=\textwidth]{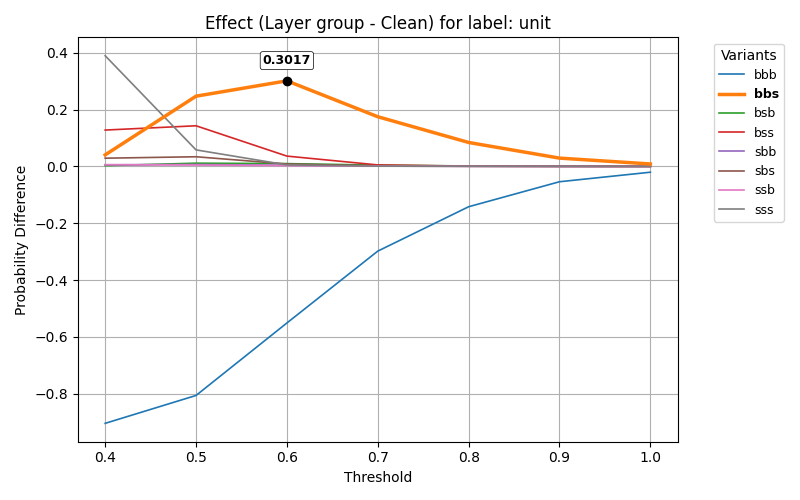}
        \caption{Unit Circuit intervention}
    \end{subfigure}
    \hfill
    \begin{subfigure}[b]{0.32\textwidth}
        \centering
        \includegraphics[width=\textwidth]{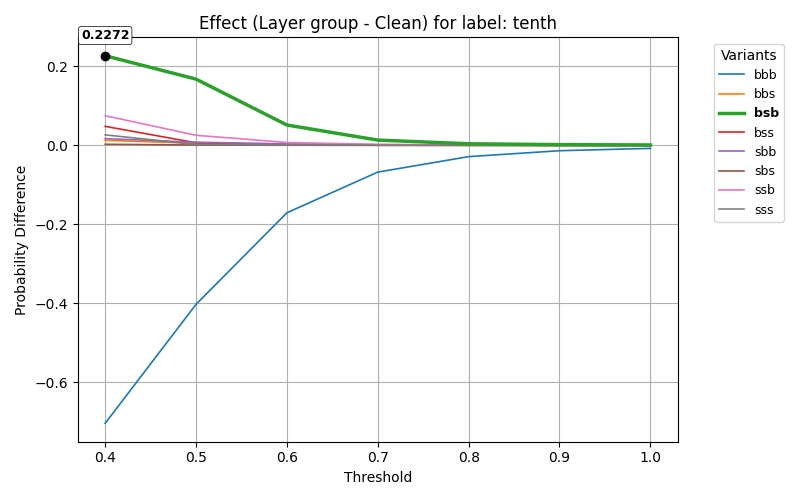} 
        \caption{Tens Circuit intervention}
    \end{subfigure}
    \begin{subfigure}[b]{0.32\textwidth}
        \centering
        \includegraphics[width=\textwidth]{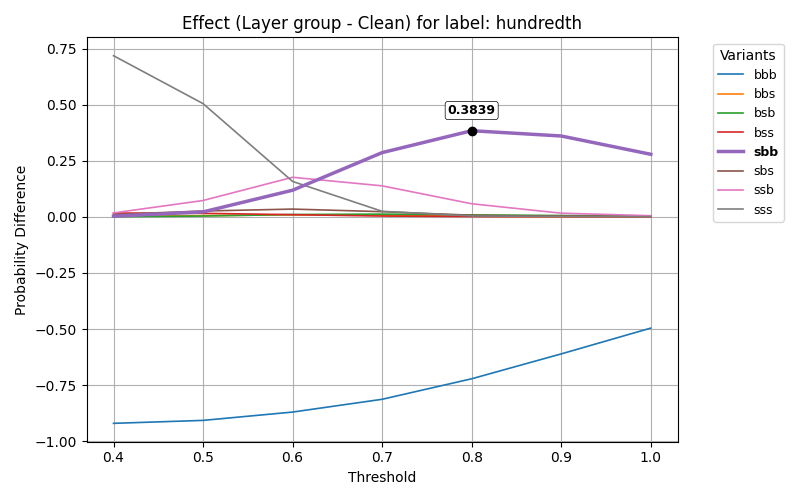}
        \caption{Hundreds Circuit Intervention}
    \end{subfigure}
    \caption{\textbf{Llama 3 8B}, \(D_{add, op1}\): Effect size of circuit specific interventions with different thresholds for neuron circuit membership, on circuit neurons in layers \(L = \{ 16, ..., 24\}\).}
    \label{fig:Threshold_Results_Llama8B_Addition_Op1}
\end{figure*}

\begin{figure*}[h!]
    \centering
    \hfill
    \begin{subfigure}[b]{0.32\textwidth}
        \centering
        \includegraphics[width=\textwidth]{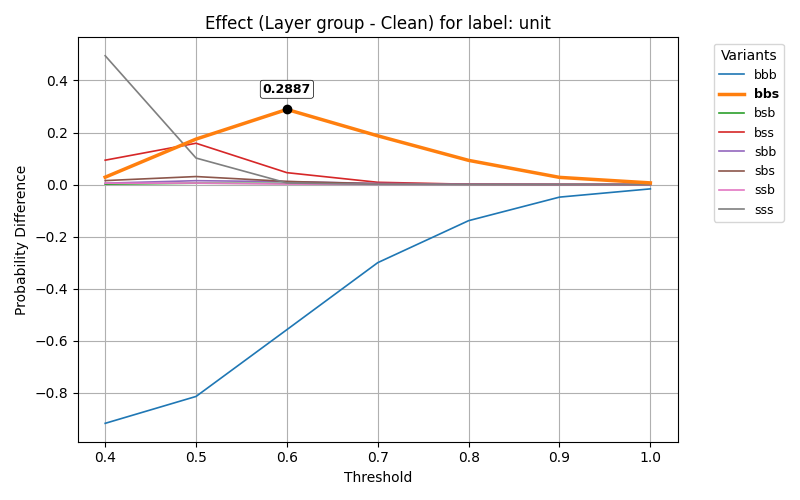}
        \caption{Unit Circuit intervention}
    \end{subfigure}
    \hfill
    \begin{subfigure}[b]{0.32\textwidth}
        \centering
        \includegraphics[width=\textwidth]{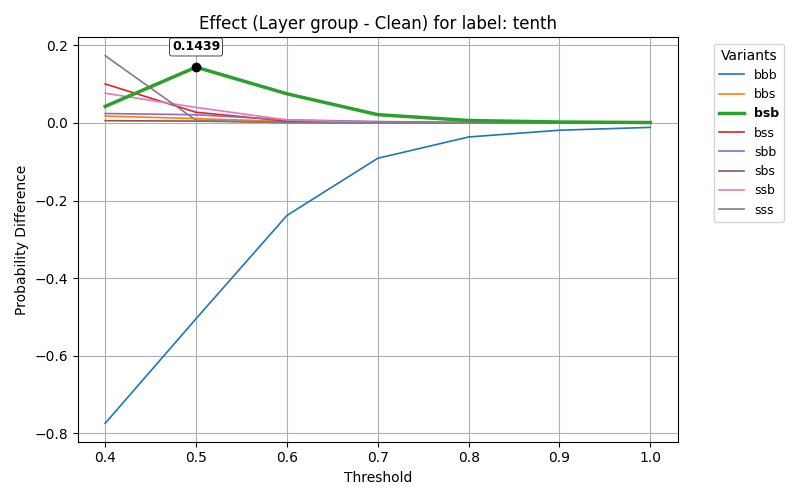} 
        \caption{Tens Circuit intervention}
    \end{subfigure}
    \begin{subfigure}[b]{0.32\textwidth}
        \centering
        \includegraphics[width=\textwidth]{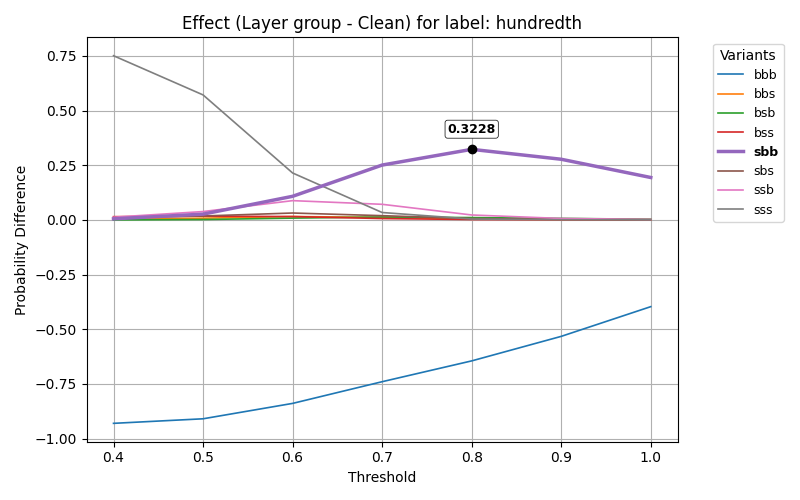}
        \caption{Hundreds Circuit Intervention}
    \end{subfigure}
    \caption{\textbf{Llama 3 8B}, \(D_{sub, op1}\): Effect size of circuit specific interventions with different thresholds for neuron circuit membership, on circuit neurons in layers \(L = \{ 16, ..., 28\}\).}
    \label{fig:Threshold_Results_Llama8B_Subtraction_Op1}
\end{figure*}

\begin{figure*}[h!]
    \centering
    \hfill
    \begin{subfigure}[b]{0.32\textwidth}
        \centering
        \includegraphics[width=\textwidth]{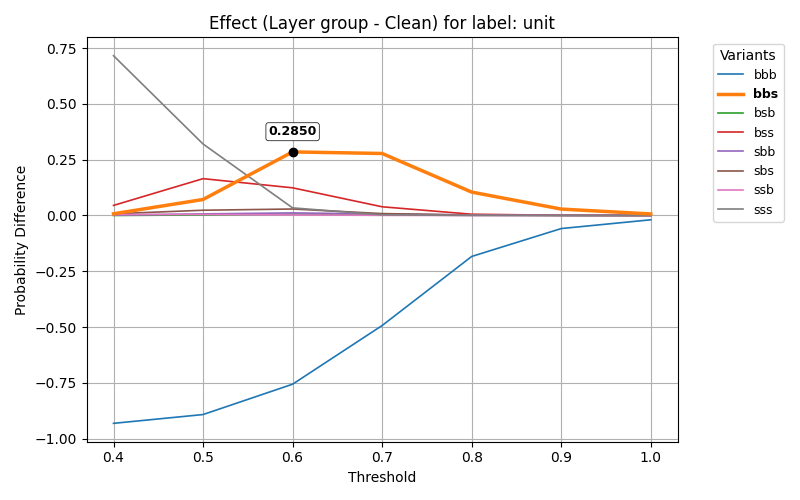}
        \caption{Unit Circuit intervention}
    \end{subfigure}
    \hfill
    \begin{subfigure}[b]{0.32\textwidth}
        \centering
        \includegraphics[width=\textwidth]{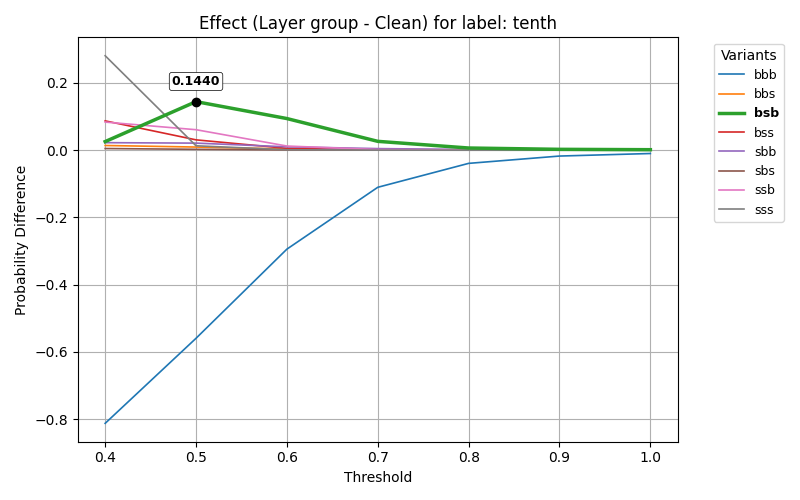} 
        \caption{Tens Circuit intervention}
    \end{subfigure}
    \begin{subfigure}[b]{0.32\textwidth}
        \centering
        \includegraphics[width=\textwidth]{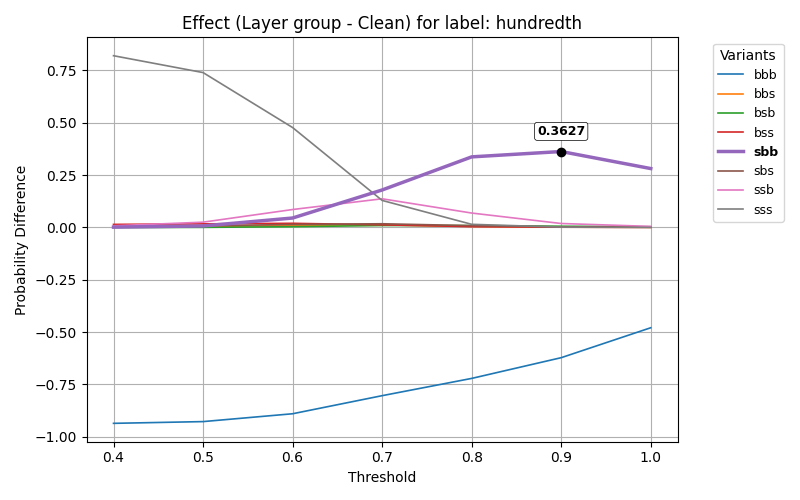}
        \caption{Hundreds Circuit Intervention}
    \end{subfigure}
    \caption{\textbf{Llama 3 8B}, \(D_{sub, op2}\): Effect size of circuit specific interventions with different thresholds for neuron circuit membership, on circuit neurons in layers \(L = \{ 15, ..., 28\}\). }
    \label{fig:Threshold_Results_Llama8B_Subtraction_Op2}
\end{figure*}

\begin{figure*}[h!]
    \centering
    \hfill
    \begin{subfigure}[b]{0.32\textwidth}
        \centering
        \includegraphics[width=\textwidth]{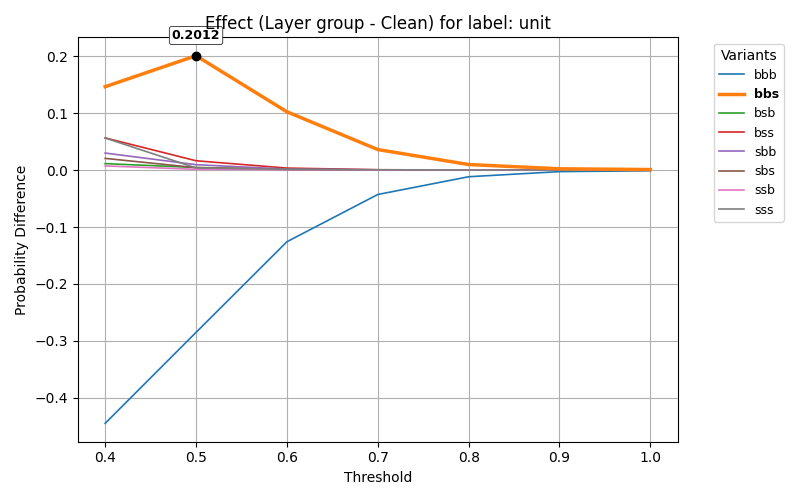}
        \caption{Unit Circuit intervention}
    \end{subfigure}
    \hfill
    \begin{subfigure}[b]{0.32\textwidth}
        \centering
        \includegraphics[width=\textwidth]{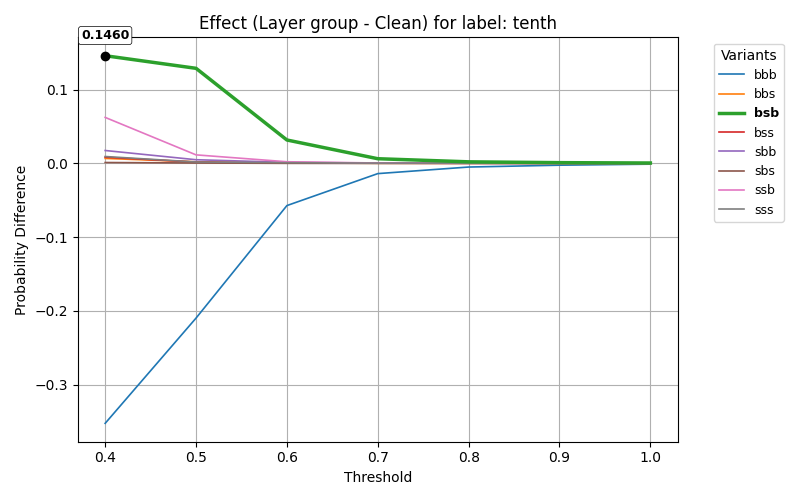} 
        \caption{Tens Circuit intervention}
    \end{subfigure}
    \begin{subfigure}[b]{0.32\textwidth}
        \centering
        \includegraphics[width=\textwidth]{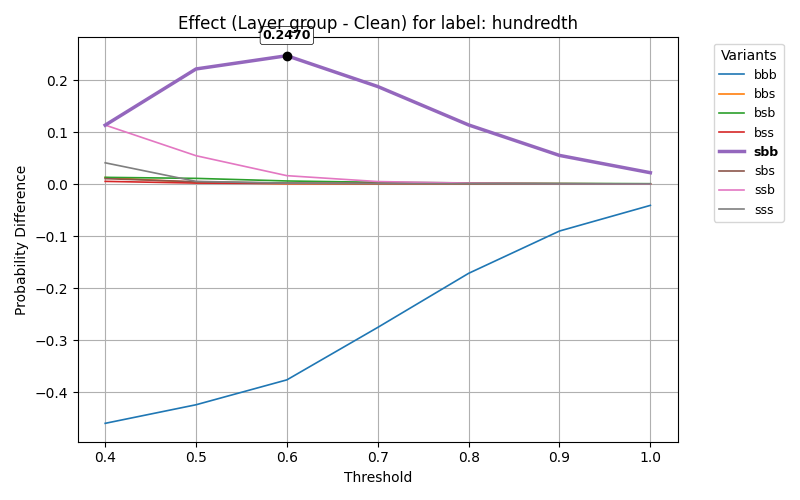}
        \caption{Hundreds Circuit Intervention}
    \end{subfigure}
    \caption{\textbf{Llama3-70B}, \(D_{add, op1}\): Effect size of circuit specific interventions with different thresholds for neuron circuit membership, on circuit neurons in layers \(L = \{ 39, ..., 56\}\).}
    \label{fig:Threshold_Results_Llama70B_Addition_Op1}
\end{figure*}

\begin{figure*}[h!]
    \centering
    \hfill
    \begin{subfigure}[b]{0.32\textwidth}
        \centering
        \includegraphics[width=\textwidth]{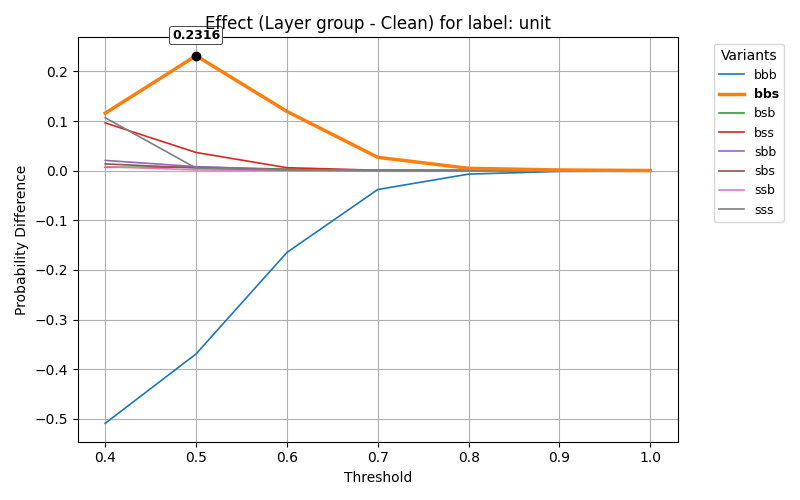}
        \caption{Unit Circuit intervention}
    \end{subfigure}
    \hfill
    \begin{subfigure}[b]{0.32\textwidth}
        \centering
        \includegraphics[width=\textwidth]{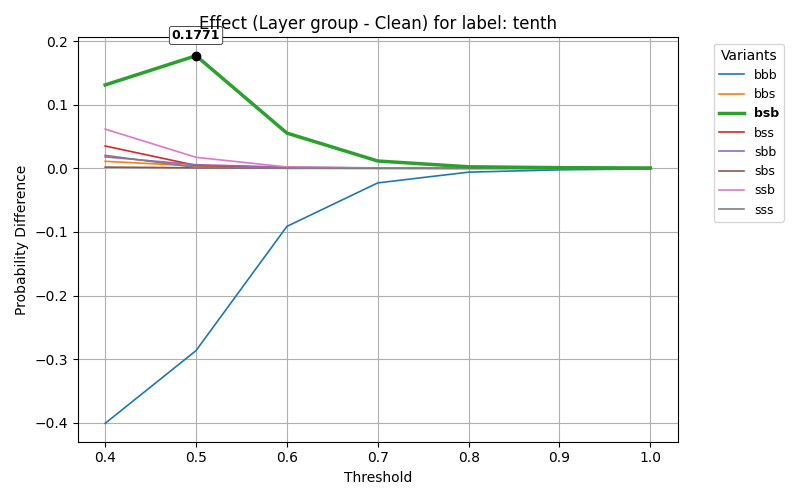} 
        \caption{Tens Circuit intervention}
    \end{subfigure}
    \begin{subfigure}[b]{0.32\textwidth}
        \centering
        \includegraphics[width=\textwidth]{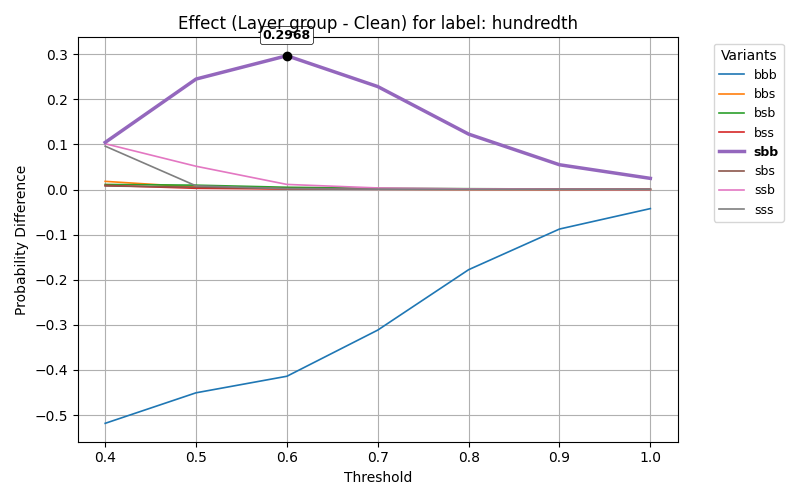}
        \caption{Hundreds Circuit Intervention}
    \end{subfigure}
    \caption{\textbf{Llama3-70B}, \(D_{add, op2}\): Effect size of circuit specific interventions with different thresholds for neuron circuit membership, on circuit neurons in layers \(L = \{ 39, ..., 56\}\).}
    \label{fig:Threshold_Results_Llama70B_Addition_Op2}
\end{figure*}

\begin{figure*}[h!]
    \centering
    \hfill
    \begin{subfigure}[b]{0.32\textwidth}
        \centering
        \includegraphics[width=\textwidth]{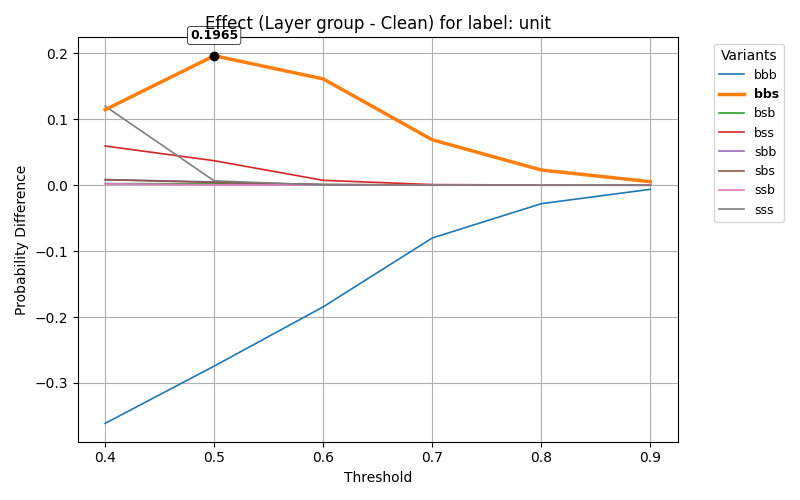}
        \caption{Unit Circuit intervention}
    \end{subfigure}
    \hfill
    \begin{subfigure}[b]{0.32\textwidth}
        \centering
        \includegraphics[width=\textwidth]{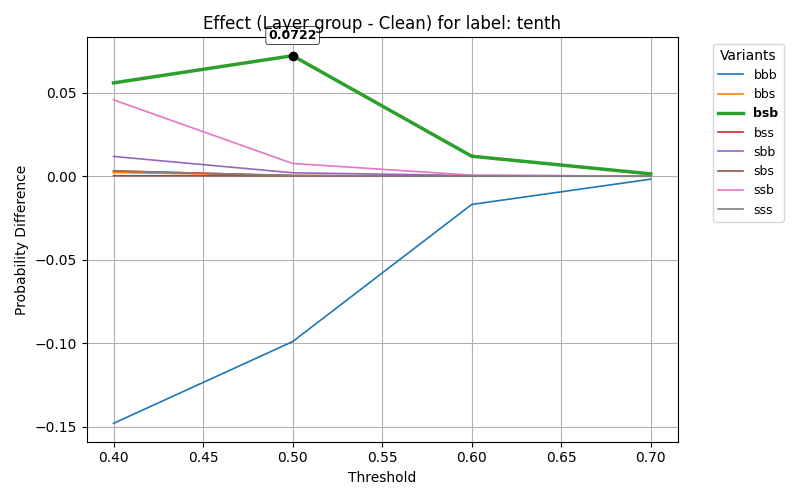} 
        \caption{Tens Circuit intervention}
    \end{subfigure}
    \begin{subfigure}[b]{0.32\textwidth}
        \centering
        \includegraphics[width=\textwidth]{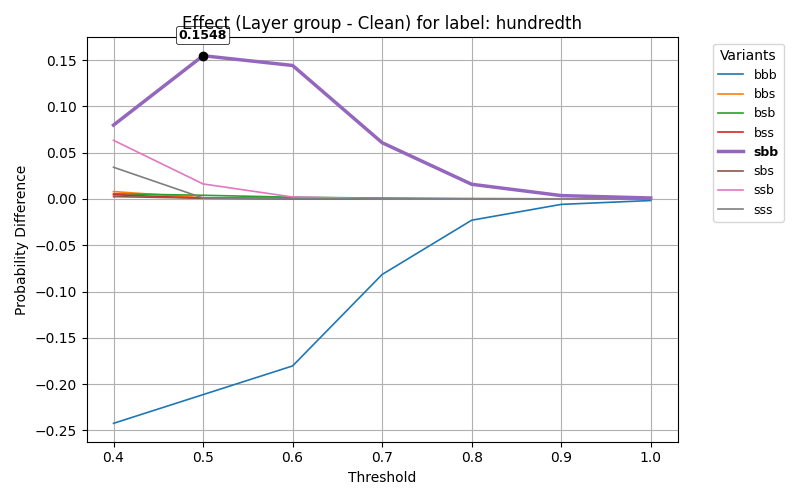}
        \caption{Hundreds Circuit Intervention}
    \end{subfigure}
    \caption{\textbf{Llama3-70B}, \(D_{sub, op1}\): Effect size of circuit specific interventions with different thresholds for neuron circuit membership, on circuit neurons in layers \(L = \{ 39, ..., 58\}\).}
    \label{fig:Threshold_Results_Llama70B_Subtraction_Op1}
\end{figure*}

\begin{figure*}[h!]
    \centering
    \hfill
    \begin{subfigure}[b]{0.32\textwidth}
        \centering
        \includegraphics[width=\textwidth]{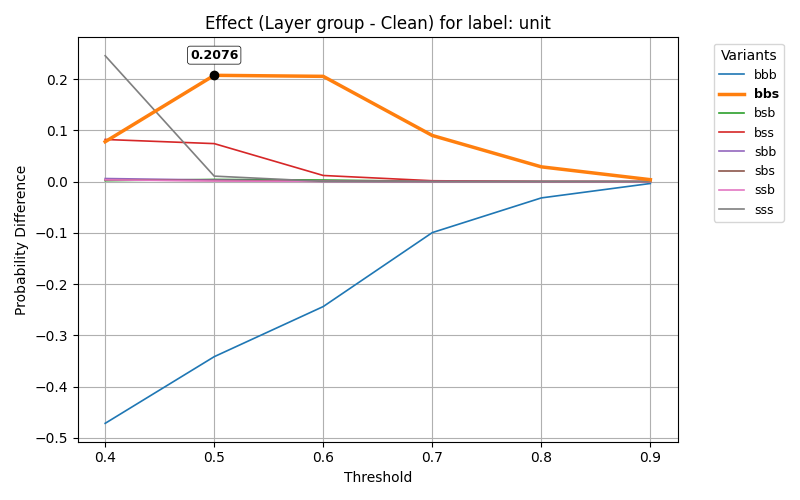}
        \caption{Unit Circuit intervention}
    \end{subfigure}
    \hfill
    \begin{subfigure}[b]{0.32\textwidth}
        \centering
        \includegraphics[width=\textwidth]{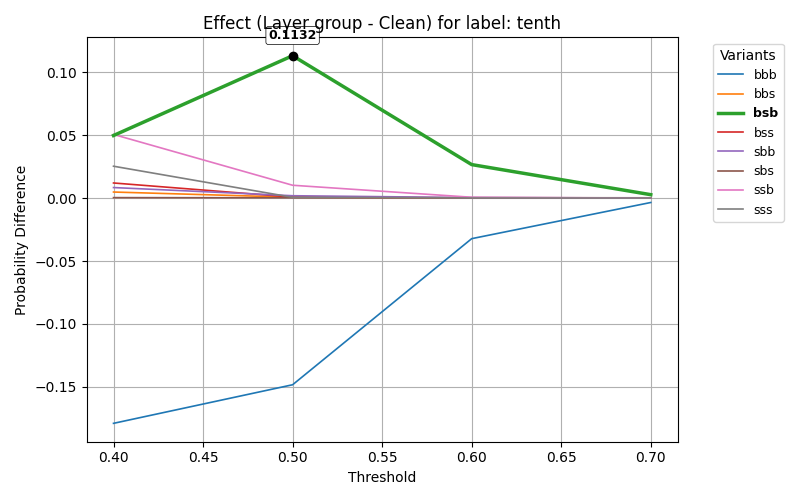} 
        \caption{Tens Circuit intervention}
    \end{subfigure}
    \begin{subfigure}[b]{0.32\textwidth}
        \centering
        \includegraphics[width=\textwidth]{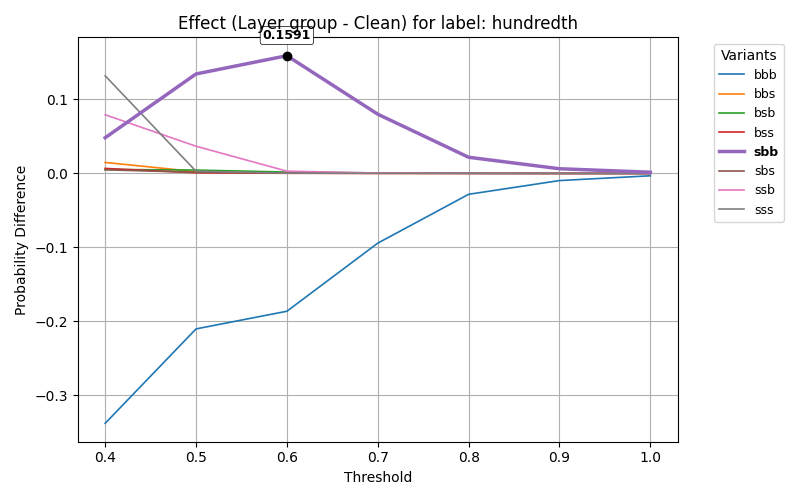}
        \caption{Hundreds Circuit Intervention}
    \end{subfigure}
    \caption{\textbf{Llama3-70B}, \(D_{sub, op2}\): Effect size of circuit specific interventions with different thresholds for neuron circuit membership, on circuit neurons in layers \(L = \{ 39, ..., 58\}\).}
    \label{fig:Threshold_Results_Llama70B_Subtraction_Op2}
\end{figure*}

\begin{figure*}[h!]
    \centering
    \hfill
    \begin{subfigure}[b]{0.32\textwidth}
        \centering
        \includegraphics[width=\textwidth]{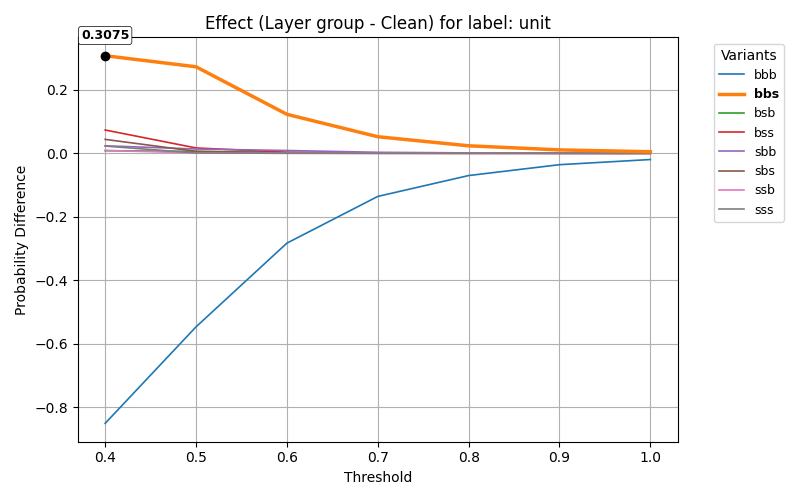}
        \caption{Unit Circuit intervention}
    \end{subfigure}
    \hfill
    \begin{subfigure}[b]{0.32\textwidth}
        \centering
        \includegraphics[width=\textwidth]{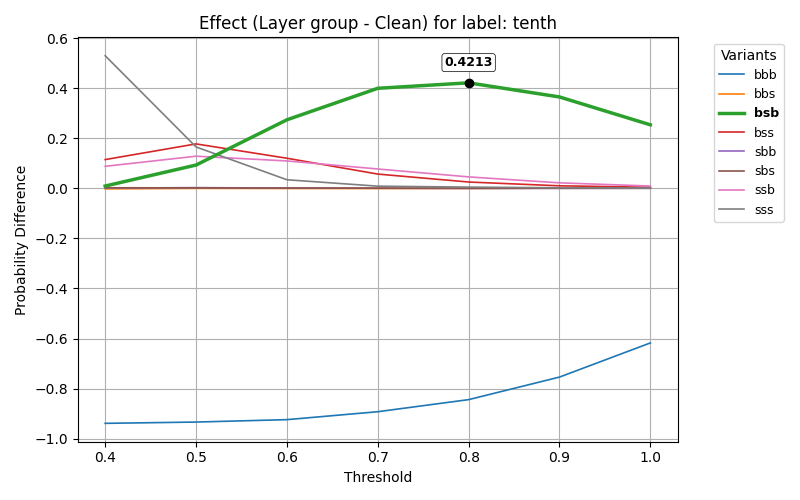} 
        \caption{Tens Circuit intervention}
    \end{subfigure}
    \begin{subfigure}[b]{0.32\textwidth}
        \centering
        \includegraphics[width=\textwidth]{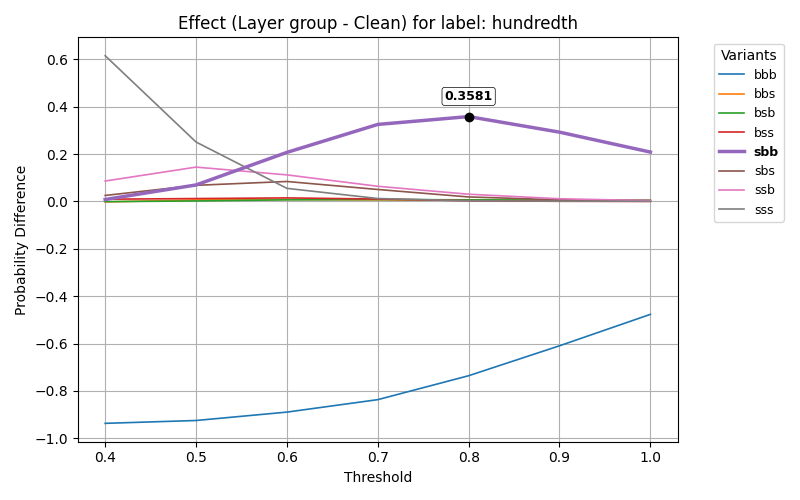}
        \caption{Hundreds Circuit Intervention}
    \end{subfigure}
    \caption{\textbf{Olmo 2 7B}, \(D_{add, op1}\): Effect size of circuit specific interventions with different thresholds for neuron circuit membership, on circuit neurons in layers \(L = \{ 17, ..., 30\}\).}
    \label{fig:Threshold_Results_Olmo2_Addition_Op1}
\end{figure*}

\begin{figure*}[h!]
    \centering
    \hfill
    \begin{subfigure}[b]{0.32\textwidth}
        \centering
        \includegraphics[width=\textwidth]{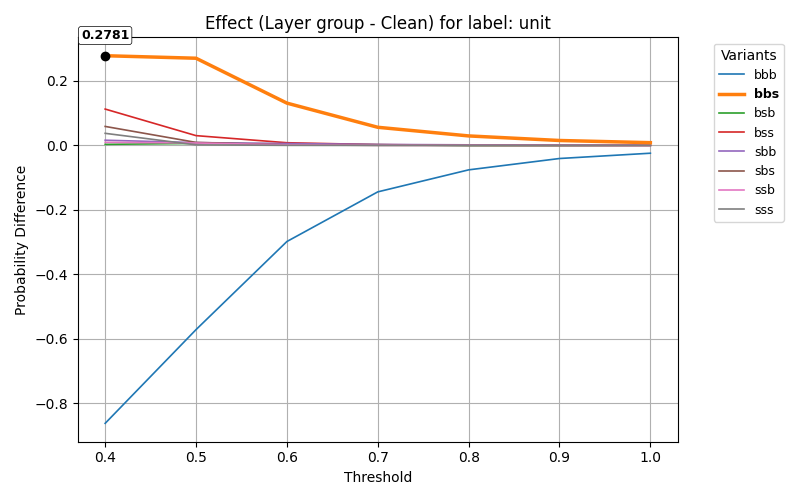}
        \caption{Unit Circuit intervention}
    \end{subfigure}
    \hfill
    \begin{subfigure}[b]{0.32\textwidth}
        \centering
        \includegraphics[width=\textwidth]{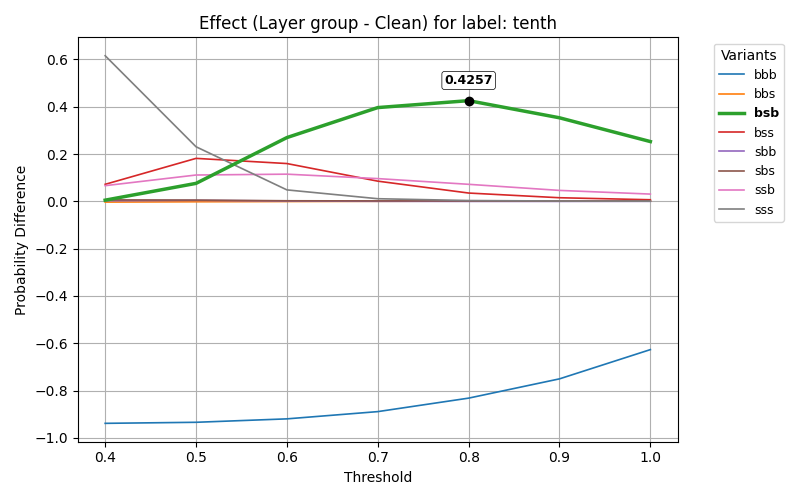} 
        \caption{Tens Circuit intervention}
    \end{subfigure}
    \begin{subfigure}[b]{0.32\textwidth}
        \centering
        \includegraphics[width=\textwidth]{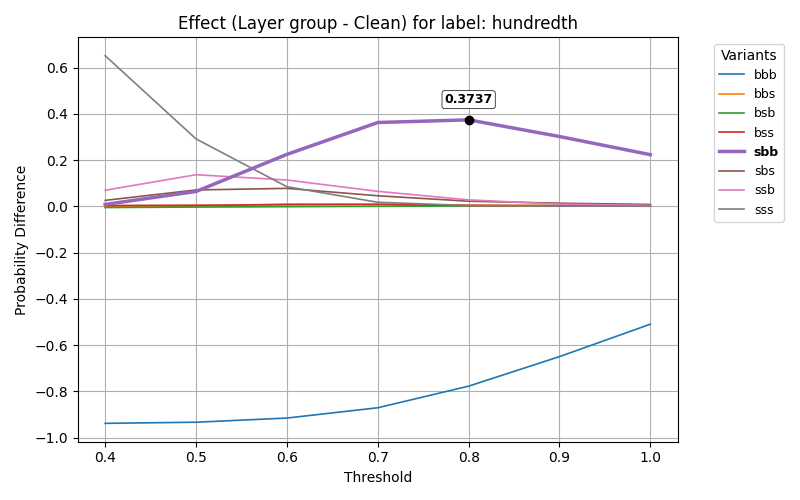}
        \caption{Hundreds Circuit Intervention}
    \end{subfigure}
    \caption{\textbf{Olmo 2 7B}, \(D_{add, op2}\): Effect size of circuit specific interventions with different thresholds for neuron circuit membership, on circuit neurons in layers \(L = \{ 17, ..., 30\}\).}
    \label{fig:Threshold_Results_Olmo2_Addition_Op2}
\end{figure*}

\begin{figure*}[h!]
    \centering
    \hfill
    \begin{subfigure}[b]{0.32\textwidth}
        \centering
        \includegraphics[width=\textwidth]{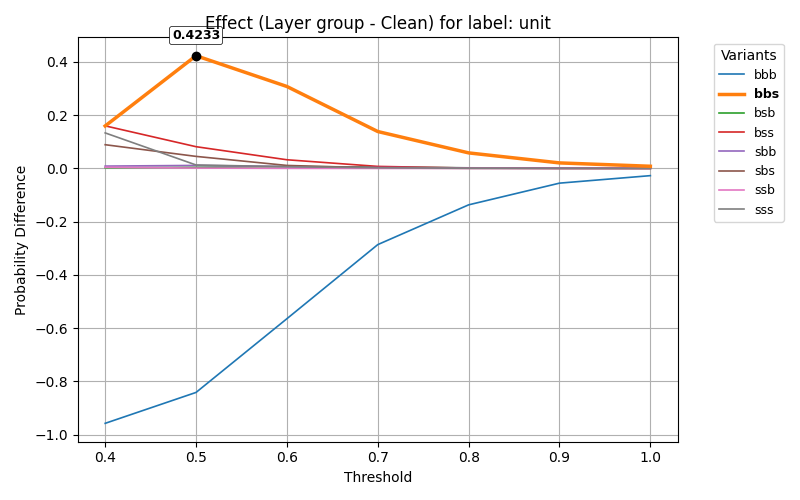}
        \caption{Unit Circuit intervention}
    \end{subfigure}
    \hfill
    \begin{subfigure}[b]{0.32\textwidth}
        \centering
        \includegraphics[width=\textwidth]{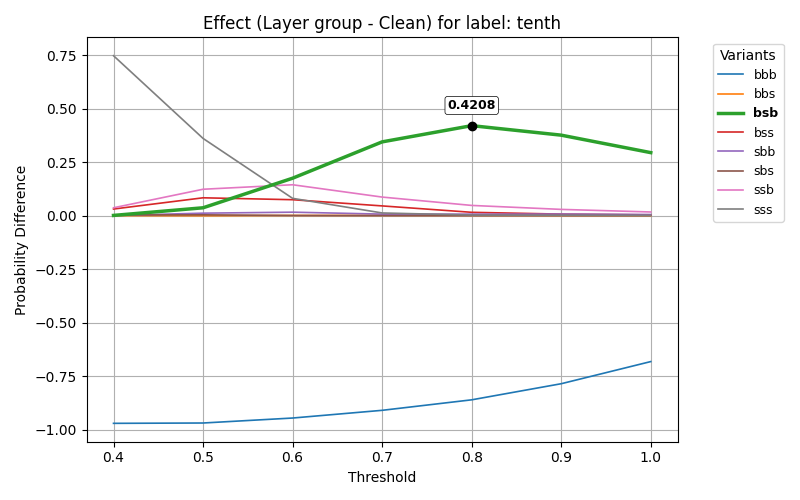} 
        \caption{Tens Circuit intervention}
    \end{subfigure}
    \begin{subfigure}[b]{0.32\textwidth}
        \centering
        \includegraphics[width=\textwidth]{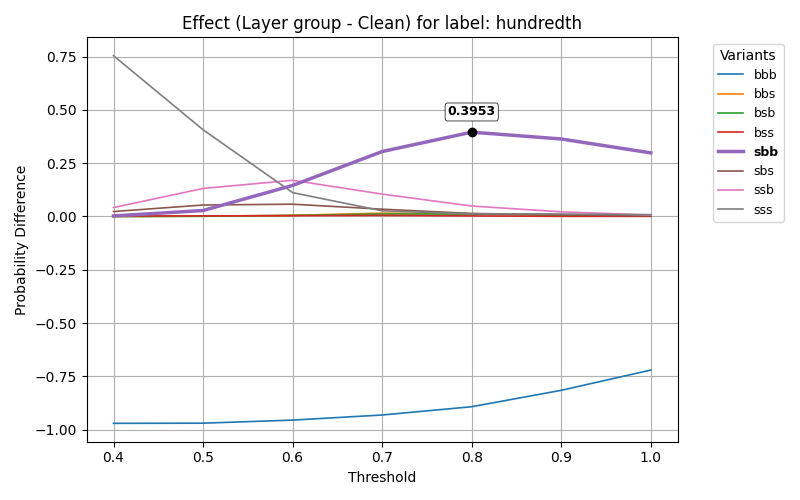}
        \caption{Hundreds Circuit Intervention}
    \end{subfigure}
    \caption{\textbf{Olmo 2 7B}, \(D_{sub, op1}\): Effect size of circuit specific interventions with different thresholds for neuron circuit membership, on circuit neurons in layers \(L = \{ 19, ..., 27\}\).}
    \label{fig:Threshold_Results_Olmo2_Subtraction_Op1}
\end{figure*}

\begin{figure*}[h!]
    \centering
    \hfill
    \begin{subfigure}[b]{0.32\textwidth}
        \centering
        \includegraphics[width=\textwidth]{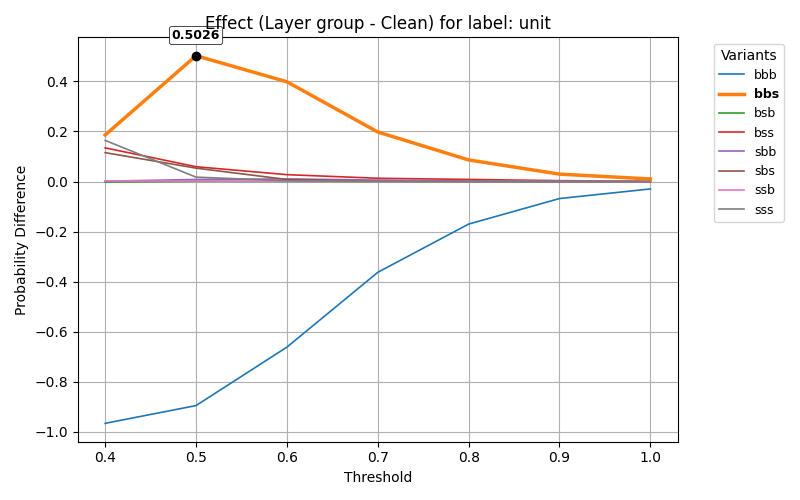}
        \caption{Unit Circuit intervention}
    \end{subfigure}
    \hfill
    \begin{subfigure}[b]{0.32\textwidth}
        \centering
        \includegraphics[width=\textwidth]{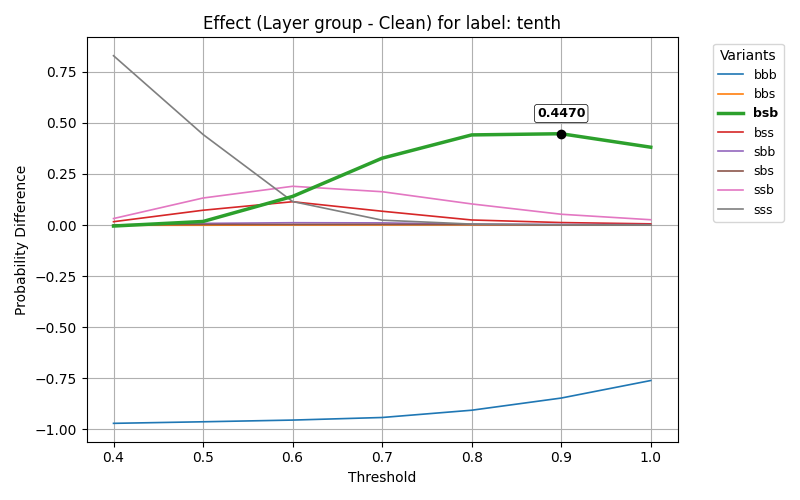} 
        \caption{Tens Circuit intervention}
    \end{subfigure}
    \begin{subfigure}[b]{0.32\textwidth}
        \centering
        \includegraphics[width=\textwidth]{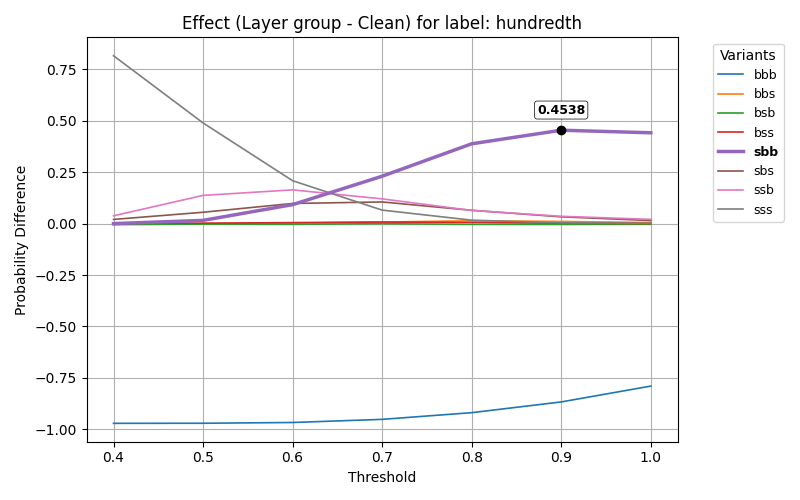}
        \caption{Hundreds Circuit Intervention}
    \end{subfigure}
    \caption{\textbf{Olmo 2 7B}, \(D_{sub, op2}\): Effect size of circuit specific interventions with different thresholds for neuron circuit membership, on circuit neurons in layers \(L = \{ 17, ..., 27\}\).}
    \label{fig:Threshold_Results_Olmo2_Subtraction_Op2}
\end{figure*}

\clearpage

\begin{table*}[ht]
\section{Intervention Results on \(D_{add, op1}\) and \(D_{sub, op1}\)}
\small
\centering
\resizebox{\textwidth}{!}{
\begin{tabular}{|c|c|c|c|c|c|c|c|c|c|c|c|}
\hline
\(m\) &\(\mathrm{o}\) & \(d\) &\(t*\) & bbb & \textbf{bbs} & \textbf{bsb} & \textbf{sbb} & bss & sbs & ssb & sss\\
\hline
\multicolumn{1}{|c|}{} & 
\multicolumn{1}{c|}{} & 
\multicolumn{1}{c|}{} & 
\multicolumn{1}{c|}{} & 
\multicolumn{8}{c|}{\textbf{Absolute change in prediction probability \textbf{\(\Delta p\)} in percentage points (after - before).}} \\
\hline
\multirow{6}{*}{Llama 3 8B} & \multirow{3}{*}{\(+\)} & unit & \(0.6\) & -55.74\% & \textbf{+30.17\%} & +1.01\% & +0.41\% & +3.65\% & +0.87\% & +0.30\% & +0.54\% \\
                           &  & tens & \(0.4\) & -70.44\% & +1.27\% & \textbf{+22.72\%} & +1.71\% & +4.82\% & +0.28\% & +7.52\% & +2.67\% \\
                           & & hun.s & \(0.8\) & -72.17\% & +0.25\% & +0.87\% & \textbf{+38.39\%} & +0.13\% & +0.71\% & +5.83\% & +0.40\% \\
                           \cline{2-12} 
                           & \multirow{3}{*}{\(-\)} & unit & \(0.6\) & -55.75\% & \textbf{+28.87\%} & +0.46\% & +1.12\% & +4.57\% & +1.20\% & +0.19\% & +0.63\% \\
                         &       & tens & \(0.5\) & -50.48\% & +1.12\% & \textbf{+14.39\%} & +2.11\% & +2.76\% & +0.48\% & +3.97\% & +0.69\% \\
                         &       & hun.s & \(0.8\) & -64.48\% & +0.46\% & +1.07\% & \textbf{+32.28\%} & +1.77\% & +0.62\% & +2.26\% & +0.41\% \\
                           \hline
\multirow{6}{*}{\text{Llama 3 70B}} 
& \multirow{3}{*}{\(+\)} 
& \text{unit} & 0.5 & -28.51\% & \textbf{+20.12\%} & +0.48\% & +0.97\% & +1.66\% & +0.45\% & +0.15\% & +0.40\% \\
&  & \text{tens} & 0.4 & -35.24\% & +0.67\% & \textbf{+14.60\%} & +1.75\% & +0.84\% & +0.12\% & +6.24\% & +0.93\% \\
&  & \text{hun.s} & 0.6 & -37.65\% & +0.08\% & +0.61\% & \textbf{+24.70\%} & +0.12\% & +0.20\% & +1.63\% & +0.18\% \\
\cline{2-12}
& \multirow{3}{*}{\(-\)} 
& \text{unit} & 0.5 & -27.45\% & \textbf{+19.65\%} & +0.22\% & +0.49\% & +3.72\% & +0.43\% & +0.07\% & +0.66\% \\
&  & \text{tens} & 0.5 & -9.52\% & +0.41\% & \textbf{+7.22\%} & +0.21\% & +0.04\% & +0.18\% & +0.76\% & +0.03\% \\
&  & \text{hun.s} & 0.5 & -21.14\% & +0.13\% & +0.40\% & \textbf{+15.47\%} & +0.08\% & +0.09\% & +1.62\% & +0.11\% \\
\hline
\multirow{6}{*}{Olmo 2 7B} & \multirow{3}{*}{\(+\)} & unit & \(0.4\) & -85.04\% & \textbf{+30.75\%} & +0.83\% & +7.34\% & +2.07\% & +4.39\% & +0.90\% & +2.28\% \\
                       &      & tens & \(0.8\) & -84.41\% & -0.06\% & \textbf{+42.13\%} & +2.51\% & -0.05\% & +0.04\% & +4.56\% & +0.47\% \\
                       &      & hun.s & \(0.8\) & -73.61\% & +0.55\% & +0.69\% & \textbf{+35.81\%} & +0.30\%  & +1.88\% & +3.04\% & +0\% \\
                           \cline{2-12} 
                           & \multirow{3}{*}{\(-\)} & unit & \(0.5\) & -84.73\% & \textbf{+42.33\%} & +0.42\% & +1.13\% & +8.14\% & +4.50\% & +0.13\% & +1.30\% \\
                           &  & tens & \(0.8\) & -86.03\% & +0.14\% & \textbf{+42.08\%} & +0.73\% & +1.58\% & +0.03\% & +4.80\% & +0.30\% \\
                           &  & hun.s & \(0.8\) & -89.26\% & +1.29\% & +1.11\% & \textbf{+39.53\%} & +0.25\% & +1.40\% & +0.49\% & +1.22\% \\
\hline
\end{tabular}
}
\caption{Main Results: For all detected circuits (across models, operators, and digit positions) we report the change in prediction probabilities for result variants after interventions on digit-position-specific arithmetic circuits for the best threshold \(t*\) in each circuit, on datasets \(D_{add, op1}\) and \(D_{sub, op1}\). The increase in prediction probability for the correct digit-specific result variant is shown in bold.}
\label{tab:mainresultsop1}
\end{table*}

\begin{table*}[h]
\centering
\small
\begin{tabular}{|c|c|c|c|c|}
\hline
\(m\) &\(\mathrm{o}\) & \(d\) &\(t*\) & Flip Rate\\
\hline
\multirow{6}{*}{Llama 3 8B} & \multirow{3}{*}{\(+\)} & unit & \(0.6\) & 42.5\% \\
                           &  & tens & \(0.4\) & 38.5\% \\
                           & & hun.s & \(0.8\) & 58.5\% \\
                           \cline{2-5} 
                           & \multirow{3}{*}{\(-\)} & unit & \(0.6\) & 36.5\% \\
                           &  & tens & \(0.5\) & 19\% \\
                           & & hun.s & \(0.8\) & 47.5\% \\
                           \hline
\multirow{6}{*}{Llama 3 70B} & \multirow{3}{*}{\(+\)} & unit & \(0.5\) & 17.5\% \\
                           &  & tens & \(0.4\) & 13\% \\
                           & & hun.s & \(0.6\) & 24.5\% \\
                           \cline{2-5} 
                           & \multirow{3}{*}{\(-\)} & unit & \(0.5\) & 16.5\% \\
                           &  & tens & \(0.5\) & 3.5\% \\
                           & & hun.s & \(0.5\) & 11\% \\
                           \hline
\multirow{6}{*}{Olmo 2 7B} & \multirow{3}{*}{\(+\)} & unit & \(0.4\) & 50.5\% \\
                           &  & tens & \(0.8\) & 53\% \\
                           & & hun.s & \(0.8\) & 45\% \\
                           \cline{2-5} 
                           & \multirow{3}{*}{\(-\)} & unit & \(0.5\) & 56\% \\
                           &  & tens & \(0.8\) & 53\% \\
                           & & hun.s & \(0.8\) & 48\% \\
                           \hline
\end{tabular}
\caption{Flip rate from \texttt{bbb} result to the intended digit-specific result variant (Unit: \texttt{bbb} $\rightarrow$ \texttt{bbs},Tens: \texttt{bbb} $\rightarrow$ \texttt{bsb}, Hundreds: \texttt{bbb} $\rightarrow$ \texttt{sbb}), results given for the best threshold \(t*\) for each circuit, on datasets \(D_{add, op1}\) and \(D_{sub, op1}\).}
\label{tab:fliprateop1}
\end{table*}
\begin{figure*}[h!]
\section{Ablation: Deeper Circuits}
\label{app:threshold_layer28}
    \centering
    \hfill
    \begin{subfigure}[b]{0.32\textwidth}
        \centering
        \includegraphics[width=\textwidth]{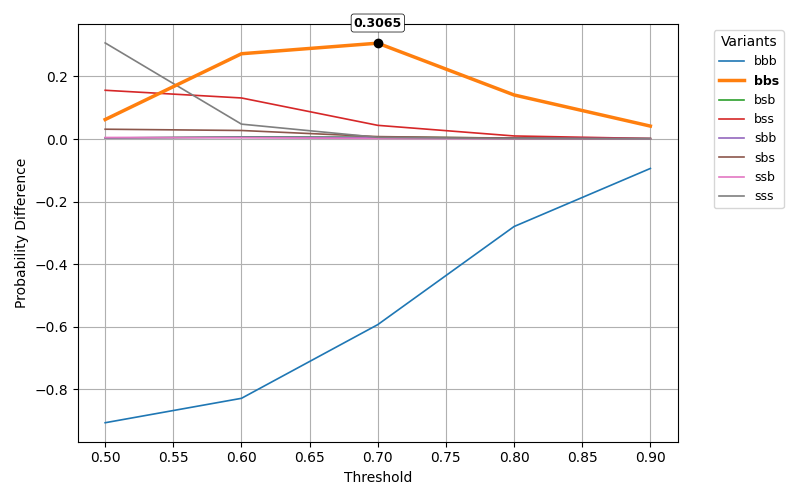}
        \caption{Unit Circuit intervention}
    \end{subfigure}
    \hfill
    \begin{subfigure}[b]{0.32\textwidth}
        \centering
        \includegraphics[width=\textwidth]{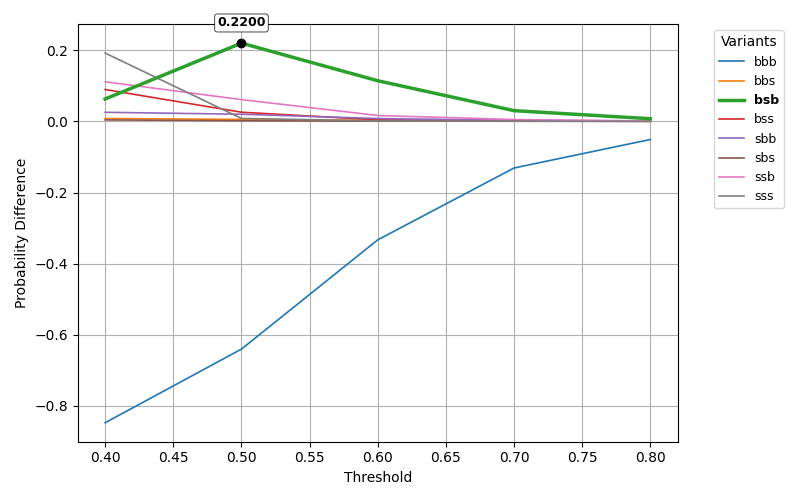} 
        \caption{Tens Circuit intervention}
    \end{subfigure}
    \begin{subfigure}[b]{0.32\textwidth}
        \centering
        \includegraphics[width=\textwidth]{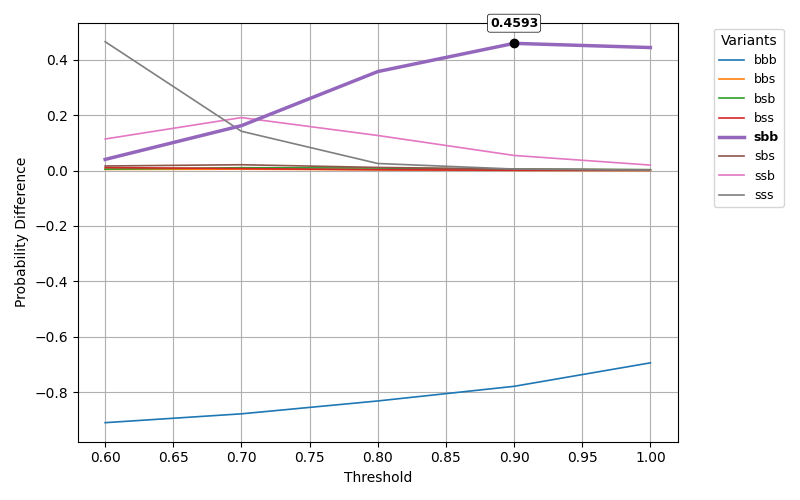}
        \caption{Hundreds Circuit Intervention}
    \end{subfigure}
    \caption{Effect size of circuit specific interventions on Llama 3 8B and \(D_{add, op2}\). We intervene on a deeper circuit (\(L = \{15, ..., 28\}\)) and find no significant difference to the shallower circuit depth (\(L = \{15, ..., 24\}\)) chosen based on the statistics on circuit sufficiency and size.}
    \label{fig:Threshold_Results_28}
\end{figure*}

\clearpage

\begin{table*}[h!]
\section{Similarity of Addition and Subtraction Circuits}
\label{app:simaddsub}
\centering
\scriptsize
\begin{tabular}{|c|c|ccccc|}
\hline
\textbf{Layer} & \textbf{Digit Pos} & \textbf{Top-50} & \textbf{Top-100} & \textbf{Top-250} & \textbf{Top-500} & \textbf{Top-1000} \\ \hline
\multirow{3}{*}{15} & unit & 12.0\% & 17.0\% & 12.8\% & 7.2\% & 3.6\% \\
 & tens & 4.0\% & 2.0\% & 0.8\% & 0.4\% & 0.2\% \\
 & hundreds & 10.0\% & 10.0\% & 15.6\% & 18.2\% & 16.0\% \\
\hline
\multirow{3}{*}{16} & unit & 16.0\% & 14.0\% & 5.6\% & 2.8\% & 1.4\% \\
 & tens & 8.0\% & 4.0\% & 1.6\% & 0.8\% & 0.4\% \\
 & hundreds & 16.0\% & 15.0\% & 11.6\% & 5.8\% & 2.9\% \\
\hline
\multirow{3}{*}{17} & unit & 36.0\% & 32.0\% & 36.4\% & 26.0\% & 13.0\% \\
 & tens & 2.0\% & 1.0\% & 0.4\% & 0.2\% & 0.1\% \\
 & hundreds & 18.0\% & 18.0\% & 20.0\% & 10.0\% & 5.0\% \\
\hline
\multirow{3}{*}{18} & unit & 24.0\% & 25.0\% & 28.0\% & 15.6\% & 7.8\% \\
 & tens & 6.0\% & 3.0\% & 1.2\% & 0.6\% & 0.3\% \\
 & hundreds & 40.0\% & 32.0\% & 28.4\% & 31.0\% & 22.4\% \\
\hline
\multirow{3}{*}{19} & unit & 16.0\% & 18.0\% & 8.0\% & 4.0\% & 2.0\% \\
 & tens & 6.0\% & 3.0\% & 1.2\% & 0.6\% & 0.3\% \\
 & hundreds & 48.0\% & 44.0\% & 44.0\% & 42.8\% & 44.8\% \\
\hline
\multirow{3}{*}{20} & unit & 14.0\% & 7.0\% & 2.8\% & 1.4\% & 0.7\% \\
 & tens & 6.0\% & 3.0\% & 1.2\% & 0.6\% & 0.3\% \\
 & hundreds & 44.0\% & 45.0\% & 42.0\% & 40.0\% & 36.8\% \\
\hline
\multirow{3}{*}{21} & unit & 28.0\% & 29.0\% & 29.2\% & 14.6\% & 7.3\% \\
 & tens & 26.0\% & 34.0\% & 21.6\% & 10.8\% & 5.4\% \\
 & hundreds & 28.0\% & 26.0\% & 29.2\% & 25.0\% & 12.5\% \\
\hline
\multirow{3}{*}{22} & unit & 4.0\% & 7.0\% & 3.6\% & 1.8\% & 0.9\% \\
 & tens & 26.0\% & 26.0\% & 20.4\% & 10.2\% & 5.1\% \\
 & hundreds & 4.0\% & 3.0\% & 6.0\% & 5.6\% & 5.5\% \\
\hline
\multirow{3}{*}{23} & unit & 18.0\% & 17.0\% & 12.4\% & 6.2\% & 3.1\% \\
 & tens & 4.0\% & 4.0\% & 5.6\% & 3.6\% & 1.8\% \\
 & hundreds & 2.0\% & 5.0\% & 2.8\% & 1.4\% & 0.7\% \\
\hline
\multirow{3}{*}{24} & unit & 26.0\% & 24.0\% & 24.4\% & 24.2\% & 12.1\% \\
 & tens & 12.0\% & 12.0\% & 7.6\% & 3.8\% & 1.9\% \\
 & hundreds & 0.0\% & 0.0\% & 0.0\% & 0.0\% & 0.0\% \\
\hline
\end{tabular}
\caption{Overlap of Top-K Fisher Score neurons between addition and subtraction circuits across layers and digit positions.}
\label{tab:add_sub}
\end{table*}
\begin{figure}[t]
\section{Examples of Neuron Heuristics}
\label{app:heuristics}
    \centering
    \includegraphics[width=0.44\textwidth]{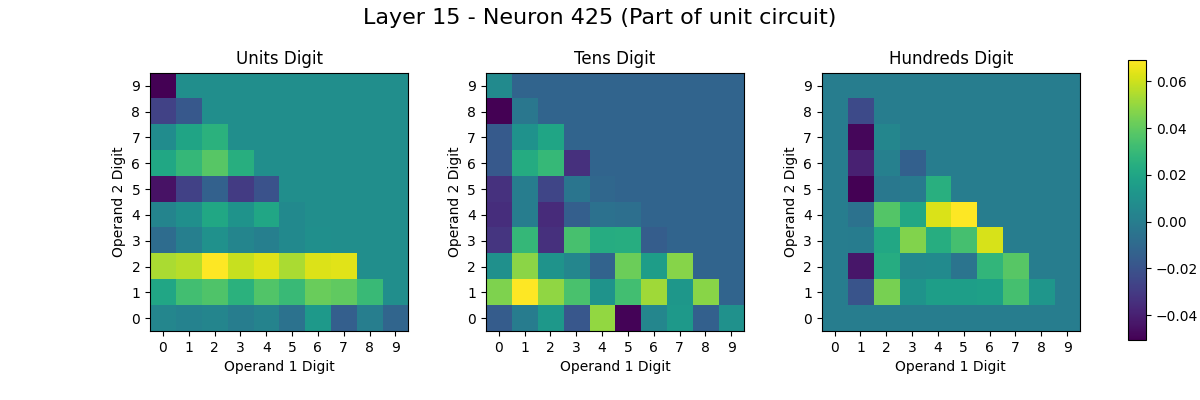} %
    \small
    \caption{MLP neuron \(N_{15, 425}\) is part of unit circuit - Heuristic: Operand 2 is \(2\) in unit digit position.}
    \label{fig:heurisit_op2}
\end{figure}

\begin{figure}[t]
    \centering
    \includegraphics[width=0.44\textwidth]{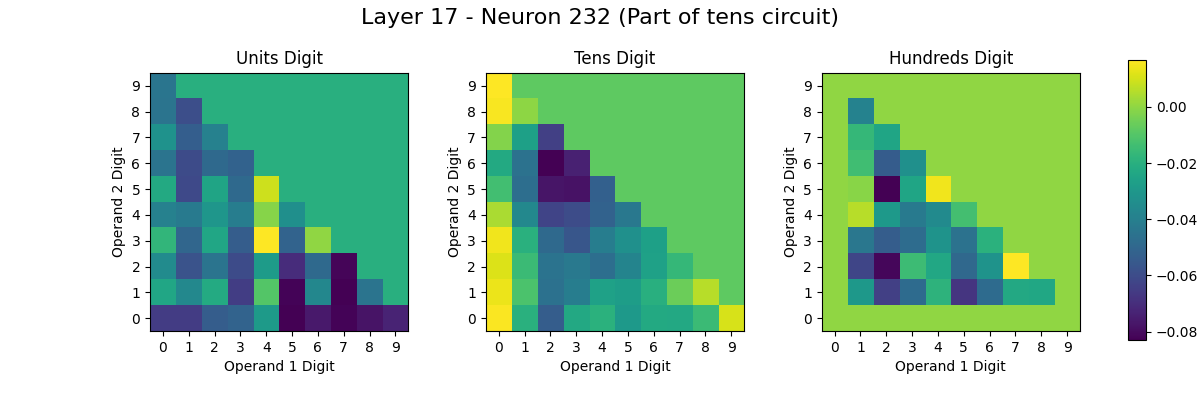} %
    \small
    \caption{MLP neuron \(N_{17, 232}\) is part of tens circuit - Heuristic: Operand 1 is \(0\) in tens digit position.}
    \label{fig:heurisit_op1}
\end{figure}

\begin{figure}[t]
    \centering
    \includegraphics[width=0.44\textwidth]{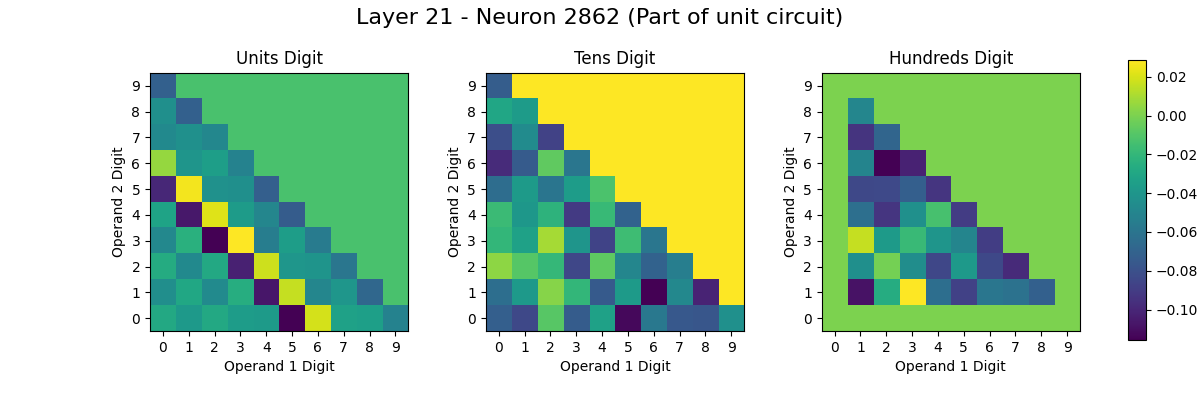} 
    \small
    \caption{MLP neuron \(N_{21, 2862}\) is part of unit circuit - Heuristic: Result is \(6\) in unit digit position.}
    \label{fig:heurisit_resultunit}
\end{figure}

\end{document}